\documentclass{article}

\usepackage{PRIMEarxiv}

\usepackage[utf8]{inputenc} 
\usepackage[T1]{fontenc}    
\usepackage{hyperref}       
\usepackage{url}            
\usepackage{booktabs}       
\usepackage{amsfonts}       
\usepackage{nicefrac}       
\usepackage{microtype}      
\usepackage{lipsum}
\usepackage{fancyhdr}       
\usepackage{graphicx}       
\usepackage{amsmath}
\graphicspath{{media/}}     

\title{Review of Inference-Time Scaling Strategies: Reasoning, Search and RAG}

\author{
    Zhichao Wang\thanks{Corresponding author} \\ 
    Inflection AI \\
    \texttt{zcwang0201@gmail.com}
    \And
    Cheng Wang \\
    Georgia Institute of Technology
    \And
    Dong Nie \\
    ChatAlpha AI
}

\begin{document}
\maketitle

\begin{abstract}
The performance gains of LLMs have historically been driven by scaling up model size and training data. However, the rapidly diminishing availability of high-quality training data is introducing a fundamental bottleneck, shifting the focus of research toward inference-time scaling. This paradigm uses additional computation at the time of deployment to substantially improve LLM performance on downstream tasks without costly model re-training. This review systematically surveys the diverse techniques contributing to this new era of inference-time scaling, organizing the rapidly evolving field into two comprehensive perspectives: Output-focused and Input-focused methods. Output-focused techniques encompass complex, multi-step generation strategies, including reasoning (e.g., CoT, ToT, ReAct), various search and decoding methods (e.g., MCTS, beam search), training for long CoT (e.g., RLVR, GRPO), and model ensemble methods. Input-focused techniques are primarily categorized by few-shot and RAG, with RAG as the central focus. The RAG section is further detailed through a structured examination of query expansion, data, retrieval and reranker, LLM generation methods, and multi-modal RAG.
\end{abstract}


\begin{figure}
    \centering    \includegraphics[width=\linewidth]{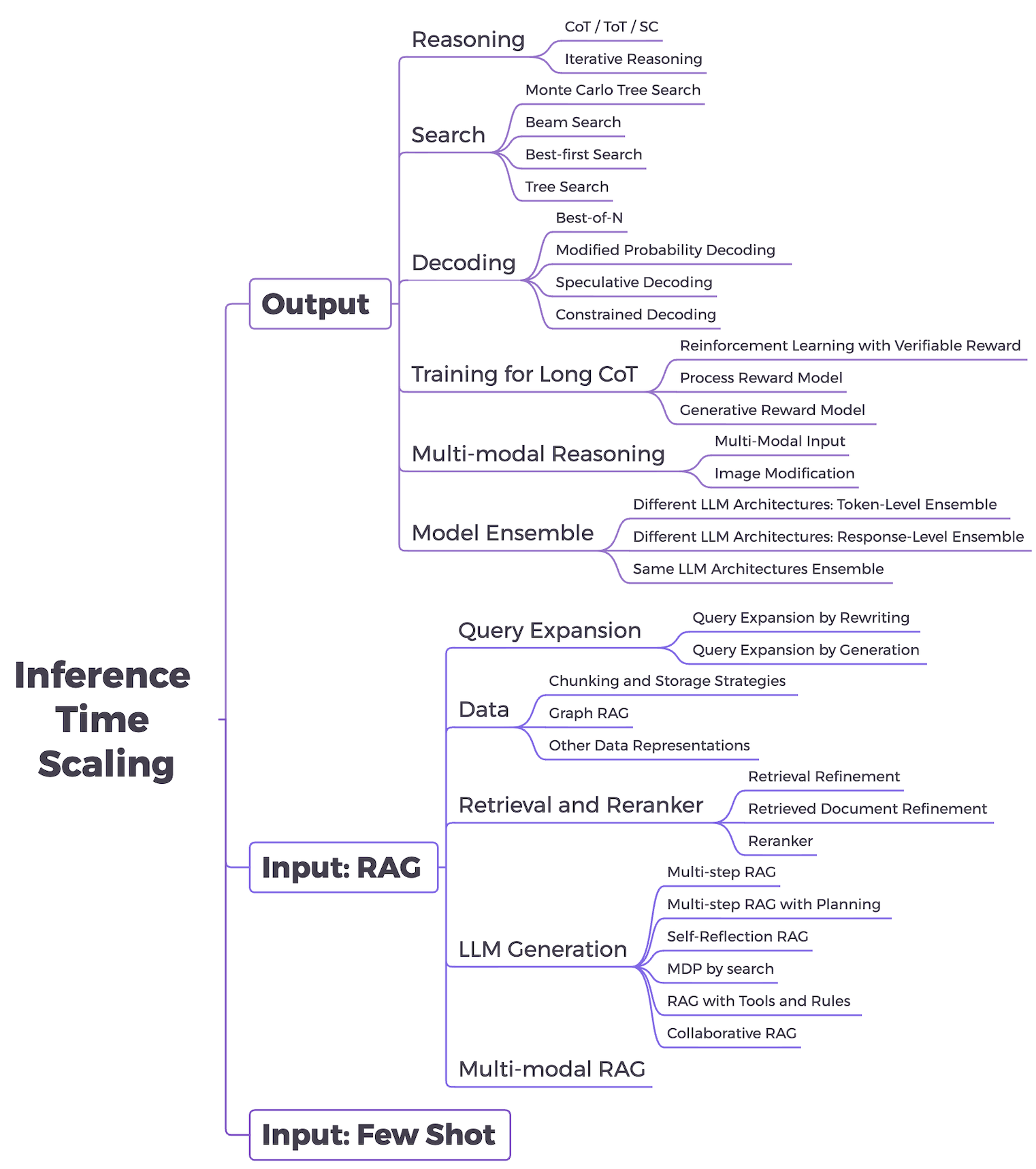}
    \caption{From the output side, different techniques will be discussed, including: 1) reasoning methods like CoT, ToT, and ReAct; 2) search methods like MCTS and beam search; 3) decoding methods like Best-of-N, speculative decoding, and constrained decoding; 4) training for long CoT like RLVR and GRPO; 5) multi-modal reasoning; and 6) model ensemble. For the input side, it is further divided into RAG and Few-Shot. In RAG, it will be discussed from the perspectives of: 1) query expansion, 2) data, 3) retrieval and reranker, 4) LLM generation, and 5) multi-modal RAG.}
    \label{fig:InferenceTimeScaling}
\end{figure}

\section{Introduction}

The rapid advancement of large language models (LLMs) transformed the landscape of natural language processing (NLP), enabling breakthroughs in tasks from text generation to complex problem-solving. Historically, researchers focused on scaling training processes by expanding computational resources—measured in floating-point operations (FLOPs)—and enlarging both model size and training dataset diversity. This strategy, known as pre-training scaling, drove remarkable progress in systems such as GPT-4 and its successors \cite{openai2024gpt4ocard}. However, the supply of high-quality and diverse training data became a critical bottleneck. Curating massive, reliable, and unbiased datasets was resource-intensive and challenging, rendering further gains through larger models unsustainable. As a result, researchers began to explore new directions for improving LLM capabilities without increasing model size.

In recent years, a paradigm shift emerged toward inference-time scaling, which enhanced model performance by allocating additional computation during inference. Unlike pre-training scaling, which consumed vast resources during model training, inference-time scaling optimizes the use of computation after deployment. This approach allowed models to refine outputs or process inputs more effectively, leading to higher accuracy, improved reasoning, and greater adaptability without costly retraining. This paper comprehensively reviews inference-time scaling techniques, organized into output-focused and input-focused methods as shown in Figure \ref{fig:InferenceTimeScaling}.

From an output perspective, we examined techniques that improve the quality of model-generated results during inference. These included reasoning-based approaches such as Chain-of-Thought (CoT), which encouraged step-by-step problem-solving; Tree-of-Thought (ToT), which explored multiple reasoning paths; and Reason+Act (ReAct), which combined reasoning and action in interactive settings. Search-based methods such as Monte Carlo Tree Search (MCTS) and beam search balanced exploration and exploitation in sequence generation. We also reviewed decoding strategies—Best-of-N sampling, speculative decoding, and constrained decoding—that optimized fluency, efficiency, or structure. In addition, we discussed fine-tuning strategies such as Reinforcement Learning with Verifiable Rewards (RLVR) and Group Relative Policy Optimization (GRPO), which trained models to produce more coherent reasoning chains. Finally, we included multi-modal reasoning and model ensembling, both of which expanded LLM capabilities across tasks and modalities.

From the input perspective, we analyzed methods that increased the length of prompt to scale inference FLOPs and improved the quality of inputs. These included few-shot learning, which enabled rapid adaptation from limited examples, and Retrieval-Augmented Generation (RAG), which integrated external knowledge into generation. While few-shot learning often appeared as an engineering practice rather than a core research focus, RAG became central to input-based scaling. We reviewed its major components—query expansion, data, retrieval and reranker, LLM generation, and multi-modal extensions.

Finally, we selected topics based on whether they increased computational FLOPs during inference to enhance downstream task performance. Traditional inference-time scaling focused mainly on output-level methods like CoT or MCTS. In this review, we also included input-based approaches, such as few-shot prompting and RAG, as longer prompts proportionally increase inference FLOPs within the context window. Although some techniques—like speculative decoding for acceleration, constrained decoding for structured output, or model merging for performance enhancement—did not directly increase FLOPs, they nonetheless improved inference-time capability and thus warranted inclusion. Through this synthesis, we aimed to establish a clearer understanding of inference-time scaling as a unifying framework for advancing LLM efficiency and intelligence.

\section{Output}
\subsection{Reasoning Techniques in LLMs}

\begin{figure}
    \centering
    \includegraphics[width=0.9\linewidth]{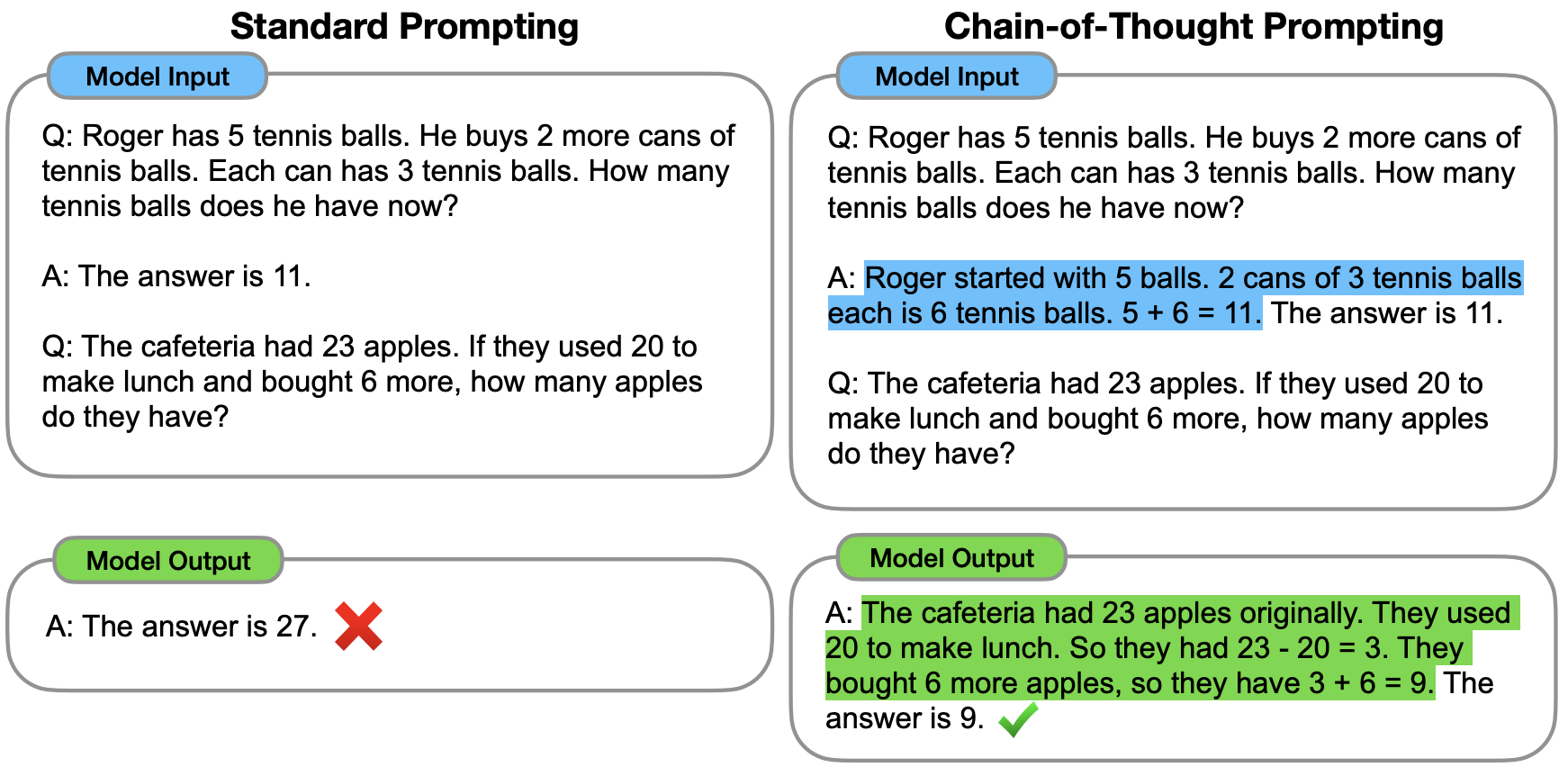}
    \caption{CoT: LLM is asked to generate chain of thought before generating the final answer with few-shot examples of prompt, CoT and answer. For zero-shot case, it will use "Let’s think step by step" to encourage LLM to think before generating the final answer.}
    \label{fig: CoT}
\end{figure}

\subsubsection{Prompting-Based Reasoning Techniques}

Prompting-based reasoning techniques enable LLMs to perform multi-step reasoning through structured prompt design, without modifying model parameters. These methods progressively evolve from simple step-by-step prompting to more sophisticated frameworks that enhance reasoning diversity, structure, and consistency. Starting with Chain-of-Thought (CoT), which elicits explicit intermediate reasoning steps, subsequent approaches such as Self-Consistency (SC), Tree-of-Thought (ToT), and Graph-of-Thought (GoT) introduce mechanisms like ensemble reasoning, tree search, and graph-based deliberation to improve accuracy, robustness, and scalability in complex reasoning tasks.

\textbf{CoT} As one of the earliest prompting-based reasoning techniques, CoT laid the foundation for structured reasoning in LLMs. CoT \cite{wei2023chainofthoughtpromptingelicitsreasoning} guides models to solve problems by presenting examples of step-by-step reasoning that lead to the final answer, as illustrated in Figure~\ref{fig: CoT}. In Zero-shot CoT, the prompt appends a simple instruction, such as "Let us think step by step," to encourage the model to reason sequentially. In Few-shot CoT, the prompt includes a few examples of queries, intermediate reasoning steps, and final answers, followed by the same verification instruction. When faced with new problems, the model generates its own reasoning steps before producing the answer, which improves interpretability. Experiments on arithmetic, commonsense, and symbolic reasoning tasks show that CoT significantly boosts performance for large-scale models (over 100 billion parameters) on complex tasks. However, for simpler tasks, improvements are minimal, as the model may skip unnecessary steps.

\begin{figure}
    \centering
    \includegraphics[width=\linewidth]{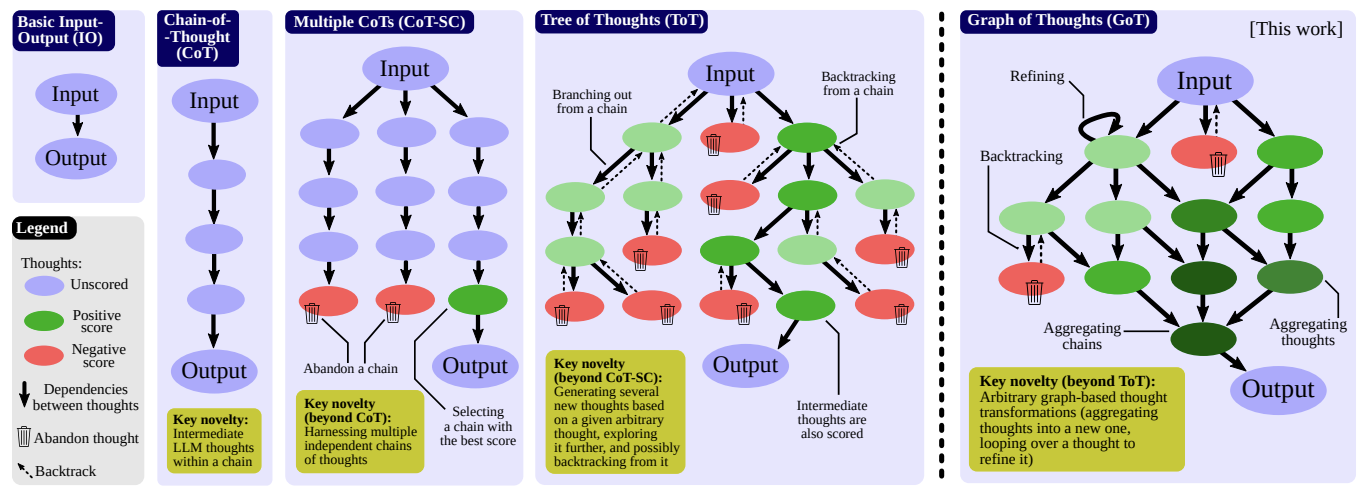}
    \caption{Comparison among: 1. CoT, 2. SC, 3. ToT, and 4. GoT}
    \label{fig:CoT_ToT_SC_GoT}
\end{figure}

\textbf{SC} Building on CoT, the SC approach refines the reasoning process by incorporating diversity and consensus during decoding. In \cite{wang2023selfconsistencyimproveschainthought}, the authors extend CoT into CoT-SC by replacing conventional greedy decoding with a majority voting mechanism to enhance reasoning reliability:
\begin{itemize}
    \item \textit{Paths Generation}: produces multiple paths using (temperature) sampling.
    \item \textit{Majority Voting}: selects the final answer by aggregating the most frequent outcome.
\end{itemize}
This sample-and-marginalize approach leverages the insight that complex problems often have multiple valid reasoning paths, and convergence on the same answer across independent paths increases confidence in its correctness. Building on this, \cite{chen2023universalselfconsistencylargelanguage} proposes universal self-consistency (USC), extending the self-consistency framework to free-form generation tasks where no single correct answer exists. USC generates multiple reasoning paths using a LLM and employs the LLM as a judge to evaluate responses. The model selects the most consistent response based on a prompt, such as "Evaluate these responses and select the most consistent based on majority consensus". Further advancing the concept, \cite{wan2023betterzeroshotreasoningselfadaptive} introduces a two-stage framework to improve zero-shot reasoning by dynamically generating few-shot examples. In the first stage, an LLM produces multiple answers in a zero-shot setting, retaining only those that align with the majority prediction (self-consistency). These answers are further filtered based on three criteria: low entropy, low repetitiveness, and high diversity, resulting in a set of top-k samples. In the second stage, a new prompt incorporates these top-k samples to generate a fresh set of answers, with the final answer selected via majority voting. This approach enhances the robustness and adaptability of zero-shot reasoning in LLMs.

\textbf{ToT} Moving beyond linear reasoning, the ToT framework extends CoT into a structured search paradigm that explores multiple reasoning paths in parallel. Proposed by Yao et al.~\cite{yao2023treethoughtsdeliberateproblem}, ToT generalizes the CoT approach by representing reasoning as a tree search over coherent intermediate steps (“thoughts”) rather than a linear, top-down sequence, as illustrated in Figure \ref{fig:CoT_ToT_SC_GoT}. At each step, multiple candidate thoughts are generated and self-evaluated, which allows for:
\begin{itemize}
    \item \textit{Lookahead}: anticipates future steps to guide decision-making effectively.
    \item \textit{Pruning}: eliminates less promising steps to focus on viable steps.
    \item \textit{Backtracking}: revisits earlier steps to correct or explore alternative steps.
\end{itemize}
This deliberate problem-solving process significantly enhances performance on tasks requiring planning, exploration, or global consistency. In \cite{haji2024improvingllmreasoningmultiagent}, the ToT framework is enhanced by integrating a thought validator to filter out invalid branches. When a query is received, multiple reasoner agents perform parallel ToT processes to decompose the query and generate diverse next steps. Each generated thought is evaluated for (1) logical consistency, (2) factual accuracy, and (3) completeness. Flawed paths are discarded, and feedback is provided for refinement. A consensus voting mechanism then determines the final answer: if consensus is reached, the answer is returned; otherwise, additional iterations of reasoning are conducted.

The Graph of Thoughts (GoT) framework, proposed in \cite{Besta_2024}, further extends CoT and ToT, as shown in Figure \ref{fig:CoT_ToT_SC_GoT}. GoT introduces two key operations: (1) refinement, where a thought is iteratively improved, and (2) aggregation, where multiple thoughts are combined into a new thought. Lastly, a comparison among CoT, SC, ToT, GoT is in Table \ref{tab:reasoning_techniques}.

\begin{table}[h]
\centering
\begin{tabular}{l|ll}
\hline
\textbf{Technique} & \textbf{Key Mechanism} & \textbf{Advantages} \\
\hline
CoT\cite{wei2023chainofthoughtpromptingelicitsreasoning} & Step-by-step prompting & Improves interpretability on complex tasks \\
SC\cite{wang2023selfconsistencyimproveschainthought} & Ensemble voting on diverse paths & Enhances reliability and robustness \\
ToT\cite{yao2023treethoughtsdeliberateproblem} & Tree search with lookahead/pruning & Better for planning and exploration \\
GoT\cite{Besta_2024} & Graph with refinement/aggregation & Scalable for complex deliberation \\
\hline
\end{tabular}
\caption{Comparison of Prompting-Based Reasoning Techniques}
\label{tab:reasoning_techniques}
\end{table}

\subsubsection{Iterative Reasoning Paradigms in LLMs}
This subsection examines iterative reasoning approaches in LLMs, which improve problem-solving through repeated cycles of reasoning, action, and refinement. These methods address challenges in complex tasks, such as mathematical reasoning, decision-making, and long-term planning, by breaking them into manageable steps. Key approaches include: ReAct, which combines reasoning and actions in a loop with environmental feedback; self-improving techniques like Self-Refine and Self-Debug, where models critique and enhance their own outputs; memory-based systems like Reflexion and Cumulative Reasoning, which store and reuse past knowledge; planning frameworks (e.g., LLM+P and Plan-and-Solve) which structure tasks into sequential steps; and selection-inference models or tool-using methods (e.g., PAL, ToolLLM), which leverage external computation or APIs for accuracy. Together, these paradigms enhance LLMs' ability to generalize, reduce errors, and produce reliable, interpretable results across diverse applications.

\textbf{Iterative Reasoning with Task Decomposition}

Building on the foundation of reasoning techniques, iterative reasoning with task decomposition enhances complex problem-solving in LLMs by breaking a query into manageable sub-tasks and progressively refining intermediate results. Zhou et al.~\cite{zhou2023leasttomostpromptingenablescomplex} introduced least-to-most prompting to address limitations in CoT prompting for complex problem generalization. This two-stage framework includes::
\begin{itemize}
    \item \textit{Decompose}: decomposes a query into sub-queries using (a). few-shot examples of query decomposition and (b). the current query.
    \item \textit{Solve}: solves sub-queries sequentially with a prompt containing (a). constant examples of sub-problem solutions, (b). previously answered sub-questions and their solutions and (c). the next sub-question to address.
\end{itemize}
This progressive approach enables LLMs to tackle tasks requiring symbolic manipulation, compositional generalization, and mathematical reasoning. Wang et al.~\cite{wang2022iterativelypromptpretrainedlanguage} developed an iterative prompting framework to enhance multi-step reasoning in complex tasks. A context-aware prompter, trained while keeping the LLM fixed, generates a new sub-query based on the query and previously retrieved knowledge. The LLM then uses this sub-query to produce a sub-answer; the loop continues until a stopping condition is met. This method improves performance on tasks requiring iterative refinement of reasoning steps.

\begin{figure}
    \centering
    \includegraphics[width=\linewidth]{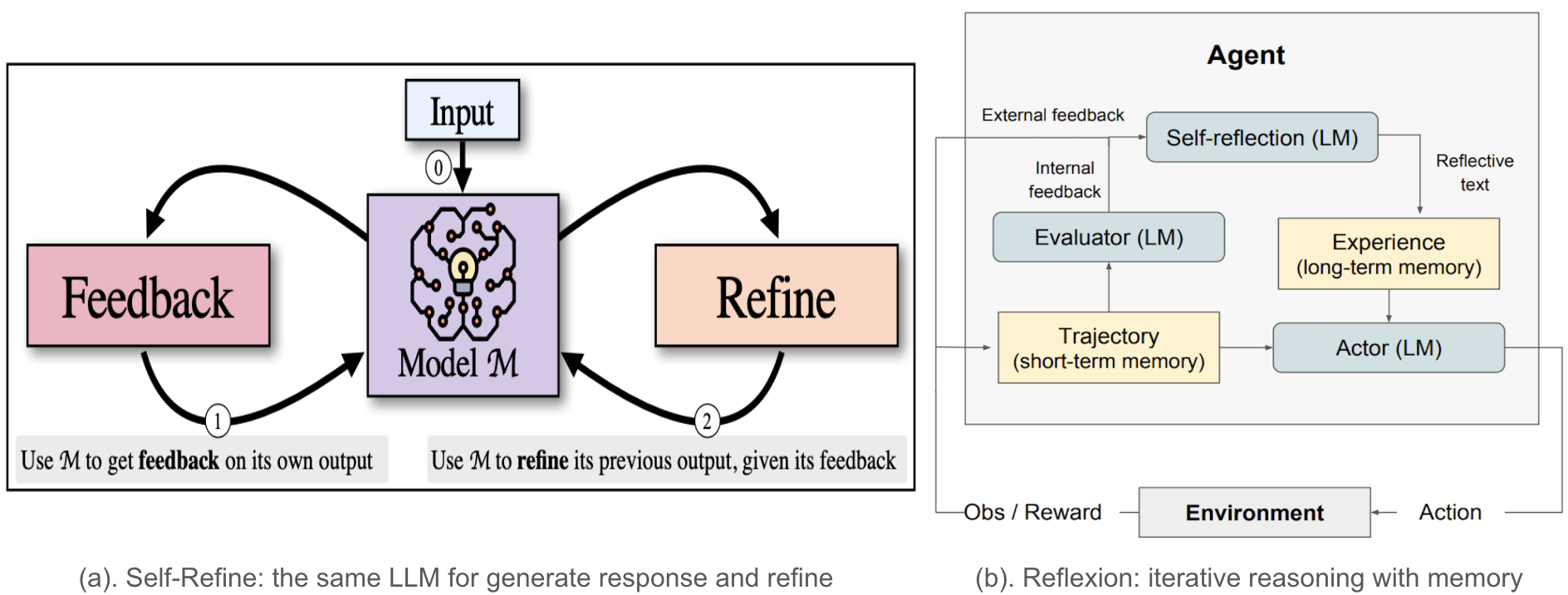}
    \caption{(a). Self-Refine: the same LLM is utilized for generating the response and providing the feedback, (b). Reflexion: iterative reasoning with memory}
    \label{fig: SelfRefine_Reflexion}
\end{figure}

\textbf{Iterative Reasoning with Feedback}

Extending beyond decomposition, iterative reasoning with feedback introduces an adaptive loop where LLMs continually refine their outputs through self-assessment or environmental responses. As proposed by Yao et al.~\cite{yao2023reactsynergizingreasoningacting}, \textsc{ReAct} (Reason+Act) is a paradigm that integrates reasoning and action in large language models (LLMs) through a reasoning-action-observation loop. The process involves three steps:
\begin{enumerate}
    \item \textit{Reason}: The LLM generates free-form thoughts, such as decomposing tasks, creating or updating plans, injecting commonsense knowledge, or handling uncertainty.
    \item \textit{Act}: The LLM interacts with an external environment, such as querying Wikipedia or navigating text-based systems.
    \item \textit{Observe}: Feedback from the environment informs subsequent reasoning and actions.
\end{enumerate}
This cycle continues until a final answer is reached. ReAct, a prompt-based approach, leverages few-shot in-context examples and outperforms reasoning-only (Chain-of-Thought, CoT) and action-only baselines on various tasks. By producing grounded, fact-driven reasoning traces, ReAct enhances human interpretability and reduces hallucinations. Optionally, ReAct can be combined with CoT-SC to effectively integrate internal knowledge and external feedback. In \cite{madaan2023selfrefineiterativerefinementselffeedback}, Self-Refine was proposed to enhance the capabilities of an LLM, as shown in Figure~\ref{fig: SelfRefine_Reflexion}(a). Initially, the LLM is guided by a prompt to generate an initial response based on a user query. The same LLM then takes the input and initial response to generate feedback. Finally, the response is refined based on the feedback, starting another iteration. The process stops when the maximum number of iterations is reached or a stop token is output. In Self-Debug \cite{chen2023teachinglargelanguagemodels}, inspired by rubber duck debugging, the authors propose a method for teaching LLMs to iteratively debug their own code without human intervention or additional model training. Self-Debug involves three main steps: 1. generate candidate code, 2. code execution, 3. explain the code line-by-line in natural language, and 4. produce a feedback message based on the code explanation and execution results for debugging. In \textsc{Refiner}~\cite{paul2024refinerreasoningfeedbackintermediate}, the authors propose an interaction-based framework where a generator language model produces intermediate reasoning steps, and a critic model provides fine-grained structured feedback on errors. During inference, the generator produces a response, which is sent to the critic for feedback. The feedback is then incorporated into the generator to refine the response, and this iteration continues until the critic outputs ``no hint'' or the maximum iteration threshold is reached. Lastly, in Self-Correct \cite{welleck2022generatingsequenceslearningselfcorrect}, a corrector is trained using self-corrective learning. In inference, a generator generates a response and the corrector iteratively improves the response until the stopping condition. A comparison of these different iterative reasoning with feedback is in Table \ref{tab:feedback_paradigms}.

\begin{table}[h]
    \centering
    \caption{Comparison of Key Iterative Reasoning Paradigms with Feedback}
    \label{tab:feedback_paradigms}
    \begin{tabular}{c|ll}
        \hline
        \textbf{Technique} & \textbf{Feedback Source} & \textbf{Loop} \\
        \hline
        \textsc{ReAct} \cite{yao2023reactsynergizingreasoningacting} & External Environment (Observation) & Reason-Act-Observe \\
        Self-Refine \cite{madaan2023selfrefineiterativerefinementselffeedback} & Same LLM (Self-critique) & Reason-Assess-Refine \\ 
        Self-Debug \cite{chen2023teachinglargelanguagemodels} & External Execution \& LLM (Explanation) & Reason-Execute-Explain-Debug \\
        \textsc{Refiner} \cite{paul2024refinerreasoningfeedbackintermediate} & Separate Critic LLM & Reason-Assess-Refine \\        
        \hline
    \end{tabular}%
\end{table}

\textbf{Iterative Reasoning with Memory} 

While feedback-based methods focus on short-term refinement, memory-augmented reasoning extends this concept, enabling models to accumulate, recall, and leverage past experiences over time. In Reflexion \cite{shinn2023reflexionlanguageagentsverbal}, the authors propose a verbal reinforcement framework that enables language agents to learn from experience without parameter updates as shown in part (b) of Figure \ref{fig: SelfRefine_Reflexion}. The agent consists of four modules
\begin{itemize}
    \item \textit{Actor}: generates responses based on prompt and context.
    \item \textit{Evaluator}: provides external feedback from environment or internal feedback from the model.
    \item \textit{Self-Reflection Model}: transforms sparse signals (e.g., binary success/fail) into rich, semantic guidance.
    \item \textit{Memory}: stores reflections episodically.
\end{itemize}
Finally, this memory is appended to the prompt in the next iteration, allowing the agent to iteratively improve its performance. Cumulative Reasoning (CR), proposed by \cite{zhang2025cumulativereasoninglargelanguage}, enhances LLM problem-solving capability by emulating human-like cumulative reasoning. CR contains four modules: 1. Proposer for idea generation, 2. Verifier for fact checking via another LLM or external tools such as code interpreters or theorem provers and 3. Reporter for generating final responses from accumulated knowledge and 4. Memory to store verified knowledge in a direct acyclic graph (DAG) for subsequent reasoning. In \cite{brooks2023largelanguagemodelsimplement}, the authors propose In-Context Policy Iteration (ICPI), a method that uses a LLM to implement policy iteration without gradient updates or expert demonstrations. A memory buffer stores past trajectories (state, action, reward, next state), which are used to prompt the LLM to act as both a world model (predicting next states and rewards) and a rollout policy (choosing actions). At each step, the LLM simulates rollouts based on sampled trajectories from the buffer to compute Q-values. Actions are selected greedily with respect to these Q-values, and new experiences are added to the buffer. This iterative process continues until convergence, improving the policy solely through in-context learning. 

\textbf{Iterative Reasoning with Plan}

In parallel, planning-based iterative reasoning introduces structured decision-making, where LLMs devise and refine plans before executing them to handle long-horizon or multi-step tasks. LLM+P \cite{liu2023llmpempoweringlargelanguage} is a framework that enables large language models (LLMs) to solve long-horizon planning problems by leveraging classical planners. The system translates a natural language description of a planning problem into PDDL using an LLM (with in-context learning), solves it with a classical planner, and then converts the solution back into natural language. In Plan-and-Solve prompting \cite{wang2023planandsolvepromptingimprovingzeroshot}, the authors improve CoT by addressing three common error types: 1. calculation errors, 2. missing-step errors and 3. semantic misunderstanding errors by introducing two instruction prompts: 1. PS Prompting and 2. PS+ Prompting. PS Prompting first instructs the LLM to devise a plan and then execute it step by step: "Let’s first understand the problem and devise a plan to solve the problem. Then, let’s carry out the plan and solve the problem step by step." PS+ Prompting adds detailed guidance based on PS Prompting: “extract relevant variables and their corresponding numerals” and “calculate intermediate results (pay attention to calculation and commonsense)”. AdaPlanner \cite{sun2023adaplanneradaptiveplanningfeedback} first utilizes an LLM to generate an initial plan based on task description, candidate actions and optional expert demonstrations to generate an intial plan with Python-style function to 1. decompose tasks, 2. generate actions, 3. check results and 4. parse information. During execution, AdaPlanner adapts the plan using two refinement strategies: in-plan refinement, which extracts useful information from environmental feedback for upcoming actions without modifying the current plan, and out-of-plan refinement, which revises the plan when unexpected errors occur (refine-then-resume). Lastly, the system also uses skill discovery to store successful plans, improving sample efficiency. 

The following works focus on iterative reasoning with plan on embodied tasks. In \cite{huang2022innermonologueembodiedreasoning}, the authors propose Inner Monologue, a framework where a pretrained LLM acts as an interactive planner for embodied robotic tasks. The LLM iteratively plans actions, executes them via pre-trained robotic skills, receives textual feedback from the environment including success detection, passive, and active scene descriptions, and human feedback, and replans as needed. This closed-loop reasoning allows the LLM to adapt to failures, gather information, and accomplish complex long-horizon tasks without additional training. Text2Motion \cite{Lin_2023} combines LLMs with motion planning for robots. The LLM translates a language instruction into possible goal states, and then a planner searches for a sequence of skills to reach one of these goals. The planner uses two approaches: generating full candidate plans and checking if they are feasible, or building a plan step by step while checking each action. By performing verification on both the language and motion aspects, Text2Motion achieves much higher success rates on long-horizon tasks than earlier LLM-based methods. In \cite{wang2024describeexplainplanselect}, the authors propose DEPS (Describe, Explain, Plan, Select), an interactive LLM-based planning framework for multi-task embodied agents in open-world environments. DEPS addresses two challenges of open-world planning: (1) long-horizon tasks with strict sub-goal dependencies and (2) state-dependent feasibility of alternative sub-goals. The framework works by (a) generating an initial plan via an LLM, (b) using a descriptor to summarize execution outcomes, (c) prompting the LLM as an explainer to diagnose failures, and (d) replanning accordingly. To further improve efficiency, DEPS introduces a learned goal selector that ranks parallel candidate sub-goals based on predicted completion horizon.

\textbf{Iterative Reasoning with Selection}

Complementing planning approaches, selection-based reasoning focuses on identifying the most relevant information or hypotheses across multiple inference cycles. In \cite{creswell2022selectioninferenceexploitinglargelanguage}, selective inference (SI), a two-stage framework, was proposed to enhance LLM's capability on multiple-steps reasoning. In the first stage, as a selection module, LLM with few-shot examples of good selection, scores the probability of facts and the facts with the highest scores will be chosen. In the second inference stage, an LLM, provided with few-shot examples, is prompted to generate conclusion from the selected facts. The new generated conclusion is added to the knowledge base for a new round of selection, and this cycle repeats until reaching the final answer. In \cite{creswell2022faithfulreasoningusinglarge}, the authors propose a faithful multi-step reasoning framework built on a Selection–Inference (SI) backbone with 1. the Halter and 2. the Value Function. The Halter is a LLM to decide when to stop the cycle through answering "Given {inference}, can you answer {question}?". If yes, the answer will be generated from the last inference. Otherwise, an answer of "Unknown" will be returned. The Value Function predicts a score for the (selection, inference) pair and prunes the ones with lower score given multiple generations.

\begin{figure}
    \centering
    \includegraphics[width=\linewidth]{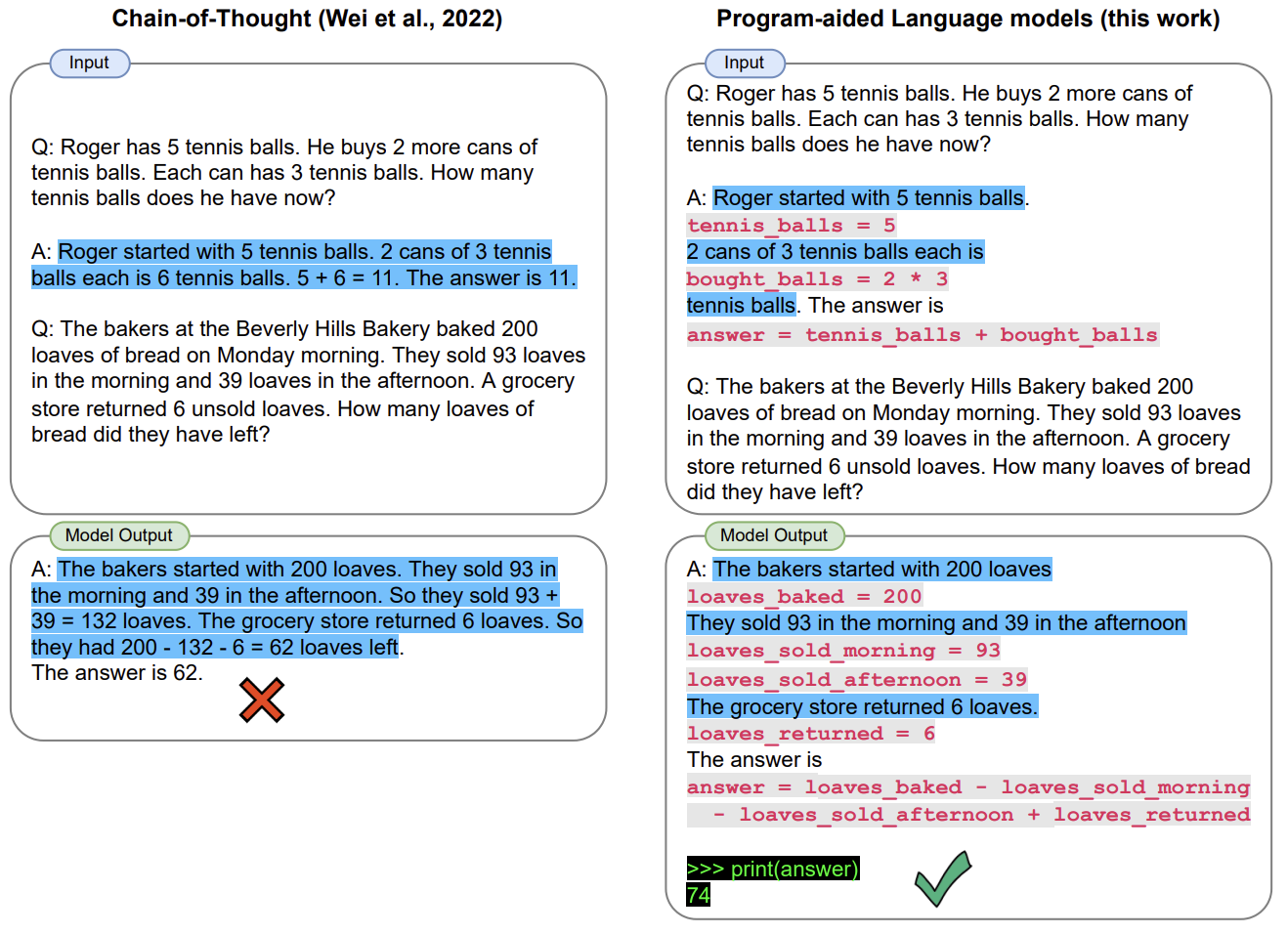}
    \caption{Program aided language model transforms the natural language problem into Python programs so that the results are obtained from the execution of the Python program.}
    \label{fig: PAL}
\end{figure}

\textbf{Iterative Reasoning with Tool}

Finally, tool-augmented reasoning extends LLM capabilities beyond text, allowing models to interact with external computation engines, APIs, or symbolic systems for more accurate and interpretable results. In program-aided language model (PAL) \cite{gao2023palprogramaidedlanguagemodels}, the LLM is used to translate natural language problems into Python programs that define variables and operations as intermediate reasoning steps to avoid calculation errors during CoT, as shown in Figure \ref{fig: PAL}. Then, the Python interpreter executes the code to find the final answer. In Program of Thoughts (PoT) prompting \cite{chen2023programthoughtspromptingdisentangling}, LLMs are prompted to generate reasoning steps in the form of Python programs, which are then executed by an interpreter. In Faithful CoT \cite{lyu2023faithfulchainofthoughtreasoning}, a query was sent to LLM to output a reasoning chain that interleaves 1. natural language (NL) like a sub-query, dependency among sub-queries and rationale to answer each sub-query and 2. symbolic language (SL) which is the executable program to solve each sub-query. Next, a deterministic external solver will execute the code in SL to generate the final answer. In \cite{Drori_2022}, the authors show that OpenAI’s Codex can solve, explain, and generate university-level math problems at scale. Their method takes a math question, synthesizes a Python program via zero-shot or few-shot prompting, executes it to obtain the solution, and then produces step-by-step explanations. In Hypothesis search \cite{wang2024hypothesissearchinductivereasoning}, given a prompt and multiple samples of input-output pairs, the authors proposed to utilize LLM to generate, filter, and test multiple hypotheses. For each remaining hypotheses, LLM generates a Python code to realize the hypothesis and tests it on the prepared examples, where programs that pass all tests are selected and the ones that incurred errors will be sent back to LLM for debugging. Eventually, the best program is selected. In \cite{surís2023vipergptvisualinferencepython}, ViperGPT handles complex visual queries by using a code-generation LLM that decomposes the query into interpretable steps, generates executable Python code to call pretrained vision and language modules, and sequentially executes each step to produce the final answer.

Another direction is calling API tools to solve different problems. In ToolLLM \cite{qin2023toolllmfacilitatinglargelanguage}, the initial LLM was fine-tuned on ToolBench, a dataset of 16k APIs over 49 categories and muti-step tool reasoning. Eventually, given a query from the user, relevant APIs are retrieved from the pool through a pretrained retriever and the ToolLLM can build the API call with suitable parameters and derive the final answer through multi-tool and multi-step reasoning. TORA \cite{gou2024toratoolintegratedreasoningagent} introduces Tool-integrated Reasoning Agents for mathematical problem-solving via an iterative loop: 1. natural language rationale, 2. program/tool call, 3. tool output and 4. refined rationale, blending analytical reasoning with computational tools like SymPy solvers. In ToolChain* \cite{zhuang2023toolchainefficientactionspace}, the A* algorithm is employed to search through a decision tree of possible actions. The root node corresponds to the user query and each node is a possible tool, i.e., API function call. At each step, the algorithm evaluates candidate leaf nodes by computing their total cost as the sum of the cumulative past cost and a heuristic estimate of future cost. The node with the lowest total cost is then expanded, and its past cost functions are updated accordingly. To improve reliability, long-term memory and self-consistency mechanisms are incorporated to penalize erroneous branches. This iterative process continues until the termination condition is satisfied.

\subsection{Search}

This section provides an overview of search strategies used to guide LLMs in reasoning and sequence generation. We cover tree-based methods such as MCTS and value-function-guided search, which systematically explore multiple reasoning paths. Sequential decoding strategies, including beam search and stochastic beam search, are discussed for generating high-quality and diverse token sequences. We also examine best-first that adaptively allocate computational resources to promising reasoning paths. Finally, multi-agent LLM frameworks are introduced, highlighting how iterative debate and collaboration among multiple models can improve reasoning accuracy and robustness.

\begin{figure}
    \centering
    \includegraphics[width=\linewidth]{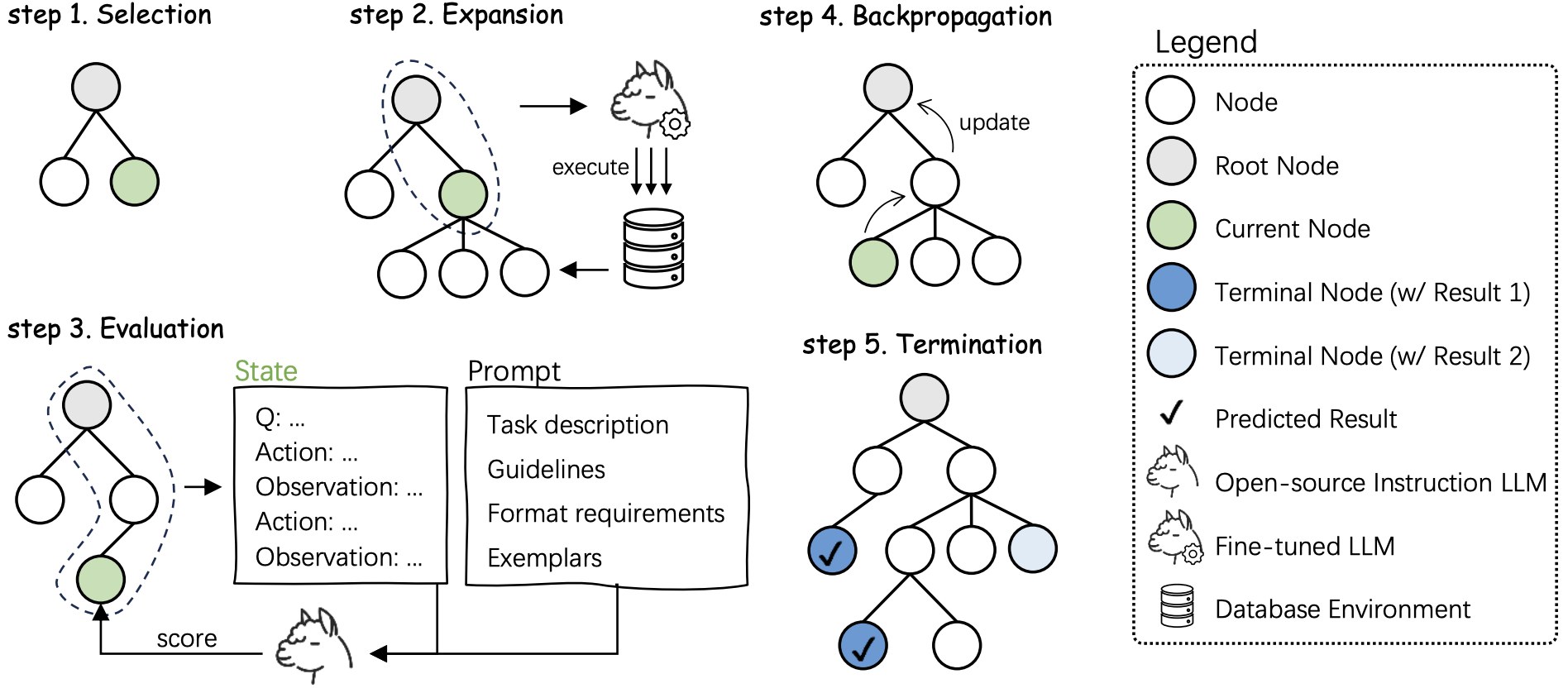}
    \caption{MCTS: Monte Carlo Tree Serarch includes four stages 1. selection, 2. expansion, 3. evaluation, 4. backpropagation and 5. termination}
    \label{fig: MCTS}
\end{figure}

\subsubsection{Monte Carlo Tree Search}
Monte Carlo Tree Search (MCTS) is a heuristic search algorithm used to make optimal decisions in large, uncertain, or partially explored decision spaces---most famously in games like Go, Chess, and general reinforcement learning settings. It balances \textbf{exploration} (trying new actions) and \textbf{exploitation} (using known good actions) through repeated simulations as shown in Figure \ref{fig: MCTS}.

MCTS operates in \textbf{four main steps}, repeated many times to grow and refine a search tree:

\begin{enumerate}
    \item \textit{Selection}: Starting from the root (initial state), MCTS traverses the existing tree by selecting child nodes according to a policy that balances exploitation and exploration commonly using the Upper Confidence Bound (UCB) formula:
    \begin{equation}
    UCT = r_j + C \sqrt{\frac{\ln N}{n_j}}
    \end{equation}
    where $r_j$ is the average reward of node $j$, $N$ is the total number of visits to the parent, $n_j$ is the number of visits to node $j$, and $C$ controls the level of exploration.
    This step continues until a leaf node (a node not fully expanded or terminal) is reached.

    \item \textit{Expansion}: If the selected node is not a terminal state, one or more new child nodes (representing possible next actions) are added to the tree. This introduces unexplored actions for future simulations.

    \item \textit{Simulation / Rollout}: From the newly expanded node, the algorithm performs a random or heuristic-based simulation until a terminal state (or a predefined depth) is reached. The outcome of this simulation (e.g., win/loss, score, reward) provides an estimate of the node’s value.

    \item \textit{Backpropagation}: The simulation result is then propagated back up the tree. Each node along the path from the expanded node to the root updates its statistics (e.g., visit count and average reward). This gradually improves the estimated value of each action and helps guide future selections.
\end{enumerate}

After many iterations, MCTS converges toward optimal decisions: the action corresponding to the child of the root with the highest estimated value is chosen. In short, MCTS efficiently builds a partial search tree guided by stochastic sampling, making it powerful for problems with large or complex state spaces where exhaustive search is infeasible.

\textbf{MCTS for LLM reasoning} 

In MCTS for LLMs, step-by-step reasoning is modeled as a Markov decision process (MDP), where states are intermediate reasoning steps and actions are next-step generations from LLMs. In reasoning as planning (RAP) \cite{hao2023reasoninglanguagemodelplanning}, a LLM is utilized as both a reasoning agent to predict the next action and a world model to predict the next state. Next, planning is performed using MCTS guided by rewards including: 1. likelihood of action, 2. confidence of state, 3. self-evaluation to check the correctness of the reasoning step, and 4. task-specific heuristics, i.e., custom reward signals based on the nature of the task to encourage goal-directed reasoning. Eventually, multiple reasoning paths are generated and the final answer is derived by aggregating them. Building upon this framework, LE-MCTS \cite{park2024ensemblinglargelanguagemodels} ensembles multiple LLMs to improve reasoning robustness. In each iteration, a leaf node is selected by balancing exploration and exploitation, expanded by generating candidate next steps from multiple LLMs, and evaluated using a preference reward model (PRM). The step with the highest reward is then propagated upward through optimistic backpropagation, which prioritizes the best child rather than averaging over all children. This iterative process continues until a predefined depth or computational budget is reached.

A different line of work explores value-function-guided MCTS, where the reward evaluation relies on learned value estimators. In \cite{fu2023acceleratingmontecarlotree}, probability tree state abstraction (PTSA) was applied to speed up MCTS by reducing the size of the search tree by grouping or clustering nodes with similar states as a single node. When a new path comes in, it is compared with the saved paths using PTAS function through Jensen-Shannon (JS) divergence between the probability distributions of their predicted action values (Q-values). If the new path is merged into the old path list, the path with the lower estimated value will be removed. Eventually, PTSA can reduce the computational cost by 10\%-45\% without deteriorating the performance of MCTS. In PPO-MCTS \cite{liu2024dontthrowawayvalue}, a decoding algorithm leverages the value model from PPO to guide token-level MCTS. Specifically, the Q-value of each expanded node is initialized using the parent node’s V-value to encourage exploration. After running multiple simulations, the probability of selecting a token is made proportional to its visit count, which reflects higher expected rewards. In \cite{feng2024alphazeroliketreesearchguidelarge}, a shared LLM-based value function and outcome reward model (ORM) are trained. Then, tree search optimizes cumulative rewards by one of five algorithms—BFS-V/DFS-V, classic MCTS (Monte Carlo backpropagation from terminals), MCTS-$\alpha$ (AlphaZero-like with value approximations and visit-count selection), or MCTS-Rollout (hybrid restarting from root with intermediate backups)—to explore paths. Lastly, different reasoning paths are aggregated via majority vote, ORM-max, or ORM-vote to select high-reward outputs.

While the above methods rely on structured search guided by external or model-based reward signals, subsequent works introduced self-refinement and human-like evaluation mechanisms into MCTS. Building upon this structured reasoning paradigm, Monte Carlo Tree Self-refine (MCTSr) built a reasoning algorithm that augments LLMs with a structured search-and-refinement process \cite{zhang2024accessinggpt4levelmathematical}. MCTSr includes five stages: 1. selection, 2. expansion, 3. self-refine, 4. self-evaluation and 5. backpropagation. In self-refine, candidate solutions are improved through iterative refinement. In self-evaluation, candidates were scored with self-reward mechanism using 1. prompt constraint 2. full score suppression and 3. repeated sampling. This cycle iterates until termination and yields the final reasoning paths. Complementary to these self-refining approaches, Cooperative Reasoning (CoRe) \cite{Zhu_2023} proposed a human-like dual-system approach including: 1. a generator to generate potential reasoning paths and 2. a verifier to score the individual step and the whole response for solving math word problems. In inference, CoRe utilized MCTS to balance exploration and exploitation through the verifier's score on the generator's response. 

Finally, while the above approaches treat each response or reasoning step as a single unit of action, PPL-MCTS \cite{chaffin2022pplmctsconstrainedtextualgeneration} shifts the focus to token-level search and evaluation. This method enables constrained text generation without fine-tuning by applying MCTS at the token-level, where leaf nodes are selected using the PUCT formula, allowing fine-grained control over the generated content.

\textbf{Tree-based Search}

Unlike linear reasoning, tree search–based approaches—distinct from MCTS—expand into a branching structure that examines multiple reasoning paths and systematically assesses alternative explanations and outcomes. In Maieutic Prompting \cite{jung2022maieuticpromptinglogicallyconsistent}, the authors proposed "logical integrity": a proposition is integral if the LM gives opposite truth values to Q and !Q consistently as shown in part (a) of Figure \ref{fig:Tree_Beam_Search}. For example, in Proposition Q: “The sky is blue.” and Negation !Q: “The sky is not blue.”, the LLM demonstrates logical integrity if it assigns 1. True for Q and False for !Q or 2. False for Q and True for !Q. Instead of directly predicting True/False or relying on a single explanation, the method builds a maieutic tree: for each statement, the LM generates abductive explanations for both "True" and "False", then recursively expands these explanations into deeper ones. At each step, explanations are checked for logical integrity, i.e., whether the LM gives consistent opposite labels to a statement and its negation. Non-integral branches are pruned and the resulting tree contains only logically integral propositions, which are scored by the LM’s belief in each proposition and its consistency with the original question. Eventually, these scores are formulated as weighted constraints and the MAX-SAT solver assigns truth values across the tree to maximize overall consistency. In this way, the workflow transforms noisy, inconsistent generations into a structured reasoning process that yields more reliable truth judgments.

\subsubsection{Beam Search}
Unlike greedy search, which selects the single most probable token at each step, beam search keeps track of the top k tokens (called the beam width) at each generation step. By exploring multiple candidate sequences simultaneously, it balances exploration and exploitation, increasing the likelihood of finding higher-probability overall sequences.

\textbf{Beam Search} 

The following works introduced minor modifications to beam search to enhance its performance. In \cite{massarelli2020decodingstrategiesaffectverifiability}, the authors discovered that sampling based method produced less repetitive, more creative, and more human-like text, while likelihood strategies like beam search produced more factual and verifiable text. To make a balance between sampling and likelihood strategies, the authors proposed delayed beam search, and it started the first L tokens using sampling based method and the remaining tokens through beam search. In \cite{wang2024chainofthoughtreasoningprompting}, the authors proposed to force the LLM to decode the top-k tokens for the first token and then decode greedily for the remaining tokens to generate different responses and select the best one for answering. To begin with, the authors discovered that if the LLM’s generated responses utilized CoT, the top-1 candidate will have a much larger probability over the top-2 candidate for each token. However, this is reversed if the LLM response did not utilize CoT. By forcing the model try all different top-k candidates at the first token can increase the probability of CoT. In addition, the author proposed to utilized the confidence score:
\begin{equation}
\delta = \frac{1}{|answer|} \sum_{i=1}^{|answer|} (P(y_i^1|x, y_1, \cdots, y_{i-1}) - P(y_i^2|x, y_1, \cdots, y_{i-1}))
\end{equation}
where $P(y_i^1|x, y_1, \cdots, y_{i-1})$ referred to the top-1 candidate and $P(y_i^2|x, y_1, \cdots, y_{i-1})$ referred to the top-2 candidate for each token and the confidence score will be larger if the model is more confident. Eventually, the answer with the highest confidence score will be selected. 

Beyond improving token-level probability, recent work has focused on integrating external evaluators to guide beam search toward correct reasoning paths. In \cite{li2023makinglargelanguagemodels}, the authors propose DIVERSE (Diverse Verifier on Reasoning Step), a method to improve language model reasoning on GSM8K. The approach has three components: (1) generating diverse prompts to elicit multiple reasoning paths, (2) using a verifier to score each path and perform weighted voting instead of naive majority voting, and (3) introducing a step-aware verifier that labels and evaluates each reasoning step individually, allowing the model to identify partially correct reasoning paths and filter out faulty steps. In \cite{zhu2024deductivebeamsearchdecoding}, the authors introduce Deductive Beam Search (DBS), which integrates beam search with a pretrained deductive verifier to check factual consistency. This ensures that each step in the reasoning chain is logically deducible from the previous one, thereby mitigating error accumulation.

Verifier-guided beam search has also been extended to physically grounded planning tasks, where language models must generate sequences that satisfy real-world constraints, such as robot actions and safety considerations. In SayCan \cite{ahn2022icanisay}, LLM was combined with a robot's affordance function to ground LLM's plan with reality. In "Say", given the current state in physical world, available skills of the robot and the final goal, the LLM predicted the probability of each action. In the meantime, the robot estimated the possibility of each candidate action through the affordance function, which was referred as "Can". Eventually, the probability of "SayCan" was multiplication of the probabilities of "Say" and "Can", and the highest scored action was selected. The robot executed this action and moved to the next state for the next action selection until the goal is achieved. In \cite{hazra2024saycanpayheuristicplanninglarge}, SayCanPay framed LLM planning as a heuristic search problem where Say and Can are the same as before while Pay referred to Long-Term Value Estimation to avoid short-visoned actions. Eventually, the final score was the multiplication of probabilities of "Say", "Can" and "Pay" and it was utilized in the beam search for generating the best possible responses. In \cite{huang2023groundeddecodingguidingtext}, the authors propose Grounded Decoding, a method that guides large language models with grounding functions to ensure generated plans are both semantically meaningful and physically realizable for robots. At each decoding step, token probabilities from the LLM are combined with probabilities from grounded models (e.g., affordances, safety constraints, preferences), effectively framing the problem as probabilistic filtering. In Guiding chain-of-thought ReAsoning with a CorrectnEss Discriminator (GRACE) \cite{khalifa2023gracediscriminatorguidedchainofthoughtreasoning}, a stepwise approach was proposed to solve the problem that LLMs often assign high probability to incorrect reasoning steps. To begin with, a correctness discriminator was trained through 1. sampling steps, 2. finding the exact mistakes like missing or incorrect steps and 3. discriminator training with the contrastive max-margin loss. After training, a stepwise decoding strategy was applied with the score of each step as the multiplication of 1. LM likelihood and 2. discriminator score for the candidate step.

\textbf{Stochastic Beam Search}

\begin{figure}
    \centering
    \includegraphics[width=\linewidth]{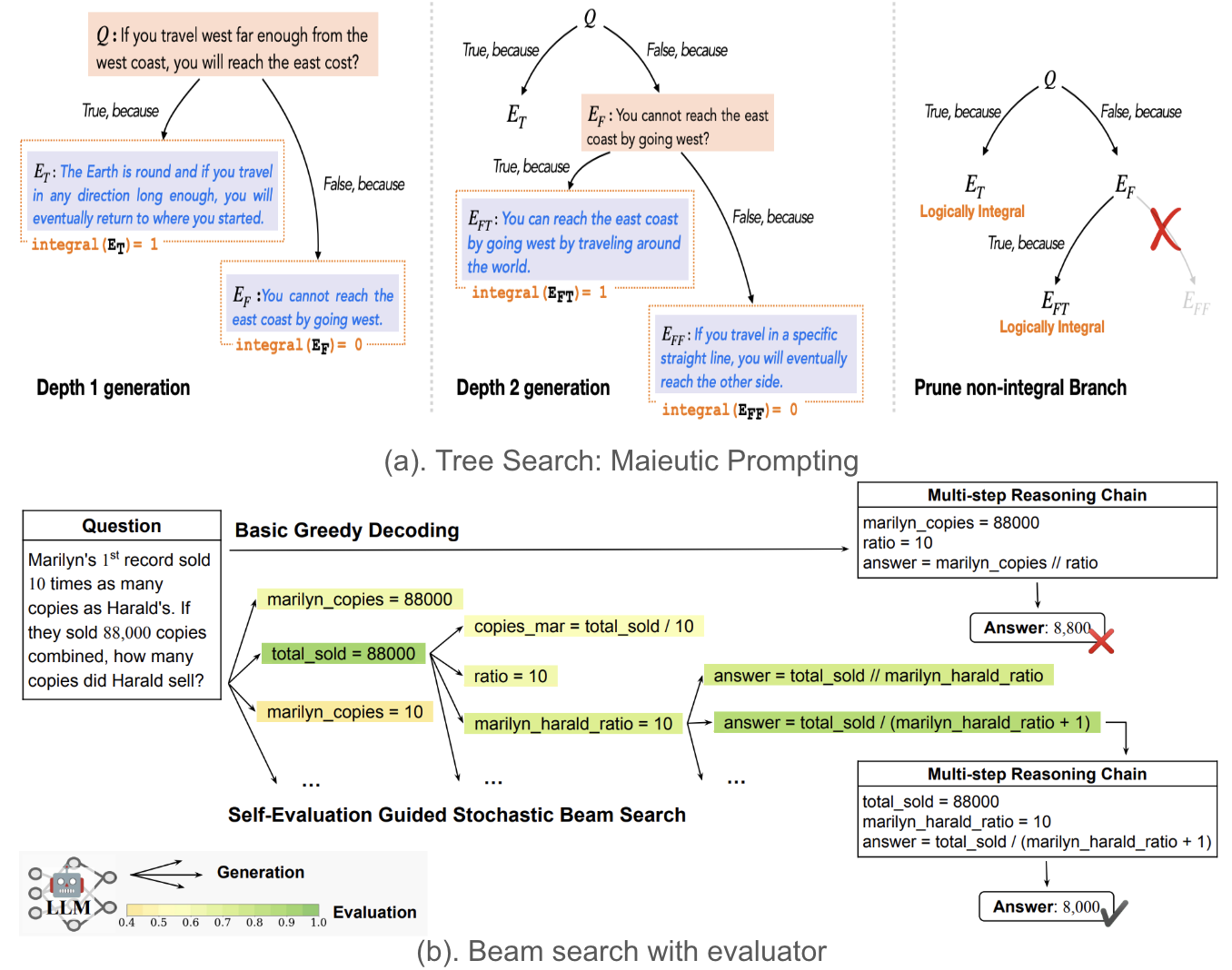}
    \caption{(a). Tree search through Maieutic Prompting to ensure integrity, (b). Beam search with evaluator}
    \label{fig:Tree_Beam_Search}
\end{figure}

Another line of research addresses the lack of diversity in conventional beam search by introducing stochasticity into the decoding process, enabling the generation of more varied yet high-quality sequences. In \cite{kool2019stochasticbeamsthemgumbeltopk}, the authors introduce stochastic beam search (SBS), an algorithm that implicitly applies the Gumbel-Top-k trick to sequence models without enumerating the exponentially large sequence space. The Gumbel-Max trick samples one element from a categorical distribution, adds independent Gumbel noise to each element's log-probability and takes the argmax. Here, replacement refers to whether a sampled element is put back into the pool before the next sampling. Gumbel-Top-k samples the top-k elements rather the argmax. In stochastic beam search, for each step, the log probability based on its parental sequence was predicted, sampled with a Gumbel perturbation and finally pruned to maintain the top-k subsequences. Empirically, SBS yields more diverse yet high-quality translations. In \cite{xie2023selfevaluationguidedbeamsearch}, the LLM played two roles: 1. next step generation and 2. next step generation’s self evaluation as shown in part (b) of Figure \ref{fig:Tree_Beam_Search}. In each step, multiple next steps were generated and each candidate was evaluated by the combination of 1. LLM’s probability of generating the response and 2. correctness confidence through self-evaluation. Then, a constrained SBS is utilized to sample answers by utilizing the weighted score rather than greedily selecting the top-k next paths. The procedure proceeds step by step until the full reasoning chain is constructed for producing the final answer. In \cite{meister2023conditionalpoissonstochasticbeam}, Conditional Poisson Stochastic Beam Search (CPSBS) was utilized to address two limitations of standard beam search: high candidate overlap and biased expectation estimates. Rather than selecting the top-k using argmax, it selects the next tokens through Conditional Poisson (CP) sampling. The CP sampling procedure begins with an empty set. At each iteration, every token in the vocabulary is considered for inclusion. For each token, the probability of being added to the set is determined by its weight, the number of remaining slots, and a precomputed normalization factor. The process continues until the set reaches its predefined size.

\subsubsection{Best-first Search}

To further enhance reasoning without retraining, best-first search methods scale inference-time computation by exploring reasoning paths adaptively—allocating more compute to the most promising trajectories rather than treating all reasoning chains equally. In \cite{polu2020generativelanguagemodelingautomated}, GPT-f was utilized to prove mathematical theorems by starting with a goal and generates multiple proof steps, and these proof steps are first verified by a theorem verifier, i.e., Metamath, and then search the next steps by either 1. cumulative log probability or 2. value function. Instead of merely searching for the next steps sequentially, the system expands a priority queue of sub-goals, repeatedly applying candidate proof steps until the entire proof tree is closed. In Q* \cite{wang2024qimprovingmultistepreasoning}, a Q-value model was trained via offline RL, rollouts, or stronger LLM completions which predicts future potential of a given immediate reasoning step, where the optimal Q value represents the total expected future reward from taking action $a_t$ in state $s_t$ and then acting optimally thereafter, where states refer the partial reasoning traces and actions are next reasoning steps. At inference, candidate next steps are scored by (1) aggregated utility (process-based rewards collected so far) and (2) heuristic value (the estimated optimal Q-value of future steps). Using an A*-style best-first search, Q* guides LLMs toward the most promising reasoning trajectories. 

\subsubsection{Multi-Agent LLMs}
In \cite{du2023improvingfactualityreasoninglanguage}, the authors propose a multi-agent debate framework where multiple instances of LLM independently generate answers to a query, critique each other’s reasoning, and iteratively update their responses across several rounds until reaching a consensus. Results showed that debate between LLMs can correct hallucinations and inconsistencies by filtering out uncertain facts. They also discovered that "stubborn" prompting encouraged LLMs to defend their reasoning to make the debates longer. 

Building on this idea of iterative debate, subsequent works investigate how increasing the number of cooperating LLM agents can systematically improve performance. In AgentForest \cite{li2024agentsneed}, the authors show that LLM performance systematically scales with the number of agents via a simple sampling-and-voting approach. For close-ended tasks, majority voting is used, while for open-ended tasks, the answer with the highest BLEU score is selected. Experiments across reasoning, generation, and coding tasks reveal that this brute-force scaling can match or exceed larger models, yields greater gains on harder tasks and weaker models, and can further boost existing methods like CoT and Debate. In multi-agent LLM training (MALT) \cite{motwani2025maltimprovingreasoningmultiagent}, the authors propose a multi-agent post-training strategy with three specialized roles: a generator, a verifier, and a refiner. Training data is automatically produced by expanding a multi-agent search tree, where outcome rewards are propagated back to each agent using value iteration. Next, the generator is fine-tuned through SFT and the verifier and refiner are fine-tuned through SFT and DPO. At inference, the generator produces candidate answers, the verifier critiques them, and the refiner integrates the critiques to generate improved responses, with majority voting over multiple iterations yielding the final answer. Beyond general reasoning, multi-agent frameworks have also been applied to structured tasks such as code generation, where specialized agent roles coordinate to produce correct outputs. In AgentCoder \cite{huang2024agentcodermultiagentbasedcodegeneration}, a multiple-agent framework consisting of 1. programmer agent for generating the code and refining the code based on error, 2. test designer agent for independently generating diverse and unbiased test cases, 3. test executer agent that runs codes in Python. If all tests are passed, the final answer is derived. Otherwise, the error messages are sent to the programmer agent for new code generation. 

\subsection{Decoding}

The most frequently used decoding strategies in generation are broadly categorized into deterministic and probabilistic approaches. Deterministic methods like greedy decoding select the most probable tokens, ensuring high quality and reproducibility but often yielding less diverse text \cite{sutskever2014sequence}. In contrast, sampling-based methods such as temperature sampling \cite{ackley1985learning}, top-k sampling and nucleus (top-p) sampling \cite{fan2018hierarchicalneuralstorygeneration, holtzman2020curiouscaseneuraltext} introduce stochasticity to enhance diversity and naturalness. In this part, more methods of decoding for inference-time computation scaling are discussed.

\subsubsection{Best-of-N}

Building on the idea of generating multiple candidate outputs to improve correctness and confidence, Best-of-N decoding selects the most promising solution from a set of candidates, effectively balancing exploration and reliability during inference. In \cite{cobbe2021trainingverifierssolvemath}, a generator–verifier framework was proposed where the generator produces multiple candidate solutions, and the verifier—trained to predict correctness of (problem, solution) pairs—selected the best one. At inference time, multiple solutions are generated and the highest-scoring one is selected for the final answer. In V-STaR \cite{hosseini2024vstartrainingverifiersselftaught}, the generator is iteratively fine-tuned on correct solutions, while a verifier is trained with Direct Preference Optimization (DPO) on both correct and incorrect solutions. At inference, the verifier ranks multiple candidate generations to select best-of-N. Another complementary approach focuses on task-specific evaluation through functional correctness rather than learned verifiers. In CODET \cite{chen2022codetcodegenerationgenerated}, the LLM is prompted to generate multiple answers and multiple test cases to evaluate these answers, where all the answers were tested over all test cases. Based on their results, the answers were clustered into different sets, where the score of the set is computed based on: 1. the number of answers in the cluster and 2. the number of cases that they pass. Eventually, the answer with the highest score was selected. 

Beyond the use of verifiers, researchers have also investigated how scaling computation at inference can improve performance. In \cite{snell2024scalingllmtesttimecompute}, the authors proved the benefits of scaling the computation budge in inference. The authors tested: (1) search with verifiers like Best-of-N, beam search, and lookahead search, and (2) iterative refinement. Easier problems benefited more from iterative refinement, while harder problems benefited more from search with verifier where the prompt hardness is defined by the pass@1. Lastly, using this adaptive strategy yields up to 4 times efficiency improvements over best-of-N baselines. In Large Language Monkeys \cite{brown2024largelanguagemonkeysscaling}, the authors explored scaling inference computation through repeated sampling to enhance LLM performance. Using positive temperature sampling, they generate multiple candidate outputs, selecting the best via automatic verifiers (e.g., unit tests, proof checkers) or heuristic methods (majority voting, reward models) for tasks without verifiers. Coverage—the likelihood of generating a correct answer—scales log-linearly with sample size, modeled as an exponential power law. 

\subsubsection{Majority Voting}

Majority voting is regarded as a special case of Best-of-N with a count-based scoring function. In \cite{lewkowycz2022solvingquantitativereasoningproblems}, the authors utilized majority voting on the fine-tuned model and observed significant improvement on performance in many benchmarks. In \cite{portillo-wightman-etal-2023-strength}, the authors discovered that LLMs have poor performance in estimating the confidence of their answers. They propose Prompt Agreement, which estimates confidence by comparing responses across multiple diverse prompts. Two approaches are introduced: 1. Multi-Prompt Log Probability utilized the majority voting to select answers and compute the average log probability and 2. Multi-Prompt Agreement (Rand Index) clustered answers into different clusters and applied Rand Index to compute the confidence on the answer. In QALIGN \cite{faria2025sampledontsearchrethinking}, an initial response is generated, and then a Markov chain Monte Carlo (MCMC) process is applied using the QUEST proposal. At each step, a random index in the sequence is chosen, the suffix is regenerated, and a Metropolis-Hastings acceptance probability—based on reward difference and model likelihood—is computed to accept or reject the new sequence. This produces a set of candidate sequences that approximate the optimal aligned distribution for the prompt. The final answer is selected via majority voting for tasks with definite answers or Minimum Bayes Risk (MBR) for open-ended tasks. 

\subsubsection{Adjusted Probability Decoding}

Beyond generating multiple candidates, another line of work focuses on adjusting token probabilities during decoding to steer the model toward more accurate or desirable outputs. In KNN-LM \cite{khandelwal2020generalizationmemorizationnearestneighbor}, a pre-inference datastore is constructed where the key is the embedding of previous tokens and the value is the next token. In inference time, the normal sampling is realized by $P(y|x)$. In contrast, datastore sampling retrieves the top 1024 nearest embeddings from the datastore, along with their corresponding next tokens. Each retrieved token is assigned a probability proportional to the negative exponential of the distance between its embedding in the datastore and the embeddings of the previous tokens in the current inference, followed by normalization to obtain $P_{\text{KNN}}(y|x)$. Eventually, the final sampling is a weighted summation of $P(y|x)$ and $P_{\text{KNN}}(y|x)$ as:
\begin{equation}
P_{\text{adjusted}}(y|x)=\alpha \times P(y|x) + (1-\alpha) \times P_{\text{KNN}}(y|x)
\end{equation}

In Proxy tuning \cite{liu2024tuninglanguagemodelsproxy}, the authors propose a lightweight decoding-time method for adapting LLMs without modifying their weights, making it applicable even to black-box LMs. Three models are required: a base large model $M$ and two smaller model, one base $m-$ and one base but fine-tuned on the specific domain $m+$. Eventually, the probability for next tokens' generation will be 
\begin{equation}
P_{\text{adjusted}}(y|x) = P_{M}(y|x) + (P_{m+}(y|x) - P_{m-}(y|x))
\end{equation}
during the inference stage.

Another representative work is Future Discriminators for Generation (FUDGE) \cite{Yang_2021}, where a binary future discriminator is trained to guide the model toward outputs with specific attributes. To begin with, the binary attribute discriminator was trained to predict $P(a|x_{1:i})$, i.e., the probability that the eventually completed sequence $x_{1:n}$ will have the desired attribute $a$. Then, FUDGE multiplies the generator’s original probabilities with the classifier’s predictions and renormalizes to sample tokens, corresponding to a Bayesian factorization 
\begin{equation}
P(x_i \mid x_{1:i-1}, a) \propto P(a \mid x_{1:i}) \cdot P(x_i \mid x_{1:i-1})
\end{equation}
Eventually, the updated probability distribution was utilized to sample the response. One more benefit of FUDGE lay in its ability to extend to various attributes, where the log probabilities from different classifiers are summed before adjusting the $P(x_i \mid x_{1:i-1}, a)$. 

\begin{figure}
    \centering
    \includegraphics[width=\linewidth]{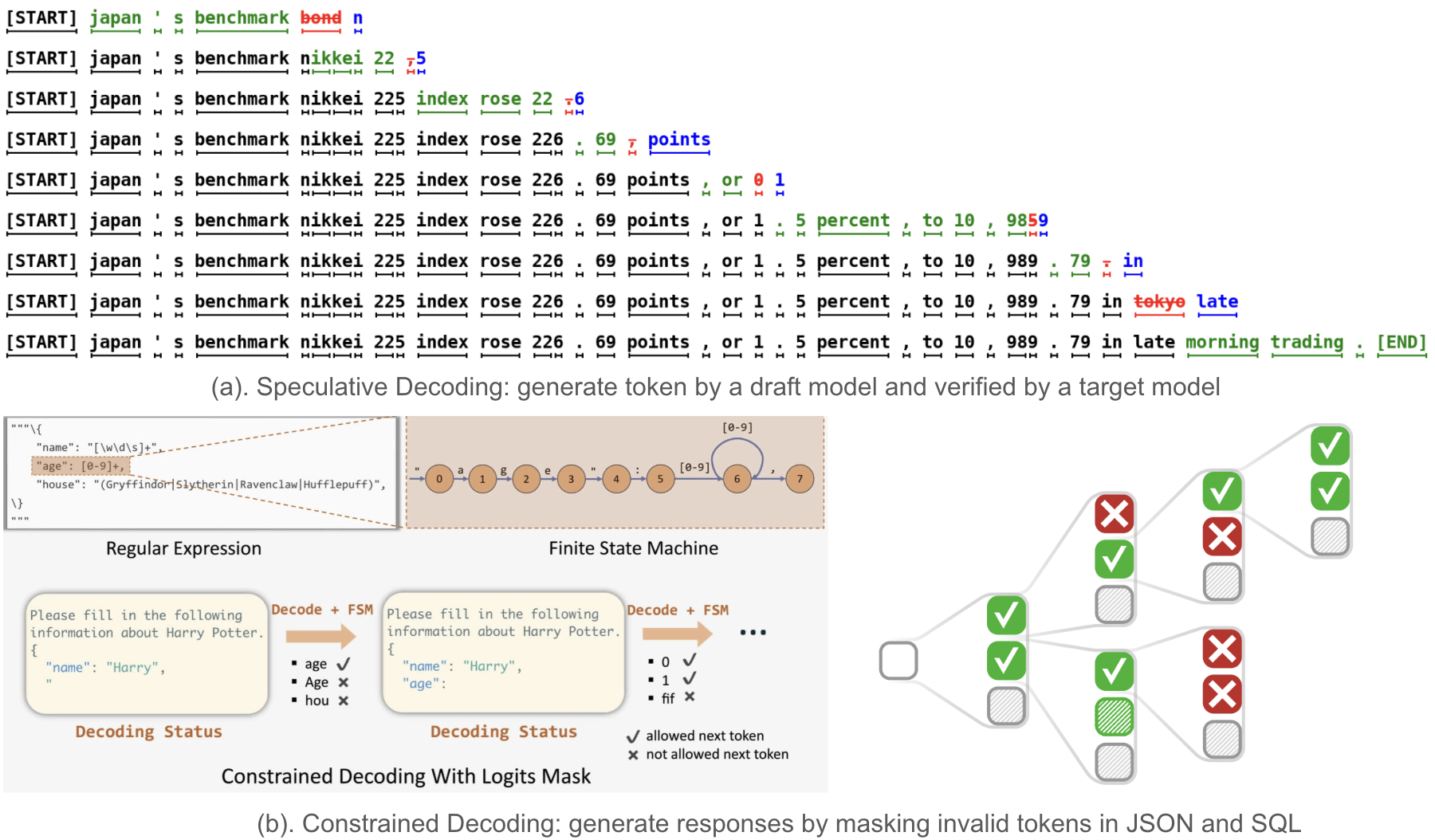}
    \caption{(a). Speculative Decoding to accelerate the generation process by generating with a small draft model in parallel, verified by a target model simultaneously, filtering invalid tokens. (b). Constrained decoding: generates valid tokens that satisfy constraint rules by masking invalid tokens.}
    \label{fig:Speculative_Constrained_Decoding}
\end{figure}

\subsubsection{Speculative Decoding} 

While probability modification refines token selection, speculative decoding accelerates the generation process by leveraging a smaller draft model in tandem with the target model, maintaining output fidelity while improving efficiency. In \cite{leviathan2023fastinferencetransformersspeculative}, the authors propose speculative decoding, a method to accelerate inference from LLM without altering their output distribution as shown in part (a) of Figure \ref{fig:Speculative_Constrained_Decoding}. A smaller draft model $q$ generates multiple candidate tokens which are then verified by the target model $p$. The acceptance rule is derived from speculative sampling: for a proposed token $x$, accept with probability $\min\!\left(1, \tfrac{p(x)}{q(x)}\right)$, reject otherwise, and if rejected, resample from the adjusted distribution $p'(x) \propto \max\!\big(0,\, p(x) - q(x)\big)$. This ensures that the final sample is exactly distributed according to $p$, i.e., identical to standard decoding.

\subsubsection{Constrained Decoding}

\textbf{Constrained Decoding} 

Constrained decoding ensures that generated sequences adhere to strict syntactic, semantic, or logical rules, making it indispensable for structured output formats like JSON, SQL, or formal reasoning tasks. This is realized by setting the logits of invalid tokens to negative infinity as shown in part (b) of Figure \ref{fig:Speculative_Constrained_Decoding}. In the small example, each time, three candidate next tokens are given, while the invalid ones that violate the constraints will be filtered out as shown in the red corssing, and the left one will be sampled by based on logits until the final response generation terminates. 

To ensure specific words or phrases appear in the generated output, researchers first explored lexically constrained decoding techniques. In \cite{hokamp2017lexicallyconstraineddecodingsequence}, grid beam search (GBS) extends standard beam search to handle lexical constraints by decoding over a 2D grid, where one axis represents time steps (t) and the other represents the number of covered constraint tokens (c). At each step, candidate hypotheses (partial generations) can expand in three ways:
\begin{enumerate}
    \item \textit{Generate}: open hypotheses can freely generate tokens from the model’s distribution.
    \item \textit{Start}: begin a new lexical constraint.
    \item \textit{Continue}: closed hypotheses (those in the middle of a constraint) must continue generating tokens within that constraint.
\end{enumerate} 
Lastly, the highest-scoring finished response is returned. Constrained beam search (CBS) \cite{anderson-etal-2017-guided} augments beam search with a finite-state machine (FSM) that encodes the constraints. Each FSM state represents which constraints have been satisfied and has its own beam of candidate sequences. When a candidate sequence starts or ends a constraint, the FSM transitions accordingly, placing the sequence into the beam of the appropriate next state. After each step, only the top-k sequences in each FSM are selected. Eventually, one beams corresponding to accepting states can output valid answers and the one with the highest probability will be selected as the final answer. In \cite{post2018fastlexicallyconstraineddecoding}, Dynamic Beam Allocation (DBA) is proposed as an efficient improvement over Grid Beam Search (GBS) for lexically constrained decoding. Unlike GBS, DBA keeps the total beam slots fixed and dynamically redistributes these beam slots among different constraint banks at each decoding step. This design ensures that the decoding cost remains constant regardless of the number of constraints.

Beyond lexical control, syntax-constrained decoding enforces grammatical and structural validity, especially important for formal or machine-readable outputs such as SQL queries. In \cite{scholak2021picardparsingincrementallyconstrained}, parsing incrementally for constrained auto-regressive decoding (PICARD) was proposed for constrained auto-regressive decoding that ensures the outputs of language models remain valid in formal languages such as SQL. PICARD uses incremental parsing to constrain decoding by selecting top-k token candidates and pruning invalid ones by assigning negative infinity to their logits. It employs three checks: (1) lexing to reject lexically invalid tokens, (2) parsing without guards to enforce grammatical correctness, and (3) parsing with guards to add semantic checks, ensuring valid SQL and correct abstract syntax tree structures. In \cite{Beurer_Kellner_2023}, Language Model Programming (LMP) extends traditional text prompting by integrating three elements: (1) natural language prompts, (2) programming constructs such as loops and conditions, and (3) explicit constraints. The process begins by constructing a query that specifies the decoder, decoding strategy (e.g., argmax or beam search), the prompt (including fixed text, holes like [VAR] for model generation, and placeholders like {VAR} for substitutions), control flow elements (such as conditions and loops), the target model, and any decoding constraints (e.g., output length or stopping phrases). This query is then parsed and executed like a Python program, where an interaction trace is built. During decoding, constraints are enforced immediately by masking invalid tokens, ensuring the generated output always satisfies the specified rules.

Moving from syntax to reasoning, logic-constrained decoding integrates symbolic constraints into the generation process to ensure logical soundness. In neurologic decoding \cite{lu2021neurologicdecodingunsupervisedneural}, a beam search-based algorithm that enforces logical constraints during text generation was proposed, where the constraints are expressed in Conjunctive Normal Form (CNF). In each step, the algorithm tracks each constraint's state, classifying it as reversible (can still be satisfied/violated) or irreversible (permanently satisfied/violated), prunes dead-end hypotheses that violate constraints irreversibly, groups the remaining candidates by their unique sets of satisfied constraints to ensure diverse solutions, and selects the best candidates from each group based on a score that balances language model likelihood and progress towards fulfilling unmet constraints. The beam search process continues until reaching out the final answer where all the constraints are satisfied. In \cite{lu2021neurologicaesquedecodingconstrained}, the authors propose NEUROLOGIC A*esque (NEUROLOGIC*) decoding, an extension of NEUROLOGIC decoding that incorporates A*-style lookahead heuristics. Unlike standard beam search, which only considers past probabilities, NEUROLOGICF augments candidate scoring with heuristic estimates of future cost. Several lookahead strategies are introduced—greedy, beam, and sampling—that approximate future continuations efficiently. For constrained generation, NEUROLOGICF builds on NEUROLOGIC’s logical CNF formulation and integrates estimates of future constraint satisfaction into the scoring function. 

Finally, constrained decoding has been applied to formal reasoning and theorem proving, where step-by-step logical validity is essential. In NATURALPROVER \cite{welleck2022naturalprovergroundedmathematicalproof}, the model was fine-tuned to generate proof with the context of reference through stepwise reference reconstruction objective. In inference, the reference is provided by either human or retriever, and stepwise constrained decoding is utilized to enforce reference usage for final prover generation. In \cite{poesia2023certifieddeductivereasoninglanguage}, the authors utilized LogicGuide, i.e., an external prover tool, with constrained decoding to guarantee reasoning reliability. To begin with, the LLM parses the query to generate assumptions, whose syntax is checked by LogicGuide. During inference, the LLM generates actions (e.g., [[object:]], [[axiom:]], [[infer:]]), and constrained semantic decoding (CSD) restricts outputs to valid logical steps based on prior context. This step iterates until reaching the answer or maximum number of threshold.

\textbf{Template+Placeholder} 

Constrained decoding realizes a strict format by modifying next tokens' logits based on different constraints. In comparison, the following methods firstly generated a template with placeholder, and then fill the placeholder with true content in the second iteration. Although they are different from constrained decoding, they are summarized here for straightforward comparison. In AutoTemplate \cite{iso2024autotemplatesimplerecipelexically}, the authors propose a simple two-stage framework for lexically constrained text generation: (1) generate a template containing placeholders conditioned on the input and constraint lexicons, and (2) replace the placeholders with the given lexicons.

\begin{figure}
    \centering
    \includegraphics[width=0.7\linewidth]{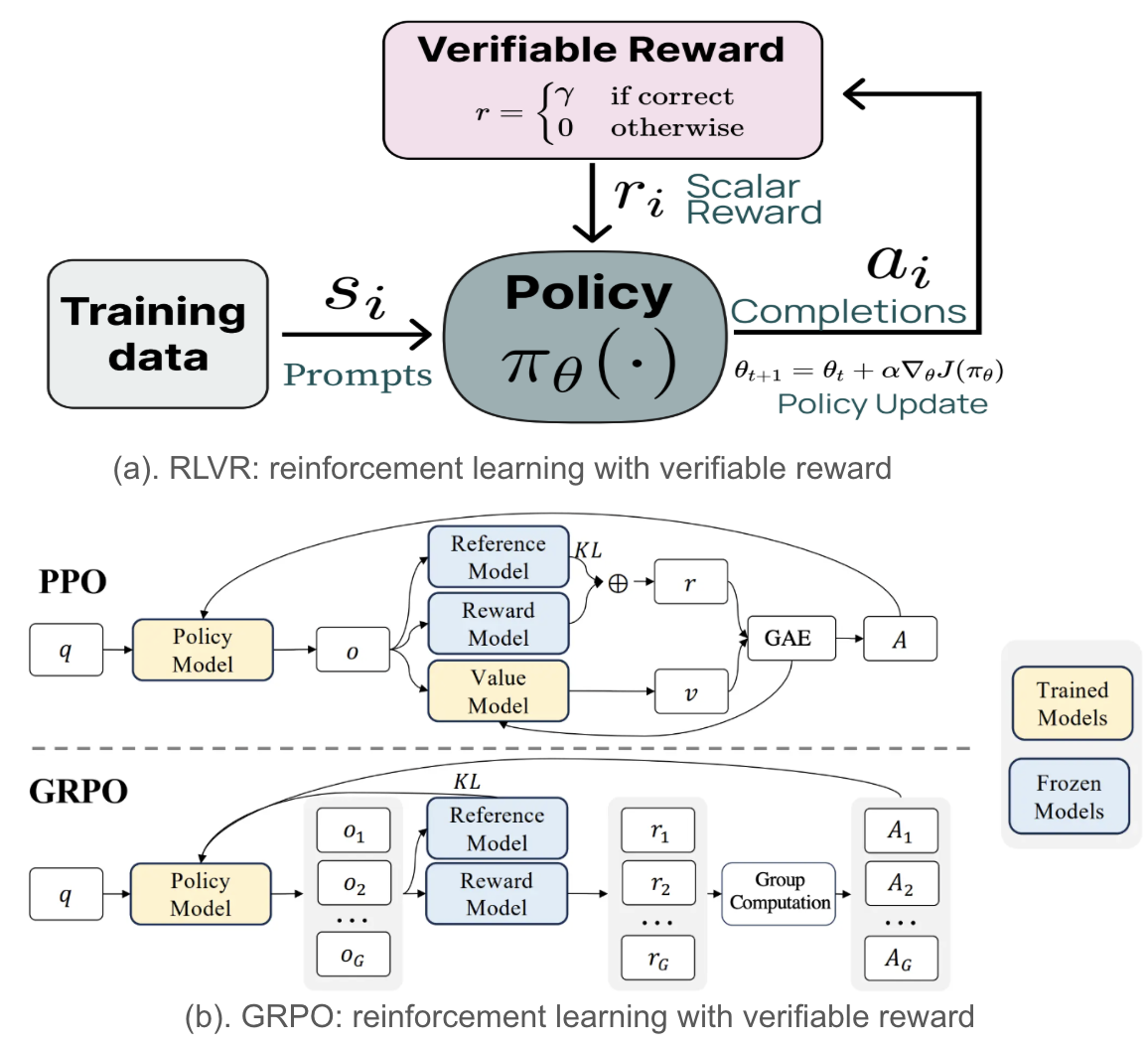}
    \caption{(a). RLVR: reinforcement learning with verifiable reward to enhance the model's capability on reasoning, (b). GRPO: group relative preference optimization to simplify PPO through alleviating the usage of value model}
    \label{fig:RLVR_GRPO}
\end{figure}

\subsection{Training for Long CoT}
This section reviews methods for enhancing LLMs' reasoning capabilities through training to generate long CoT before generating the final answer. We first discuss DeepSeek-R1, which leverages RLVR to automatically generate long CoT. Next, we examine Process Reward Models (PRMs), which provide stepwise evaluation of reasoning, and contrast them with outcome-based reward models (ORMs). Finally, we present Generative Reward Models (GRMs), which extend reward modeling with CoT reasoning.

\subsubsection{DeepSeek R1}
The authors propose two reasoning-focused LLMs—DeepSeek-R1-Zero and DeepSeek-R1—trained primarily with large-scale RL \cite{deepseekai2025deepseekr1incentivizingreasoningcapability} with verifiable reward as shown in part (a) of Figure \ref{fig:RLVR_GRPO}. First, DeepSeek-R1-Zero applies RL directly to a base model, i.e., DeepSeek-V3-Base without SFT. Using GRPO as shown in part (b) of Figure \ref{fig:RLVR_GRPO} and rule-based verifiable rewards that check results' format and accuracy, the base model gradually self-evolves into a strong reasoner with emergent behaviors like reflection, longer chains of thought and “aha” moments. However, it suffers from readability and language mixing issues. To improve usability and accelerate convergence, DeepSeek-R1 introduces a multi-stage pipeline starting from the same base model:
\begin{itemize}
    \item \textit{Step 1-Cold Start}: Collect thousands of high-quality, human-readable long CoT examples to fine-tune the base model with better readability as the starting point for RL.
    \item \textit{Step 2-RL}: Apply RL on the Cold Start SFT model with rewards for answer format, accuracy, and language consistency.
    \item \textit{Step 3-RFT}: The checkpoint from previous step generates around 800k high-quality samples, including reasoning and non-reasoning data, via rejection sampling. The base model is fine-tuned on the collected dataset for 2 epochs.
    \item \textit{Step 4-RL}: Apply RL again on the model derived from previous step, including rewards for helpfulness and harmlessness.
\end{itemize}
Last, reasoning ability is distilled into smaller open-source dense models using the 800k curated samples from DeepSeek-R1's SFT stage via SFT.

\subsubsection{Process Reward Model}

In \cite{lightman2023letsverifystepstep}, a process reward model (PRM) was trained using the PRM800K dataset and it could provide stepwise reward evaluation for LLM response. The results showed that the model trained with PRM outperformed outcome reward model (ORM). In \cite{pan2023letsreinforcestepstep}, the authors apply RLHF with both outcome-supervised reward models (ORMs) and process-supervised reward models (PRMs) to improve LLM reasoning in mathematics. They find that PRMs, which provide fine-grained stepwise rewards, significantly boost performance on simpler tasks but reduce performance on complex ones. In addition, in Best-of-N or critic-based methods, PRM can act the role of scoring function to rank different generated responses.

\begin{figure}
    \centering
    \includegraphics[width=\linewidth]{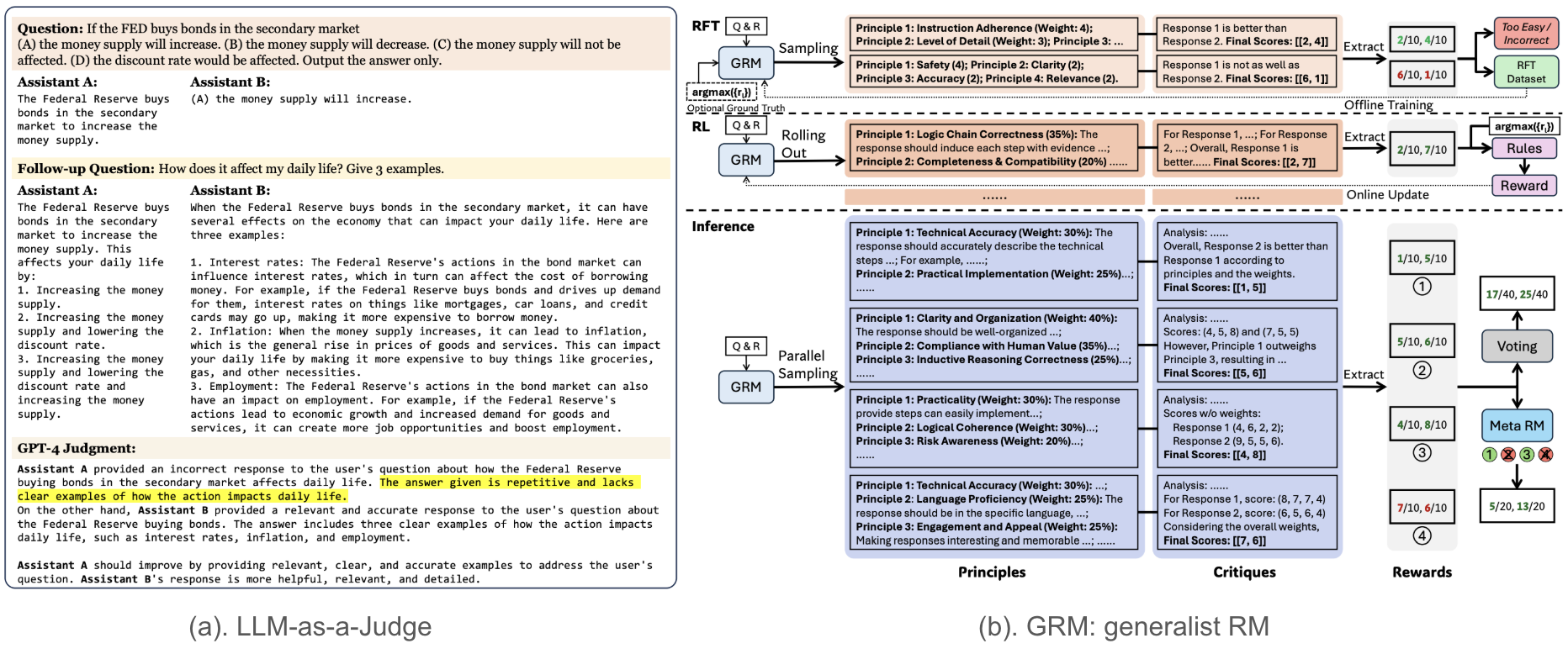}
    \caption{(a): LLM-as-a-Judge: utilize LLM to judge by outputting CoT and final evaluation, (b). GRM: generalist reward model to output principle and then judge based on the generated principle through RL.}
    \label{fig:LLMasAjudge_GRM}
\end{figure}

\subsubsection{Generative Reward Model}

\textbf{LLM-as-a-Judge and SFT}

In Reinforcement Learning from Human Feedback (RLHF), a reward model $r(x,y)$ is employed to assign a scalar score to a response $y$ given a prompt $x$. This model is typically trained using the Bradley–Terry (BT) framework to reflect human preference comparisons \cite{ouyang2022traininglanguagemodelsfollow}. While effective, such reward models provide only a numerical score without offering reasoning or interpretability. To address this limitation, the LLM-as-a-Judge paradigm was introduced, where LLMs themselves serve as scalable evaluators of conversational quality in open-ended, multi-turn settings \cite{zheng2023judgingllmasajudgemtbenchchatbot}. Instead of outputting a score directly, a model such as GPT-4 first generates a CoT rationale and then produces a final judgment. Later works further trained open-source LLMs on preference data with CoT using SFT to align reasoning with human evaluations \cite{hsu2025rateexplainciterec, wang2025directjudgementpreferenceoptimization}. Overall, this methodology improves the fidelity and robustness of evaluations through inference time scaling.

\textbf{Generative Reward Model by RLVR}

Previous studies leveraged LLMs’ inherent CoT reasoning abilities or fine-tuned them via SFT to strengthen their CoT-based response evaluation. Subsequent works introduced RLVR, where the reward signal is based on the alignment between the LLM’s judgments and human judgments, thereby enhancing the model’s capacity for long CoT reasoning and improving its overall judgment performance. In \cite{liu2025inferencetimescalinggeneralistreward}, the authors introduce Generalist Reward Modeling (GRM), designed to handle flexible inputs—such as single, paired, or multiple responses—and enable inference-time scaling of reward signals across diverse domains. To initialize the model, they employ Rejective Fine-Tuning (RFT). For each query-response pair with ground-truth labels, multiple trajectories of principles (evaluation criteria for judgment) and critiques are generated. Trajectories that are incorrect are discarded, while correct ones are used to fine-tune the GRM, ensuring proper alignment and output format. Following this, a rule-based online RL phase is conducted using GRPO. In this phase, the GRM first generates principles adaptively based on the prompt, then produces critiques conditioned on both the prompt and the principles. A binary reward is assigned (+1 if the pointwise score matches the ground truth, -1 otherwise), along with a KL penalty to avoid reward hacking. This approach encourages scalable behavior, such as producing diverse principles and accurate critiques. During inference, parallel sampling is used to generate multiple sets of principles and critiques. Final rewards are aggregated through voting or enhanced using a meta reward model (meta RM) that filters out low-quality samples. Experiments show that this method outperforms baselines on reward modeling benchmarks and demonstrates strong inference-time scaling. In \cite{wang2025helpsteer3preferenceopenhumanannotatedpreference}, a similar workflow was tested on Helpsteer v3 dataset and observed a clear improvement in reward modeling. 

In \cite{guo2025rewardreasoningmodel}, Reward Reasoning Models (RRMs) take a query with two candidate responses and are prompted to reason across multiple evaluation criteria, such as helpfulness, accuracy, and harmlessness before deciding the preferred response. To train RRMs via RLVR and GRPO, the model generates multiple reasoning paths based on different evaluation criteria and the reward is derived based on consistency with human evaluation. In addition, RRMs employ multi-response rewarding strategies, i.e., selecting the best response from multiple candidates. It is realized through (i) an ELO rating system that converts round-robin pairwise comparisons into numeric ratings, and (ii) a knockout tournament that progressively eliminates weaker responses.

\begin{figure}
    \centering
    \includegraphics[width=\linewidth]{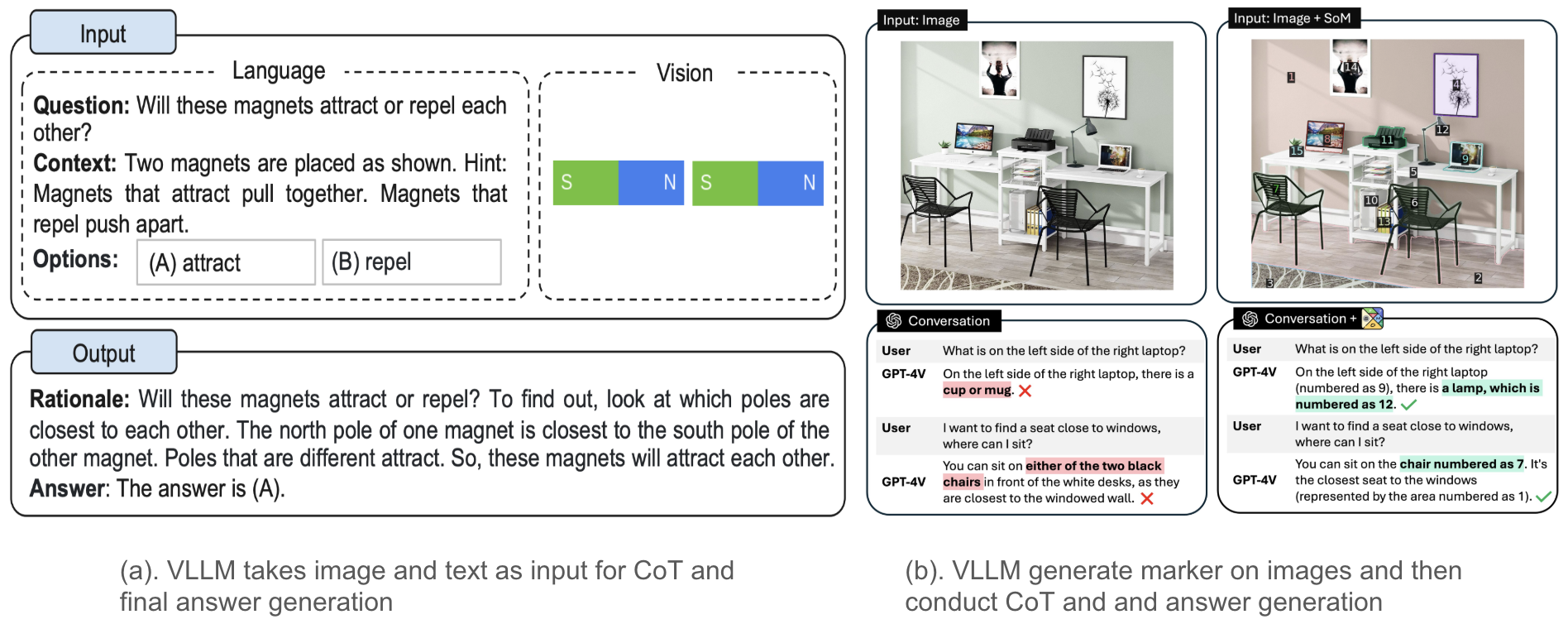}
    \caption{(a). VLLM takes image and text as input to generate CoT and final answer, (b). VLLM adds markers on the images to enhance VLLM's capability on multi-modality.}
    \label{fig:Multi-modal_CoT}
\end{figure}

\subsection{Multi-Modal Reasoning}

Recent advances in multi-modal reasoning have aimed to extend the CoT paradigm beyond text-only reasoning, enabling LLMs to jointly process and reason over heterogeneous modalities such as vision, language, and audio. These studies generally follow two complementary directions: (1) multi-modal input reasoning, which integrates visual or auditory signals directly into the reasoning process through multi-modal LLMs (MLLM) or modular expert like vision encoder; and (2) image modification approaches, which transform visual inputs—via marking, sketching, or coordinate scaffolding—to enhance grounding, spatial understanding, and iterative reasoning.

\subsubsection{Multi-Modal Input}

\textbf{MLLM} 

Recent works have explored enhancing CoT reasoning in LLMs by incorporating visual inputs, enabling models to jointly reason over text and images in a two-stage framework:.
\begin{enumerate}
    \item \textit{Rational Generation}: MLLM takes multi-modality inputs to generate rationales.
    \item \textit{Answer Generation}: MLLM takes multi-modality inputs and rationale to generate answer.
\end{enumerate}
In \cite{zhang2024multimodalchainofthoughtreasoninglanguage}, the authors propose multi-modal CoT (MCoT) through MLLM, a two-stage framework for CoT reasoning that incorporates both language and vision inputs as shown in part (a) of Figure \ref{fig:Multi-modal_CoT}. In the first stage, the MLLM generates rationales by fusing textual inputs (question, context, options) and images. In the second stage, these rationales are concatenated back with the original inputs to infer the final answer. In Image-of-Thought (IoT) \cite{zhou2024imageofthoughtpromptingvisualreasoning}, a two-stage framework was proposed 1. generate chain-of-multi-modal-rationales and 2. hybrid-rational-based answer refinement to extend CoT reasoning into the multi-modal setting. In the first stage, the image and query were sent to the MLLM to break down into sub-
tasks, to select action for each sub-task and execute the task to obtain the textual and visual rationale to form the final <Sub-Goal, Visual Rationale, Textual Rationale> triplet, i.e., Multi-modal Rationale Series (MRS). In the second stage, MRS were sent back to MLLM for the hybrid-rational-based final answer refinement. 

Expanding beyond rationale generation, \cite{mitra2024compositionalchainofthoughtpromptinglarge} introduces compositional CoT (CCoT), a zero-shot prompting method designed to improve compositional reasoning in MLLMs. CCoT first prompts the MLLM itself to generate a structured scene graph (objects, attributes, and relations) in JSON format from an input image and task prompt. In the second step, the MLLM model is prompted with the original image, task, and generated scene graph to produce the final response. This structured intermediate step encourages MLLMs to move beyond “bag-of-objects” reasoning and better capture compositionality. 

Unlike traditional CoT, which relies solely on text, Multi-modal Visualization-of-Thought (MVoT) \cite{li2025imaginereasoningspacemultimodal} enables models to generate interleaved verbal thoughts (text reasoning steps) and visual thoughts (image visualizations of reasoning traces). All the previous steps' information is utilized for the next step's generation until the termination of the generation. To improve visual quality, they introduce a token discrepancy loss that bridges the gap between text and image tokenizers in autoregressive MLLMs. 

Building on these multi-modal reasoning frameworks, Contrastive CoT (CoCoT) \cite{zhang2024cocotcontrastivechainofthoughtprompting} specifically addresses multi-image tasks, aiming to resolve two common failure modes in LMMs: insufficient fine-grained perception and unwanted blending of information across images. In the first stage, the MLLM is prompted to identify the similarities and differences among these images. Then, in the second stage, the MLLM is prompted to utilize the comparison analysis among these images to answer the question with details. 

\textbf{LLM+multi-modal experts}

To enable multi-modal reasoning for LLM without additional training, recent studies have explored modular composition frameworks that orchestrate multiple pretrained foundation models—such as LLMs, VLMs, and ALMs—through language-based prompting interfaces. In Socratic Models (SMs) \cite{zeng2022socraticmodelscomposingzeroshot}, a framework for zero-shot multi-modal reasoning was proposed to compose multiple pretrained foundation models through language-based prompting as an intermediate representation. The query is firstly decomposed into sub-queries that foundational models could answer. Then, for each step, a suitable model like LLM, VLM like CLIP or ALM like Whisper was selected and the previous step's output is combined with the current sub-query into a suitable hudrated sub-query and sent the selected foundation model to for generating the response. Here, VLM utilized CLIP for image-text pair and ALM like Whisper to transform audio into transciption. During this iteration, it may include "closed-loop" feedback to verify previous steps. This modular prompting approach leverages complementary knowledge across modalities without finetuning. In MM-ReAct \cite{yang2023mmreactpromptingchatgptmultimodal}, a LLM is combined with vision experts to conduct multi-modal complex reasoning tasks. To begin with, a query is broken down into sub-queries by the LLM. If action is needed to extract relevant information, different vision tools will be called iteratively via prompts using file paths and watchwords. Lastly, the extracted information, i.e., observation will be sent back to LLM for further reasoning and final answer generation. In \cite{zheng2023ddcotdutydistinctchainofthoughtprompting}, Duty-Distinct Chain-of-Thought (DDCoT) decomposes multi-modal query into sub-queries and explicitly separates visual recognition from textual reasoning. Sub-queries that cannot be answered without the image are tagged as Uncertain via negative-space prompting which encourage LLM not to answer if not confident and sent to visual model to obtain visual complements. Then, the LLM performs joint reasoning over the query, generated rationales, and visual complements while being prompted to critically evaluate and select valid information.

\subsubsection{Image Modification}

To further enhance multi-modal reasoning, recent studies have begun to move beyond directly feeding unprocessed multi-modal inputs into LLMs, focusing instead on transforming or augmenting visual inputs to improve the model’s ability to ground and interpret image features. In Set-of-Mark (SoM) \cite{yang2023setofmarkpromptingunleashesextraordinary}, a image was firstly segmented into different meaningful regions using external segmentation model, and each region is added with an identifiable mark to form a new image as shown in part (b) of Figure \ref{fig:Multi-modal_CoT}. This marked image, together with either plain or interleaved text prompts, is fed into GPT-4V. By explicitly associating regions with “speakable” marks, SoM unleashes GPT-4V’s emergent visual grounding ability. In \cite{shtedritski2023doesclipknowred}, the authors propose visual prompt engineering for VLLM by directly editing images instead of only manipulating text. To begin with, an object detection model is applied to detect a set of candidate region with bounding boxes and red circles are added within each bounding box to mark the objects to generate different images with each image having one object stressed. Lastly, a CLIP model will be utilized to find the object that has the largest score, i.e., most relevant to the query. In Visual Sketchpad \cite{hu2024visualsketchpadsketchingvisual}, the authors propose to equips MLLMs with a visual sketchpad, enabling them to draw intermediate sketches (e.g., lines, boxes, masks, plots) as part of their reasoning process to improve performances on math and vision tasks. In each step, the multi-modal LLM is utilized to generate the next step planning based on the query, previous text and visual reasoning, then synthesize Python code to plot new figures or utilize vision models to detect or segment objects and lastly the updated figure will be sent back to the multi-modal LLM for the next step iteration. 

While the aforementioned methods perform a single round of image modification before reasoning, subsequent research has explored iterative image refinement to progressively improve performance. In prompting with iterative visual optimization (PIVOT) \cite{nasiriany2024pivotiterativevisualprompting}, the MLLM is shown with candidate actions to choose from in a modified image iteratively to improve robots' performance on choosing suitable actions. In each iteration, multiple candidate actions were sampled and they were then labeled on the original image and sent to MLLM to choose the most potential action. After that, the selected regions are focused to sample more actions in this more fine-grained regions to start the next iteration. After several iterations, the final action will be selected and executed by the robot.

The previous works focus on image modification on 2D domain, while the following work aims at estimating 3D distances and relationships based on modification on 2D images. In 3DAxiesPrompts \cite{liu20233daxiespromptsunleashing3dspatial}, a 3D Cartesian coordinate was applied to the image through 1. origin determination, 2. axis orientation based on object's natural orientation and 3. scale marking by adding tick marks and numerical labels along each axis. Then, the modified image is combined with user query and send to LLM for answer generation to handle tasks such as measuring object dimensions, 2D-to-3D point reconstruction, 2D-to-3D point matching, and 3D object detection. In \cite{lei2024scaffoldingcoordinatespromotevisionlanguage}, the authors propose SCAFFOLD prompting, a simple visual prompting scheme to enhance vision-language coordination in multi-modal LLMs. SCAFFOLD overlays a dot matrix on the input image, with each dot labeled by Cartesian coordinates (2D for single images, extended to 3D for image sequences). These coordinates serve as visual anchors to reference positions explicitly and are also included in the textual prompt for reasoning about spatial and compositional relationships.

\subsection{Model Ensemble}
In this section, we discuss two approaches: model ensemble and model merging. In model ensemble, the output distributions of different models are combined with weights to improve performance on downstream tasks. In comparison, model merging merges different models of the same architecture. Model merging is not directly related to inference-time scaling. However, it is a counterpart of model ensemble and should be introduced.

\subsubsection{Different LLM Architectures: Token-Level Ensemble} 

\textbf{Token Vocabulary Alignment} 

The following works focus on ensembling models of different architectures by aligning their token vocabularies. In \cite{wan2024knowledgefusionlargelanguage}, the vocabulary of different LLMs are matched through minED and the probability for next token across different LLMs is merged through either 1. Minimum Cross-Entropy (MinCE), i.e., for each token, selecting the probability distribution from the source model that had the lowest cross-entropy loss or 2. Average by Cross-Entropy (AvgCE), i.e., taking a weighted average of all source models' distributions, weighted by their confidence which is the inverse of cross entropy. Eventually, the combined response and probability was utilized for better performance in fine-tuning a model. In Pack of LLMs \cite{mavromatis2024packllmsmodelfusion}, a set of LLMs was chosen, and their token vocabulary was aligned to a specific vocabulary through minimum edit distance (MinED). MinED is the minimum number of single-character edits (insertions, deletions, or substitutions) required to change one string into another. Then, the prompt are sent to the set of LLMs to compute perplexity which measures the confidence of a specific LLM on this query. Next, the weights to combine these LLMs are generated. Two methodologies were proposed: 1. simple PackLLM and 2. optimized PackLLM. In simple PackLLM, the weights were computed through softmax based on the perplexity of the prompt. In optimized PackLLM, it is realized iteratively by first combining with weights the two most reliable LLMs and then combining them to minimize perplexity, continuing with the remaining LLMs. Lastly, in inference, the probability of each token is a weighted combination of the pack of LLMs and it iterates until the termination condition. In \cite{xu2024bridginggapdifferentvocabularies}, the authors propose EVA (Ensemble via Vocabulary Alignment), a method to ensemble LLMs with different vocabularies at a fine-grained token-level during the generation process. Based on the embedding of the shared tokens across different LLMs, a linear transformation matrix is constructed and some noise reduction and variance reduction techniques are applied to build the transition dictionary. In the inference stage, the set of LLMs predicts the probability of the next token and a filtering technique is applied where if a model's top-1 probability token is not in the top-n tokens of any other LLM, this LLM will be filtered out and the remaining LLMs' next token probabilities are averaged to predict the token. Cool-Fusion \cite{liu2025coolfusionfuselargelanguage} is a training-free approach to fuse heterogeneous LLMs with different vocabularies through 1. fine-grained fusion and 2. coarse-grained fusion. In each iteration of fine-grained fusion, every LLM generates a candidate text segment—either the shortest decodable unit or an aligned segment that is decoded by all models. All candidate segments are then evaluated by every LLM using perplexity, and the one with the lowest average perplexity is selected to extend the generation. This process alternates between generation and evaluation until the completion of fine-grained fusion. Lastly, each LLM generates a separate response and combine with the fine-grained fusion response for response-level reranking by perplexity.

\textbf{Token Probability Ensemble} 

The following works focus on ensembling the next token prediction probabilities of different LLMs. In DEXPERTS \cite{liu2021dexpertsdecodingtimecontrolledtext}, a new decoding methodology was proposed utilizing three models: 1. a large base model $M$, 2. a small fine-tuned model with specific domain knowledge $m+$ and 3. a small anti-fine-tuned model with anti-domain knowledge $m-$ where our desire lies in injecting the knowledge difference between $m+$ and $m-$ to improve the performance of $M$. Given the prefix context, the probability of $M$, $m+$ and $m-$ were $P_M(y|x)$, $P_m^+(y|x)$ and $P_m^-(y|x)$ respectively. Then, the modified probability was 
\begin{equation}
P(y|x) = P_M(y|x) + \alpha \times (P_m^+(y|x) - P_m^-(y|x))
\end{equation}
Using this modified probability for iteratively decoding the next token derives the final response. This product-of-experts formulation ensures tokens are favored when they are likely under the expert but unlikely under the anti-expert. In \cite{li2024purifyinglargelanguagemodels}, a simple, provably effective method was proposed to purify LLMs from negative effects of uncurated training data—copyright infringement, data poisoning, and personally identifiable information (PII) leakage—by ensembling them with a small, benign language model at the logit level using the scaled logits ensemble, i.e., $CP-\Delta_{KL}$ algorithm.

\subsubsection{Different LLM Architectures: Response-Level Ensemble}

The following works focus on ensembling LLMs of different architectures by comparing and combining their individual responses. In LLM-Blender \cite{jiang2023llmblenderensemblinglargelanguage}, a two-stage framework was proposed to combine the responses from multiple LLMs. To begin with, given a set of LLMs, each LLM generates a response to the query, and a pairwise ranker, PAIRRANKER, is trained to predict the winner. Given $N$ responses, $N^2$ comparisons were conducted to build a comparison matrix and the methodology of MaxLogits was applied to derive the final ranking of these responses. Next, the query and the top-k responses were sent to a LLM, GENFUSER, for response improvement. In \cite{farinhas2023empiricalstudytranslationhypothesis}, the authors investigate ensembling strategies for LLM-based machine translation by combining translation candidates generated from multiple LLMs. They explore multiple strategies for generating translation candidates and aggregating them into a final translation via methods including quality-based reranking, ChooseBest, GenerateBest, and Minimum Bayes Risk (MBR) decoding. MBR decoding, which selects the hypothesis with the highest average similarity to all other candidates, is shown to be the most effective. 

Moving beyond direct response comparison, another line of research ensembles models at the knowledge level, integrating domain-specific expertise into a base LLM through retrieval and selection. In Knowledge card \cite{feng2024knowledgecardfillingllms}, the authors propose a modular framework to fill LLMs’ knowledge gaps by integrating specialized domain-specific LMs called knowledge cards. Each card serves as a parametric repository trained on targeted corpora, and at inference time, selected cards generate background documents for the base LLM. To ensure quality, 1. relevance selector, 2. pruning selector and 3. factuality selector are utilized to filter and condense the knowledge. Lastly, two different methodologies were proposed 1. bottom-up approach and 2. top-down approach for knowledge integration. In the bottom-up approach, all the knowledge cards were utilized to generate relevant documents and all the integrated knowledge were sent to LLM for answer generation. In comparison, in the top-down approach, the LLM is firstly asked if it needs external information. If no external information is needed, LLM can directly generate the answer. Otherwise, the most relevant knowledge card is selected by either 1. automatic selection or 2. explicit selection. Then, the selected knowledge card will generate new relevant document and another round of iteration starts until LLM thinks that it can generate the response without further information. 

Finally, some approaches aim to learn an explicit fusion model that adaptively combines multiple expert LLMs with complementary strengths. In \cite{wang2024fusingmodelscomplementaryexpertise}, the authors propose the fusion of experts (FoE) framework, which learns a fuser to combine outputs of multiple expert models with complementary domain knowledge in both classification and generative tasks. In inference, it is realized in standard FoE or FrugalFoE. In standard FoE, all $K$ experts compute the final layer's embedding and they are concatenated to send to the FoE for final generation. In FrugalFoE, it starts with the most promising expert and utilizes KNN on validation to estimate the benefit of incorporating the next most promising expert over the cost. This iteration continues until the termination of the incorporating the next expert. The FrugalFoE can reduce the number of experts queried while maintaining high accuracy. 

\subsubsection{Same LLM Architectures Ensemble}

\textbf{Models Merge into One Dense Model}

The most straightforward way to merge models of the same LLM architecture is through direct parameter averaging \cite{cha2021swaddomaingeneralizationseeking}. In Ensemble of Averages \cite{arpit2022ensembleaveragesimprovingmodel}, the authors propose using a simple moving average (SMA) of model parameters to stabilize performance and improve correlation between validation and test accuracy. Building on this, they introduce an ensemble of average models (EoA), which outperforms traditional ensembles of raw models. Their approach is hyperparameter-free and computationally efficient. In model soups \cite{wortsman2022modelsoupsaveragingweights}, the models derived from the combination of different hyperparameters were combined rather than selecting the best from them. Three merging methods were proposed. The first method, uniform soup took average of these models. The second method, greedy soup firstly sorted these models based on the performances on the validation set and then iteratively merge the next best model with the previous selected models as long as the merged model did not deteriorate the performance on the validation set. The final method, learned soup, optimizes the mixing coefficients and the temperature scaling parameter to combine multiple models on a validation set using AdamW. Similar ideas are found in Diverse Weight Averaging (DiWA) \cite{rame2022diverse} to average models derived from different hyperparameters, where the authors showed that model weight averaging can reduce variance in out-of-distribution (OOD) generalization through bias–variance–covariance–locality decomposition. 

Beyond simple averaging, structured merging frameworks explicitly model and resolve parameter interference between different models. In TIES-MERGING \cite{yadav2023tiesmergingresolvinginterferencemerging}, a three-step framework was proposed to merge multiple task-specific models. To begin with, Trim: the changes of the each LLM's parameters are calculated and only changes larger than a specific threshold were maintained while the others were set back to 0. Then, Elect Sign: by summing up different models, each parameter could move in the positive or the negative direction and this will be recorded. Lastly, Merge: for each parameter, if the changes in each LLM is in the same direction as the recorded direction, the changes will be averaged and updated. In FUSECHAT \cite{wan2024fusechatknowledgefusionchat}, a set of $K$ LLMs are selected and one of them will be utilized as reference and the remaining $K-1$ LLMs will be utilized for distillation where the knowledge of the original LLM will be transferred to the reference LLM. Then, the obtained $K$ LLMs utilizing the architecture of the reference LLM will be merged to obtain the final merged LLM through SCE, i.e., select, calculate and erase. In the "select", the top $\tau\%$ parameters that have been changed will be recorded. Then in "calculate", the merging weights are calculated based on the magnitude of squared changed magnitude. Lastly, in "erase", if two LLMs change in different directions, the one with smaller magnitude will be erased to avoid cancellation and noise. In RegMean \cite{jin2025datalessknowledgefusionmerging}, the authors address the challenge of dataless knowledge fusion, where multiple fine-tuned language models are available but their training data is private. The new merged model was derived by 1. directly averaging the non-linear layers and 2. using $\mathbf{W}_{M} = \left( \sum_{i=1}^{K} \mathbf{G}_i \right)^{-1} \left( \sum_{i=1}^{K} \mathbf{G}_i \mathbf{W}_i \right)$ for weighted summation in linear layers where $K$ is the number of models, $\mathbf{G}_i$ is the Gram matrix for model $i$, and $\mathbf{W}_i$ is its weight matrix. This formulation minimizes prediction differences between merged and individual models without requiring training data.

\textbf{Models Merge into One MoE Model}

To combine the strengths of multiple domain-specific models into a unified framework, several works have explored merging dense models into a single MoE architecture. In Branch-Train-Mix \cite{sukhbaatar2024branchtrainmixmixingexpertllms}, a dense model is copied multiple times, each copy is fine-tuned on domain-specific data and finally merged as a sparse MoE model , where the feedforward sub-layers serve as experts while the remaining parameters, i.e., self-attention are averaged and shared acorss experts. Lastly, a router is initialized and then fine-tuned to distribute different queries to different experts. In inference, the derived MoE model will follow the inference process of a normal MoE model. In \cite{ding2024masteringtextcodemath}, ULTRAFUSER was proposed to fuse three LLMs focusing on text, code and math by training and utilizing a token-level gating model. In inference, for each token, all three LLMs are activated which is different from MoE. The gating model will output a score between 0 and 1 to combine the token prediction from these three different LLMs. 

Building on this idea of integrating model specialization across different domains while maintaining inference efficiency, \cite{gururangan2023scalingexpertlanguagemodels} introduced Cluster-Branch-Train-Merge (c-BTM). The method first clusters documents into k clusters, then trains k copies of the model on the data within each cluster. Eventually, in inference, the top-k similar clusters were searched by TF-IDF and weighted-ensembled based on context–cluster similarity for final answer generation.

\textbf{LLM with PEFT}
The following works focus on improving LLM's performances on multiple PEFT adapters. In AdaMix \cite{wang2022adamixmixtureofadaptationsparameterefficientmodel}, different adapter were trained by randomly routing and training a batch of data. During training, input batches are stochastically routed through different adaptation modules, and consistency regularization via KL divergence is applied to stabilize learning. To keep inference efficient, the multiple modules are merged into a single adaptation per layer, matching the cost of the underlying PEFT method. In \cite{zhang2023composingparameterefficientmodulesarithmetic}, the authors propose a training-free method to compose parameter-efficient finetuning modules like LoRA through linear arithmetic in parameter space. They define two basic operators — addition (to aggregate skills) and negation (to retract skills) — and show that these are composed to create new modules for tasks such as distribution generalization, multi-task learning, unlearning, and domain transfer. In LoraHub \cite{huang2024lorahubefficientcrosstaskgeneralization}, LoRA adapters trained on diverse upstream tasks are dynamically composed via element-wise weighted combination to enable efficient cross-task generalization. For unseen tasks, few-shot examples guide coefficient optimization using gradient-free methods like Covariance Matrix Adaptive Evolution Strategies (CMA-ES), requiring no additional parameters or gradients.

\section{Input: RAG}

\textbf{RAG Background}

Prior to the era of retrieval augmented generation (RAG), there was some work focused on Q\&A with retrieval and generative models. REALM \cite{guu2020realmretrievalaugmentedlanguagemodel} included the retrieval as an augment process for the pretraining of LLM.
In \cite{izacard2021leveragingpassageretrievalgenerative}, a hybrid retriever, combining a sparse retriever (BM25 \cite{robertson2009probabilistic}) and a dense retriever (DPR \cite{karpukhin2020densepassageretrievalopendomain}), was utilized to retrieve relevant documents to answer the question. Afterward, the question and retrieved materials were combined into an encoder-decoder seq2seq model. The encoder extracted information from different materials and concatenated before being fed into the decoder for question answering. This workflow is nearly the same as the latest RAG, except that the encoder-decoder seq2seq model was replaced by a decoder-only LLM.

\textbf{RAG Workflow: Sequence and Token}

\begin{figure}
    \centering
    \includegraphics[width=\textwidth]{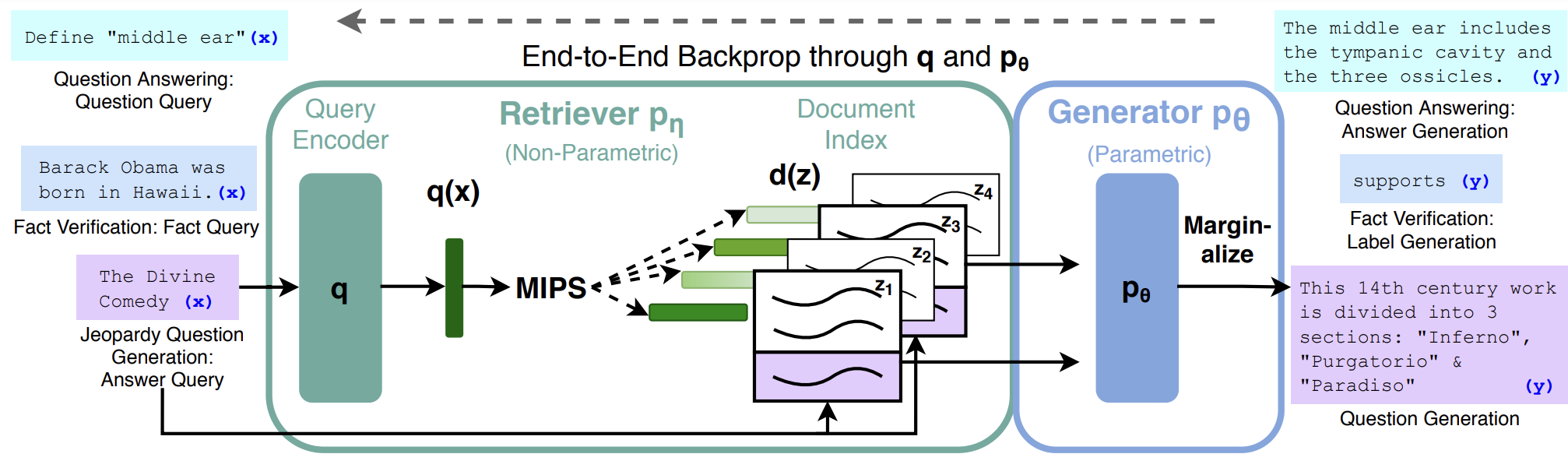}
    \caption{The initial pipeline in the RAG paper}
    \label{fig: RAG pipeline}
\end{figure}

The name of RAG comes from \cite{lewis2021retrievalaugmentedgenerationknowledgeintensivenlp} as shown in Figure \ref{fig: RAG pipeline}. This paper proposed a sequence-level RAG model as shown in Eq. \ref{eq:rag-sequence} and a token-level RAG model as shown in Eq. \ref{eq:rag-token}. 

\begin{equation}
p_{\text{RAG-Sequence}}(y|x) \approx \sum_{z \in \text{top-K}(p(\cdot|x))} p_{\eta}(z|x) \cdot \prod_{i=1}^N p_{\theta}(y_i \mid x, z, y_{1:i-1})
\label{eq:rag-sequence}
\end{equation}

\begin{equation}
p_{\text{RAG-Token}}(y|x) \approx \prod_{i=1}^N \sum_{z \in \text{top-K}(p(\cdot|x))} p_{\eta}(z|x) \cdot p_{\theta}(y_i \mid x, z, y_{1:i-1})
\label{eq:rag-token}
\end{equation}

Here, $x$ refers to the user’s query, and $y$ refers to the responses generated by the LLM. $z$ refers to retrieved documents, and they are retrieved from different retriever like BERT \cite{devlin2019bertpretrainingdeepbidirectional}, SentenceBERT \cite{reimers2019sentencebertsentenceembeddingsusing}, DPR \cite{karpukhin2020densepassageretrievalopendomain}, E5 \cite{wang2024textembeddingsweaklysupervisedcontrastive}, Deberta \cite{he2021debertadecodingenhancedbertdisentangled}. $p_{\eta}(z|x)$ referr to the probability of the retriever to retrieve this document $z$ based on the query $x$ and $p_{\theta}(y_i \mid x, z, y_{1:i-1})$ represents the probability of the generator generating the $i$-th token based on the prompt $x$, the retrieved documents $z$ and previous generated tokens $y_{1:i-1}$. The probability of RAG-sequence generating a specific response is based on summing up the probability of generating this response over each single document. By contrast, RAG-token generates each token by summing the probabilities over different documents and iterates for all tokens.

\textbf{RAG with Beam Search}

When decoding with beam search, RAG-token is directly transformed into maximizing each token's probability, i.e., $\sum_{z \in \text{top-K}(p(\cdot|x))} p_{\eta}(z|x) \cdot p_{\theta}(y_i \mid x, z, y_{1:i-1})$. On the contrary, RAG-sequence cannot solve this problem directly. Instead, RAG-sequence proposes 1. Thorough Decoding (Exact Marginalization) and 2. Fast Decoding (Approximate Marginalization) to generate multiple responses with beam search on each retrieved single document and choose the final response with the maximum probability.

In practice, RAG-token is more frequently utilized where given a query, multiple chunks of different documents are collected and combined into a hydrated prompt and sent to LLM for generation, where the attention scores in the transformer layers serve as the role of weight combination on different documents and change dynamically for each token. Another work similar to RAG from that period is found in \cite{borgeaud2022improvinglanguagemodelsretrieving}.

In this section, we will review RAG from seven perspectives: 1. Query Expansion, 2. Data, 3. Retriever and Reranker, 4. LLM Generation, 5. RAG with Tools and Rules, 6. Multi-Modal RAG, 7. Collaborative RAG.

\subsection{Query Expansion}

For Q\&A in RAG, the initial query provided might not have conveyed all the required information, so it might not have been able to make full use of the power of the LLM. To solve this problem, two types of methodologies have been proposed to expand the query. The first methodology directly enriched the query by the LLM. The second methodology augmented the query with a generated document or draft answer to help retrieve relevant materials.

\subsubsection{Query Expansion by Rewriting}

\textbf{Query Rewriting: Text} 

\begin{figure}
    \centering
    \includegraphics[width=\textwidth]{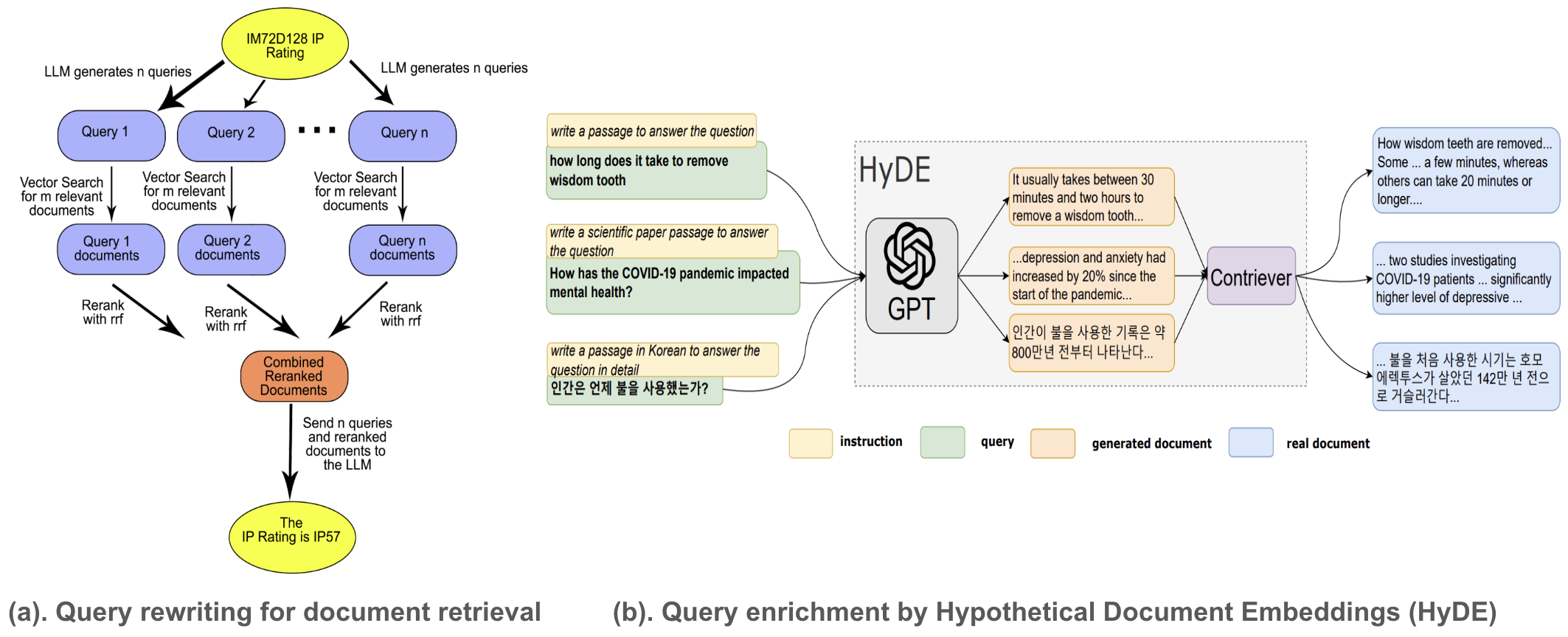}
    \caption{Query expansion: (a). rewrite the query to different versions and extract different corresponding documents; (b). use the query to generate HyDE to enhance retrieval}
    \label{fig: QueryRewriting_HyDE}
\end{figure}

The following papers directly enrich the query, especially when it is ambiguous, to facilitate the retrieval process and final answer generation in RAG. In \cite{Rackauckas_2024}, RAG-Fusion generated multiple rephrased or related queries, and multiple documents are retrieved based on each individual query as shown in part (a) of Figure \ref{fig: QueryRewriting_HyDE}. For each retrieved document, it was retrieved by multiple queries at different ranks, and Reciprocal Rank Fusion (RRF) is utilized to assign reciprocal scores to each document from multiple query results as shown in Eq. \ref{eq: RRFscore} where $X$ refers to the set of query, $x$ refers to one single query of $X$ and $z$ refers to a specific retrieved documents and k is a smoothing parameter. Based on RRFscore, the retrieved documents are reranked and the top-K documents are selected to send to LLM for final answer generation.

\begin{equation}
\text{RRFscore}(z \mid X) \;=\; 
\sum_{x \in X} \frac{1}{k + \operatorname{rank}(z \mid x)} 
\label{eq: RRFscore}
\end{equation}

In Diverse Multi-Query Rewriting (DMQR)-RAG \cite{li2024dmqrragdiversemultiqueryrewriting}, the authors proposed that the diversity of the rewritten queries in RAG-Fusion was low, and they proposed four rewriting strategies: 1. General Query Rewriting (GQR) to denoise and refine the query without losing information, 2. Keyword Rewriting (KWR) to extract key nouns/subjects and facilitate searching queries on search-engine, 3. Pseudo-Answer Rewriting (PAR) to use the LLM to generate a “fake” answer and 4. Core Content Extraction (CCE) to strip away unnecessary details, keeping only the essential parts. In addition, adaptive rewriting strategy selection was applied to prompt LLM with few-shot examples to select suitable strategies to generate multiple rewritten prompts to improve efficiency. In Diversify-verify-adapt (DIVA) \cite{in2025diversifyverifyadaptefficientrobustretrievalaugmented}, diverse queries, pseudo-interpretations are generated and their relevant documents are retrieved and top relevant documents are selected. Next, LLM is utilized to verify the usefulness of the retrieved documents on answering the query as a classifier: 1. "Useful", 2. "PartialUseful" and 3. "Useless". If it is classified as "Useful" or "PartialUseful", the retrieved documents were utilized for generation. Otherwise, no retrieved document will be utilized and LLM will directly generate the response. In Insight-RAG \cite{pezeshkpour2025insightragenhancingllmsinsightdriven}, insight identifier detects critical question and insights from the query and insight miner retrieves the passages that address these insights before being sent to LLM for final response generation.

\textbf{Query Rewriting: Embedding} 

The previous works directly rewrite the query, while another direction applies different intermediate embeddings of LLM as new query embeddings to represent different aspects of the original query. In \cite{besta2025multiheadragsolvingmultiaspect}, Multi-Head RAG (MRAG) utilizes each head of embedding in the LLM’s last attention layer to retrieve documents. For each document in each head, a relevance score is computed based on the cosine similarity of embeddings combined with the ranking selection through reranker. For each document, the scores in different heads are weighted aggregated and reranked to produce a final relevance score for each document. Finally, the top-k document chunks are selected and passed to the LLM for the final response generation.

\subsubsection{Query Expansion by Generation}

\textbf{Query Expansion by Hypothetical Document Generation} 

The other category of methods generate hypothetical documents or draft answer to augment the query for retrieving relevant documents. In \cite{gao2022precisezeroshotdenseretrieval}, the authors proposed Hypothetical Document Embeddings (HyDE) as shown in part (b) of Figure \ref{fig: QueryRewriting_HyDE}:
\begin{enumerate}
    \item \textit{Document Generation}: leverages a LLM to create multiple hypothetical documents tailored to address the user's query accurately.
    \item \textit{Retrieval}: calculates embeddings for each generated document and computes their average to enhance retrieval precision.
    \item \textit{Response Generation}: utilizes the LLM to produce a concise and accurate final response based on the retrieved documents.
\end{enumerate}
Even though the generated hypothetical document may have incorrect information, it captures the relevance pattern for the correct answer. 

Query2Doc \cite{wang2023query2docqueryexpansionlarge} also utilizes the hypothetical document generation idea with few-shot examples to enrich the query for retireval process. In particular, for sparser retirever, the initial query is repeated multiple times $x^{+} = concat(x \times n, d^{'})$ while for the dense retriever, the query is not repeated and separated with the generated document through a "[SEP]" token $x^{+} = concat(x, [SEP], d^{'})$ where $d^{'}$ refers to the generated documents. In \cite{jagerman2023queryexpansionpromptinglarge}, the authors compared Query2Doc, Query2Expansion and CoT. Query2Doc utilizes the LLM to generate documents and Query2Expansion generates expansion terms similar to Pseudo-Relevance Feedback (PRF) \cite{10.1108/eb026866}, while CoT generates chain-of-thought reasoning and final answer to break the query down and provide expansion terms. For Query2Doc, Query2Expansion and CoT, different settings including zero-shot, few-shot and PRF are tested. Their results showed that CoT had the best performance compared with Query2Doc and Query2Expansion.

\textbf{Query Expansion by Draft Answer Generation} Previous works focus on query expansion by hypothetical document generation, while the following works focus on query expansion by draft answer or initial summaries. In Language Model as Retriever (LameR) \cite{shen2023largelanguagemodelsstrong}, given the query and an initial set of retrieved documents, an LLM first generates a draft answer. This draft answer is then used to augment the query, enabling a second round of document retrieval. In \cite{shao2023enhancingretrievalaugmentedlargelanguage}, the authors iterated between 1. RAG and 2. generation-augmented retrieval (GAR) to generate better responses. In FoRAG \cite{Cai_2024}, a two-stage framework is applied for 1. outline generation to ensure coherent and multifaceted responses and 2. answer expansion based on outline to enhance the factuality and logicality of web-search enhanced long-form Q\&A. In RAT \cite{wang2024ratretrievalaugmentedthoughts}, an initial zero-shot CoT response is generated based on the query. For each step in the second round, a new retrieval sub-query is generated based on 1. the query, 2. previous reasoning steps in the second round and 3. the current reasoning step from the zero-shot CoT response in the first round. Next, relevant documents will be retrieved based on the retrieval sub-query in the second round and eventually the current reasoning step in the first round will be revised. After revising all the steps in the first round, the final answer will be generated. In \cite{mombaerts2024metaknowledgeretrievalaugmented}, the authors improved query expansion and document retrieval by incorporating metadata-guided synthetic Q\&A pairs and meta knowledge (MK) summaries through a prepare–rewrite–retrieve–read (PR3) pipeline. To begin with, different documents were classified to predefined categories, and Claude was utilized to extract Q\&A pairs from these documents. Then, MK summaries are generated for metadata-based clusters of QA pairs. When a query comes, relevant MK summaries are retrieved to augment the query into multiple more-specific sub-queries. The augmented sub-queries are applied to retrieve relevant Q\&A pairs and the original document titles. Eventually, the original query, augmented sub-queries, and retrieved Q\&A pairs are provided to the LLM for answer generation.

The O1 embedder \cite{yan2025o1embedderletretrievers} introduces a special token <emb> that allows LLMs to not only embed queries but also generate candidate reasoning paths. The embedder is jointly fine-tuned with contrastive loss for retrieval and next-token prediction for supervised fine-tuning (SFT), enabling it to align reasoning with embedding quality. At inference, the query is expanded into multiple reasoning chains, each paired with the original query and re-encoded by the O1 embedder, where the resulting embeddings are aggregated—via average pooling or voting—into a single representation for retrieval. The enriched query embedding is then passed into the standard RAG pipeline.  

\subsection{Data}
In RAG, the chunking and storage of the data play an important role. Smart chunking and storage strategies will greatly facilitate RAG. In addition, the previous RAG focuses on unstructured data, i.e., text, while other types of structured data like graph, table and tree deserve careful research. The following works deal with these problems.

\subsubsection{Chunking and Storage Strategies}
\textbf{Adaptive Chunking} For RAG with static documents, the chunking and storage of these documents in the database deserve some consideration. The most frequently utilized way to store documents is to split them into chunks, and use the embedding of each chunk as the key and the chunk as the value. To avoid the information loss, there is overlap between each chunk. However, random chunking will cause information loss and performance degradation. In ChunkRAG \cite{singh2025chunkragnovelllmchunkfiltering}, variable-length semantic chunking strategies were utilized to alleviate this problem. To be more specific, a two-stage approach is proposed including 1. variable-length semantic chunking and 2. advanced filtering to reduce hallucination and irrelevant responses in RAG. In variable-length semantic chunking, documents were split into sentences using the NLTK library, the consecutive sentences with large embedding similarity were concatenated to form variable-length chunks with dynamic greedy aggregation and lastly the embeddings of the variable-length chunks were stored in the database. Then, hybrid retrieval strategy with advanced filtering was applied by 1. query rewriting using GPT4o, 2. initial retrieval with term frequency-inverse document frequency (TF-IDF) and embedding cosine similarity,redundancy removal and removal of irrelevant content through LLM scoring, self-reflection and critic model with dynamic threshold, 3. hybrid retrieval by BM25 and LLM scoring, 4. coherence rerank to deal with the "lost-in-the-middle" problem. Eventually, LLM generated responses based on the selected documents.

\textbf{Map Chunking back to Documents} Another way of dealing with information loss in chunking stores the chunks with the mapping back to the whole documents. In \cite{zhao2024longragdualperspectiveretrievalaugmentedgeneration}, LongRAG was proposed to address “lost-in-the-middle” issues in generation \cite{liu2023lostmiddlelanguagemodels} and retrieval noise/irrelevance using dual perspectives: global (whole document) and local (chunk), linked through a mapping from chunk to document. LongRAG consists of four components: (1) a hybrid retriever, (2) an LLM-augmented information extractor, (3) a CoT-guided filter, and (4) an LLM-augmented generator. The hybrid retriever performs coarse retrieval via a dual encoder and fine-grained retrieval via a cross-encoder. The information extractor maps retrieved chunks back to their documents, where an LLM derives global information. The CoT-guided filter classifies chunks as supported or unsupported for the query, combining supported ones as local information. Finally, global and local information are fused to generate the response.

\textbf{Store and Compute the Whole Database} In another extreme, when the database is small enough, the whole database is loaded in memory and its corresponding KV cache is pre-computed and re-utilized to avoid the retrival process and speed the whole process up. In \cite{Chan_2025}, Cache-Augmented Generation (CAG) was proposed to replace RAG for a medium knowledge base to reduce latency and avoid retrieval error. The whole knowledge base is preloaded into the cache through precomputation. During inference, the user’s query is processed conditioned on this cached knowledge, enabling the model to generate responses without additional retrieval. After generation, the KV cache is efficiently reset for subsequent queries. A potential problem with CAG stems from noise or distraction when utilizing the database for answer generation.

\subsubsection{Graph RAG}

\textbf{GraphRAG} 

\begin{figure}
    \centering
    \includegraphics[width=\linewidth]{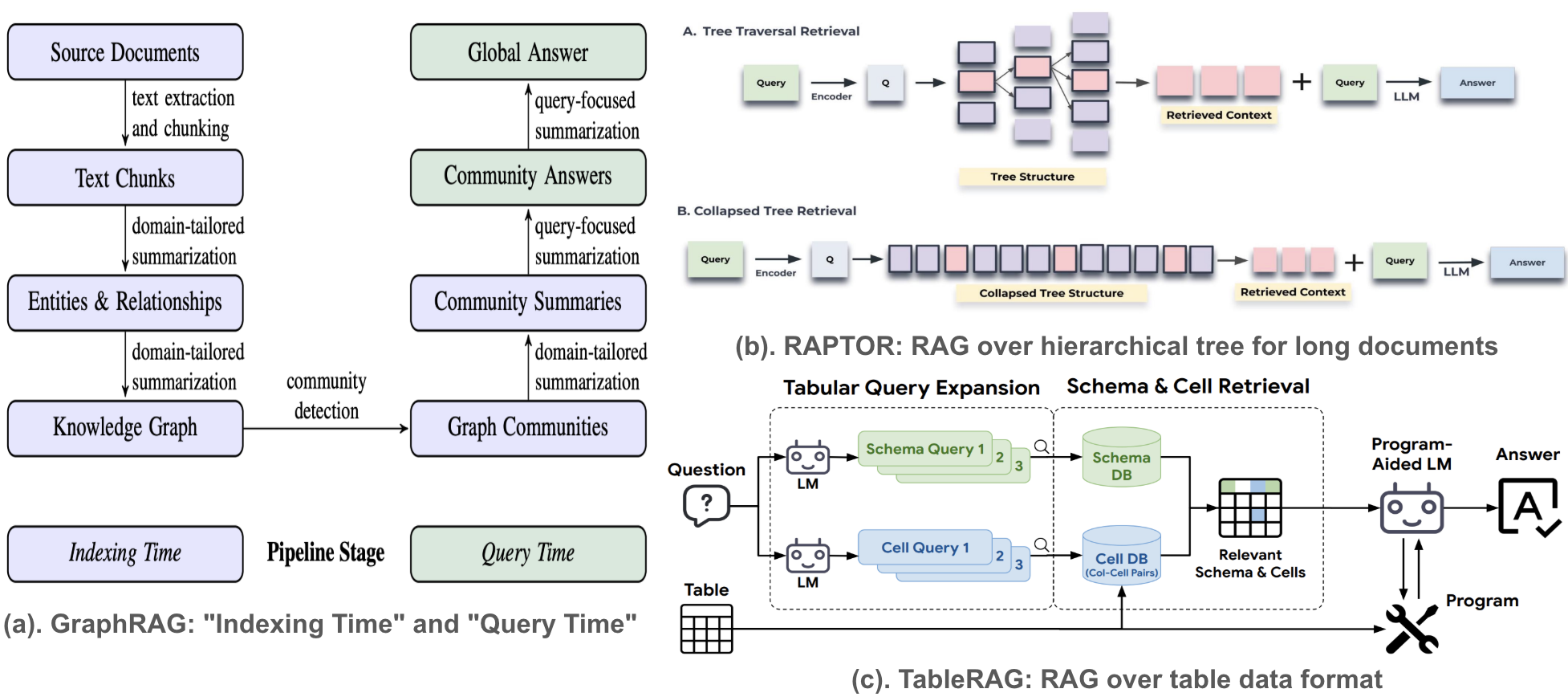}
    \caption{Data: (a). GraphRAG: "Indexing Time" and "Query Time", (b). RAPTOR: transforms long documents into tree with summary and answers query based on the tree, (c). TableRAG: retrieves relevant contents from table and utilizes the content for answer generation}
    \label{fig: GraphRAG}
\end{figure}

RAG excels at answering local questions by retrieving a small set of relevant documents, but it struggles with global questions like summarization (e.g., “What are the main themes in the dataset?”), since the documents are splitted into chunks, retrieval alone cannot capture and synthesize the full breadth of information across the entire corpus and the sequence information of these chunks are lost. To solve these problems, GraphRAG has been proposed \cite{edge2025localglobalgraphrag} as shown in part (a) of Figure \ref{fig: GraphRAG}. It includes two-stages: "Indexing Time" and "Query Time". 

In the "Indexing Time", it is composed of the following stages:
\begin{enumerate}
  \item \textit{Chunking and Extraction}: Source documents are divided into chunks, from which entities (nodes, $N$), relationships (edges, $E$), and associated claims are extracted.
  \item \textit{Knowledge Graph Construction}: Entities and relationships form a graph as triplets $(s, r, o)$ (subject, relationship, object).
  \item \textit{Community Detection}: The Leiden algorithm \cite{traag2019louvain} partitions the graph into sub-graphs (communities) for scalable summarization.
  \item \textit{Summary Generation}: Summaries are created bottom-up. Leaf-level summaries prioritize key nodes, edges, and claims within token limits. Higher-level community summaries use sub-community summaries if token limits are exceeded, ensuring scalability.
\end{enumerate}

In the "Query Time", it is composed of the following stages:
\begin{enumerate}
  \item \textit{Query Processing}: Community summaries are shuffled, divided into chunks, and combined with the query for LLM evaluation of helpfulness.
  \item \textit{Reranking and Selection}: Chunk summaries are reranked by helpfulness score and iteratively selected until the token limit is reached.
  \item \textit{Answer Generation}: Selected chunk summaries are sent to the LLM to produce the final global answer.
\end{enumerate}

\textbf{Modification on GraphRAG}

Later, Some modification works have been done based on GraphRAG with minor differences on graph construction, community detection and LLM reasoning. In CommunityKG-RAG \cite{chang2024communitykgragleveragingcommunitystructures}, the community is detected by Louvain algorithm \cite{Blondel_2008} and the embedding of each community is the average of each node's embedding within the sub-graph extracted from the BERT encoder \cite{devlin2019bertpretrainingdeepbidirectional}. In HippoRAG \cite{gutiérrez2025hipporagneurobiologicallyinspiredlongterm} and \cite{gutiérrez2025ragmemorynonparametriccontinual}, a knowledge graph and a passage–node incidence matrix are constructed through offline indexing. Then, when a query comes, seed entities are extracted and an online relevant entity search is conducted through Personalized PageRank (PPR) where node specialty is applied to modify the weights based on inverse document frequency-like signals for question answering. In addition, hierarchical information is utilized to improve GraphRAG. In tree-RAG (T-RAG) \cite{fatehkia2024traglessonsllmtrenches}, An entity hierarchy tree structure was constructed to represent organizational hierarchies. During inference, if entity is detected from the query, its hierarchical information will be extracted from the tree. If the hierarchical information has been detected, it along with retrieved documents will be sent to LLM for answer generation. GraphRAG can also run together with traditional RAG to obtain more detailed information. In HybridRAG \cite{sarmah2024hybridragintegratingknowledgegraphs}, the traditional RAG, i.e., VectorRAG is combined with GraphRAG to retrieve semantically relevant text chunks and structured relationship-aware sub-graphs of financial data and they are combined and sent to LLM for generation. To accelerate GraphRAG, LightRAG \cite{guo2025lightragsimplefastretrievalaugmented} retrieves only relevant entities and relations via vector search, instead of retrieving large text chunks or doing heavy graph traversal. During inference, LightRAG first extracts local keywords (for entity-level details) and global keywords (for broader themes) from the query. It then uses a vector database to match local keywords with candidate entities and global keywords with higher-level relations. Finally, it incorporates high-order relatedness by retrieving one-hop neighboring nodes and edges from the graph, enriching the results with structural context before sending to LLM for answer generation.

\textbf{Knowledge Graph Construction}

There have been some works to improve the quality of the constructed knowledge graph to reduce repetition and error, utilize information embedded within dataset and update with the latest knowledge or relationship extraction. In iText2KG \cite{lairgi2024itext2kgincrementalknowledgegraphs}, the authors propose a method to construct a knowledge graph to extract entity and relationship progressively while ensuring semantic uniqueness, i.e., no overlapping or duplicate meanings. To begin with, document distiller rewrites documents into semantic blocks in JSON. Then, incremental Entities Extractor extracts and deduplicates entities through cosine similarity in the embedding space, where the documents are processed sequentially. Lastly, semantic blocks and global document entities are sent to relation matcher to extract global document relations and the extracted knowledge graph is saved in graph database Neo4j. In MedGraphRAG \cite{wu2024medicalgraphragsafe}, predefined medical tags was utilized for hierarchical clustering after the initial construction of the knowledge graph. A U-retrieval is also proposed to move top-down to search the most relevant sub-graph and utilize that sub-graph for draft answer generation and then move bottom-up to utilize higher level summary tags to further refine the answer. In Retrieval-Augmented Editing (RAE) \cite{shi2024retrievalenhancedknowledgeeditinglanguage}, incorrect and latest knowledge is updated and inserted to the existing knowledge graph. To begin with, relevant knowledge sub-graph are selected based on maximizing mutual information and form the fact chain. Next, uncertainity-based fact pruning, i.e., entropy is utilized to reduce redundency and sent to LLM for final answer generation. In AgentRE \cite{Shi_2024}, the authors employed a LLM as an agent to interact with three modules: 1. retrieval module, 2. memory module and 3. extraction module to perform relation extraction (RE) in complex tasks, producing structured triples that are used to build a knowledge graph. Firstly, retrieval module store and retrieve relatively static knowledge and it is utilized for sample retrieval and relevant information retrieval, Then, memory module contains shallow memory to summarize extraction result and deep memory to summarize and reflect on historical actions. Lastly, extraction module turns text into structured triplets using ReAct style following 1. thought, 2. action and 3. observation cycle until it has enough information to generate the final response \cite{yao2023reactsynergizingreasoningacting}. 

\textbf{Iterative GraphRAG}

The following works focus on enhance the reasoning capability of LLM in GraphRAG by iteratively breaking down the query into sub-queries, retrieving relevant documents for the sub-query and generating answers based on retrieved documents. In ChainRAG \cite{zhu2025mitigatinglostinretrievalproblemsretrieval}, it mimics human-like reasoning 1. decomposing the complex question, 2. retrieving relevant context via sentence graph, 3. rewriting vague sub-questions and 4. synthesizing the final answer. To begin with, the sentence graph with entity indexing is constructed. The complex question is then broken down into multiple simpler ones, with vague pronouns such as “it” or “this” replaced by concrete terms from previous sub-queries or answers to prevent confusion. Each sub-question is retrieved with a two-step strategy with the first level seed sentence retrieval and the second level to explore the neighbor of the seed sentences until the context is sufficient or the maximum context window is reached. The retrieved documents are then utilized for answering the sub-query. Finally, the retrieved sub-documents will be reranked at sentence level and retrieved and integrate with all sub-question to generate the final answer. In \cite{ma2025thinkongraph20deepfaithful}, think-on-graph (ToG-2) iteratively reasons between 1. context-enhanced graph search to explore the neighborhood of the current entities and prune unrelated entities and 2. knowledge-guided context retrieval to retrieve related documents, build context pool, rank by relevancy and select the top-k chunks during the reasoning process to generate the final answer. After each iteration, the accumulated knowledge will be checked whether it is enough for answering the question or conducting another round of iteration. In knowledge graph based retrieval-augmented reasoning (KR-RAR) \cite{wu2025graphaugmentedreasoningevolvingstepbystep}, process level reasoning capability of GraphRAG is explored. To begin with, a process-oriented mathematical knowledge graph is constructed from PRM800 dataset to enable multi-step reasoning. 

Next, a hierarchical retrieval strategy is used where (a). step-level problem retrieval is realized by 1. filtering, 2. semantic similarity scoring and 3. context retrieval using depth-first search (DFS) and (b). step retrieval is realized by 1. restricting search to the steps of retrieved problems, 2. semantic similarity scoring and 3. context retrieval using breadth-first search (BFS). After every step of iteration, both step verification and end-of-reasoning detection will be conducted and retrieved documents will be refined by LLMs until the termination condition is met. In \cite{liang2024kagboostingllmsprofessional}, Knowledge Augmented Generation (KAG) boosts LLM reasoning in logic, computation, and semantics via three modules: KAG-Builder, KAG-Solver, and KAG-Model. KAG-Builder constructs a LLM-friendly knowledge framework with raw chunks, graph information, and a knowledge layer linked via mutual indexing. KAG-Solver uses logical-form-guided hybrid reasoning, combining LLM, knowledge, and mathematical logic reasoning iteratively. KAG-Model fine-tunes the system to improve knowledge representation and retrieval accuracy.

\textbf{Adaptive GraphRAG to Improve Performances}

The following works train a separate classifier to identify the complexity of the query or the relevancy of a triplet to a query and apply different strategies based on the classifier's result. In flexible modular KG-RAG framework (FRAG) \cite{gao2025fragflexiblemodularframework}, three modules: reasoning-aware module, flexible-retrieval module and reasoning module are included to adapt the retrieval strategy based on the complexity of the query. Reasoning-aware module is utilized to classify the query as simple or complex based on the estimated minimum hop count of reasoning paths in a knowledge graph through a trained binary classifier. Flexible-retrieval module includes a preprocessing-retrieval-postprocessing pipeline where preprocessing obtains the knowledge sub-graph, retrieval utilizes different strategies to derive reasoning paths based on query complexity and postprocessing removes redundant and irrevalent reasoning paths through Path Ranking Model. If the query is simple, breadth-first search will be utilized to retrieve all reasoning paths, while Dijkstra algorithm will be applied to find the shortest reasoning path when the query is complex. Lastly, reasoning module generates the final answer based on query and selected reasoning paths. In \cite{li2025simpleeffectiverolesgraphs}, Sub-graphRAG represents the graph as triplets, and directional distance encoding (DDE) is utilized to compute the relationship between the query and the triplet and eventually, the query, triple and DDE is sent to a pretrained MLP for classifying if the triple is relevant to the query improve the efficiency and effectivenss of GraphRAG. 

The retrieved knowledge sub-graph is generally not suitable for directly input to LLM for generation. The following papers discuss how to deal with the problem to make the knowledge sub-graph more suitable for LLM. In CoTKR \cite{wu2025cotkrchainofthoughtenhancedknowledge}, knowledge rewriting is incorporated to transform question-related triples into natural language that is consumed by LLMs. When a knowledge sub-graph is retrieved, it is transformed into contextual knowledge by iterations between 1. reasoning: decomposing the question and generating a reasoning trace and 2. summarization: summarizing the relevant knowledge. Lastly, the reasoning trace and summaries are sent to LLM for generating answers. In GRAG \cite{hu2025graggraphretrievalaugmentedgeneration}, the knowledge sub-graph is transformed to hierarchical text description and use GNN for embedding extraction to faciliate its performance on LLM. To begin with, K-hop ego-graph of all nodes are extracted and based on the query, the top K relevant sub-graphs are retrieved, merged and pruned using soft pruning using MLP-learned scaling factors based on their distance to the query. Lastly, the merged and filtered graph will be sent to LLM for answer generation with 1. text view (hard prompt) where the graph is transformed into a hierarchical text description and 2. graph view (soft prompt) where the GNN generates its embedding in the prompt. In GraphReader \cite{li2024graphreaderbuildinggraphbasedagent}, knowledge graph is utilized to handle limited context window of LLM by transforming the original long documents into knowledge graph and corresponding atomic facts summarized by the LLM. Then, a notebook to record supporting facts and a rational plan to identify missing information and the logic to identify them are constructed. Next, exploration are guided by the rational plan and recorded by the notebook at three levels: atomic fact for coarse information, chunks for detailed information and neighbors until enough information is collected to answer the query. 

\subsubsection{Other Data Representations}

In this subsection, different types of structured data—including trees, tables, and metadata—will be discussed in the context of RAG. In Recursive Abstractive Processing for Tree-Organized Retrieval (RAPTOR) \cite{sarthi2024raptorrecursiveabstractiveprocessing}, the authors enhanced RAG by structuring long documents into a hierarchical tree. The tree is built by chunking documents, embedding through SentenceBERT, clustering through Gaussian Mixture Models (GMM) and dimension reduction through Uniform Manifold Approximation and Projection (UMAP) and summarizing each chunk. During inference, the relevant documents are queried through either tree traversal or collapsed tree as shown in part (b) of Figure \ref{fig: GraphRAG}. In tree traversal, it is started from the root node and traverse from top to bottom while keeping the top-k documents. For collapsed tree, the tree is flatten into a single layer, and the top-k item is retrieved. In \cite{chen2024tableragmilliontokentableunderstanding}, the authors proposed TableRAG to improve the performances of RAG for table format. Generally, a table is too large to put into the context window of a LLM. To begin with, tabular query expansion is applied to obtain schema queries to retrieve important column-level information and cell queries to retrieve key cell values by prioritizing distinct and frequent values within a cell encoding budget. With encoder, the schema queries and cell queries are transformed into embedding and the top-k entries are selected. Lastly, a program-aided solver inspired by the ReAct framework is applied to reason and generate executable programs—e.g., SQL or Python—by leveraging the retrieved schemas and cells for answer generation. In \cite{Tan_2025}, HtmlRAG was proposed to handle long HTML contexts while maintaining their important information through utilizing its semantic and structural information. HtmlRAG included four steps in the pipeline: 1. html cleaning to remove unnecessary css and js information, 2. block tree construction to transfrom the html into a tree structure where tiny nodes were merged to reduce compute cost, 3. embedding-based coarse block pruning to remove the irrevalent nodes of the tree in the embedding space and 4. generative fine-grained pruning, where a path-aware generative model scores fine-grained block paths to select the most relevant content. After the block tree is constructed, cleaned and pruned, it will be tranformed back to html format and formatted as prompt to send to LLM for answer generation. In Multi-Meta-RAG \cite{Poliakov_2025}, metadata such as article sources and publication dates is first extracted from user queries by a lightweight LLM using few-shot prompting. This metadata is then used as a filter during chunk retrieval in the vector database, ensuring that only relevant sources and dates are considered before RAG.

\subsection{Retrieval and Reranker}

In this section, the retriever and reranker will be discussed. Given the query, how to retrieve the most relevant documents to improve downstream questions answering will be the first question. Then, there may be some errors in the documents or inconsistency with LLM's internal knowledge. Lastly, is it possible to use cross-encoder to further improve the relevancy of the retrieved documents to the query? The following sections will discuss these questions.

\subsubsection{Retrieval Refinement}
\begin{figure}
    \centering
    \includegraphics[width=\linewidth]{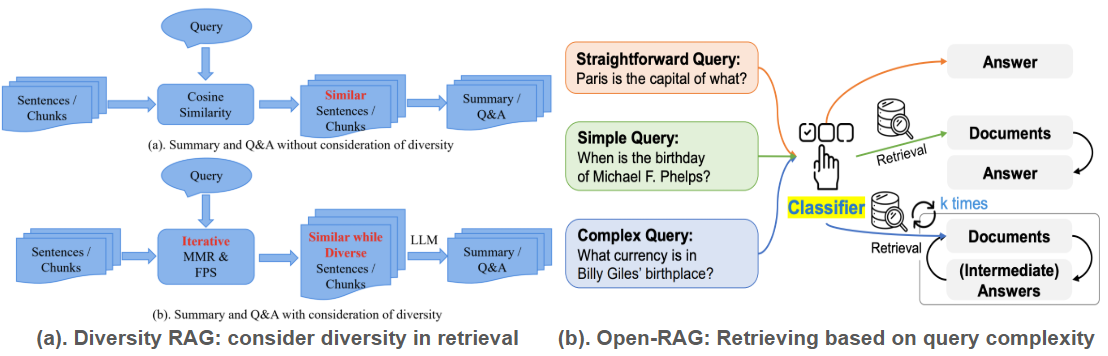}
    \caption{(a) Dvierse RAG: retrieve relevant and diverse documents; (b) The intuition behind Adaptive-RAG and Open-RAG: determine whether to retrieve based on the complexity of the query}
    \label{fig: DiverseRAG}
\end{figure}

\textbf{Retrieval Considering Diversity}

In the normal retrieval process, only paying attention to the relevance between query and documents will cause redundency and miss important materials. In DiversityRAG \cite{wang2025diversityenhancesllmsperformance}, maximal marginal relevance (MMR) and modified farthest point sampling (mFPS) were applied to retrieve documents by balancing relevance and diversity and improve the answer accuracy in downstream tasks as shown in part (a) of Figure \ref{fig: DiverseRAG}. The formula of MMR is in Eq. \ref{eq:diversityrag-mmr} where $k$ refers to the $k$-th selected document, $r_i$ is the relevancy between the query and the $i$-th document and $\max_{j \in W} \cos(i, j)$ measures the similarity between the current document and previously selected documents in $W$ to contribute diversity in a greedy manner.

\begin{equation}
k = \arg\max_{i \in R} \left[ \alpha \cdot r_i - (1 - \alpha) \cdot \max_{j \in W} \cos(i, j) \right]
\label{eq:diversityrag-mmr}
\end{equation}

\textbf{Retrieval based on Prompt Complexity}

Another direction of retrieval refinement is to retrieve based on the complexity of query. In Adaptive-RAG \cite{jeong2024adaptiveraglearningadaptretrievalaugmented}, the authors proposed to utilize different retrieval strategy for different prompt complexity to balance efficiency and accuracy as shown in part (b) of Figure \ref{fig: DiverseRAG}. They trained a classifier using cross-entropy loss to classify the complexity of the prompt. When the prompt was simple, it directly generated response without retrieval. If the prompt is moderate, it will generate response with one time retrieval. If the prompt is really complex, it will iteratively generated responses with multi-hop retrieval. In Open-RAG \cite{islam2024openragenhancedretrievalaugmentedreasoning}, [RT] or [NoRT] tokens were predicted for deciding whether to apply retrieval. If [NoRT] is predicted, it will first generate a response without a retrieval and compute the confidence score. If the confidence score is below a threshold, it will retrive relevant knowledge. If [RT] is predicted, relevant knowledge will be retrieved at the beginning. In unified active retrieval (UAR) \cite{cheng2024unifiedactiveretrievalretrieval}, the authors proposed a unified active retrieval where active retrieval refers to automatically determine when to retrieve including 1. Self-aware classifier, 2. Time-aware classifier, 3. Knowledge-aware classifier and 4. Intent-aware classifier. When a query came, these following steps were executed sequentially. To begin with, intent-aware classifier check if the user has explicitly ask for retrieval, and retrieval if true otherwise proceed. Then, the knowledge-aware classifier checks if the query requires factual knowledge. If true , retrieve otherwise proceed to the check. Next, the time-aware classifier checks if the query is time-sensitive. If so, retrieve otherwise proceed to the next check. Lastly, the self-aware check is conducted to check if the LLM has already known the answer. If so, proceed otherwise retrieve.

\textbf{Hierarchical Retrieval}

Next, hierarchical retrievers are applied to retrieve information from different granualities like chunks and documents. In Hierarchical RAG with rethink (HiRAG) \cite{zhang2024hierarchicalretrievalaugmentedgenerationmodel}, hierarchical retriever including document-level sparse retrival and chunk-lever dense retrieval is utilized to select single candidate rather than the top-k multi-candidates to reduce the noise of irrevalent documents and the pressure of context window limitation. HiRAG contained five components: 1. decomposer to decompose complex question into simpler sub-questions, 2. definer to determine if current sub-answers can solve the original query, 3. hierarchical retriever including document-level sparse retrieval and chunk-level dense retrieval to select single candidate rather than the top-k multi-candidates, 4. verifier to check if the current retrieval is sufficient for answer sub-questions, which will trigger rethinking if the retrieval is not sufficient and 5. summarier to combine all sub-answers into the final answer. By the combination of hierarchical retriever, single-candidate selection and rethinking process, HiRAG outperformed on multi-hop question answering tasks. Another work on hierarchical retriever is in \cite{wu2025graphaugmentedreasoningevolvingstepbystep} and it has been discussed in GraphRAG.

\subsubsection{Retrieved Document Refinement}

After retrieval, there may exist conflicts between internal and external knowledge. The conflicts between internal and external knowledge should be resolved firstly before generating the final answer. In FilCo \cite{wang2023learningfiltercontextretrievalaugmented}, the authors trained a filter model $M_{ctx}$ to extract the text span, i.e., sentence from the retrieved document and send the text span to LLM for final answer generation. In CRAG \cite{yan2024correctiveretrievalaugmentedgeneration}, the authors proposed to correct the problem in the retrieved documents to improve the performance of RAG. CRAG utilized a light-weight evaluator to judge the relevance score of the retrieved document as 1. correct, 2. incorrect and 3. ambiguous. If the judgment is correct, the retrieved document will be refined by splitting into smaller chunks, irrelevant chunks will be filtered out and the remaining chunks will be recomposed and then be applied for output generation. If the judgment is incorrect, the retrieved document will be discarded and will rely on web search on rewritten queries to get correct information. If the judgment is ambiguous, both document refinement and web search on rewritten queries will be conducted for the final generation. In ASTUTE RAG \cite{wang2025astuteragovercomingimperfect}, the authors desired to solve the imperfect retrieval and conflicts between internal and external knowledge. To begin with, the LLM is elicited to generate accurate, relevant and hallucination-free passages based on the query where the number of generated messages is determined by the LLM. Then, the retrieved external knowledge, internal knowledge and source labels are consolidated by iteratively comparing between them, identifying consistencies and conflicts, discarding irrelevant information and regroup the remaining contents based on consistency. Eventually, the most trustworthy, source-creditable documents are sent to LLM for final answers generation and the best answer was further selected based on reliability. In \cite{liu2024lighterbetterflexiblecontext}, FlexRAG was proposed to compress the retrieved context to improve the performance of RAG on long context problem. The retrieved context is firstly sent to a compressive encoder, a middle layer of an LLM for compatibility, to extract embedding. Then, selective compression mechanism through 1. an importance estimation at token or sentence level and 2. an embedding down-sampling are applied to extract condensed embedding with a flexible compression ratio. Lastly, query is tokenized, represented in embeddings and concatenated with the extracted condensed embedding for final answer generation.

\subsubsection{Reranker}

The retrieval of documents assessed relevancy in a fairly coarse way—usually. Reranker improves this by using a cross-encoder: the query and each retrieved document are jointly processed to produce a more precise relevance score rather than cosine similarity between embedding the query and the document. In this way, the most relevant documents rise to the top, giving a finer-grained ordering than the initial retrieval. In addition, the layout of these documents in the context plays an important role on correctly answering the query as shown in "lost-in-th-middle" \cite{liu2023lostmiddlelanguagemodels}. As a result, after the retrieval of the top-K documents, it is desired to rerank these documents for downstream generation task. 

\begin{figure}
    \centering
    \includegraphics[width=\linewidth]{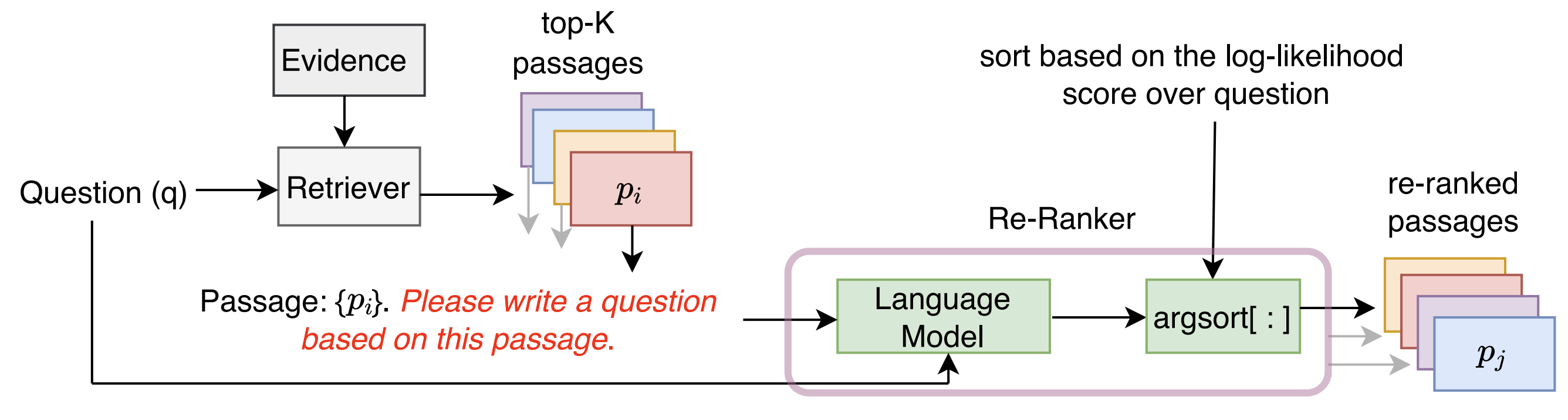}
    \caption{Rerank to select the most important document for a given query}
    \label{fig: rerank}
\end{figure}

In \cite{sachan2023improvingpassageretrievalzeroshot}, each document was utilized to predict the probability of the query, and the document with the highest probability to predict the query will be more important as shown in Figure \ref{fig: rerank}. By Bayes's law, $log P(z_i|x) \approx log P(x|z_i)$ given a uniform $P(z_i)$ assumption. Thus, the scoring mechanism is $r(x, z_i)=\frac{1}{|x|}\sum_{t} logP(x_t|x_{<t}, z_i)$ which represents the probability of LLM generating the query based on the retrieved document, and it is applied to rank the importance of different documents before sending to LLM for generation. In RankRAG \cite{yu2024rankragunifyingcontextranking}, a prompt "For the question {question}, access whether the passage is relevant to the question." is utilized, and the probability of outputting "True" is utilized as metric for ranking documents. Lastly, the reranked documents are sent to LLM for final answer generation. In RE-RAG \cite{kim2024reragimprovingopendomainqa}, the authors enhance the RAG framework by introducing a Relevance Estimator (RE) that evaluates the relevance of retrieved contexts to a given question. RE outputs a confidence score for each context, which is used to rerank contexts and guide the answer generation process.

\subsection{LLM Generation}

In this subsection, the process that LLM utilized retrieved documents for final answer generation will be investigated with the focus on multi-step RAG that decompose the query into multiple sub-queries and answer them iteratively.

\begin{figure}
    \centering
    \includegraphics[width=\linewidth]{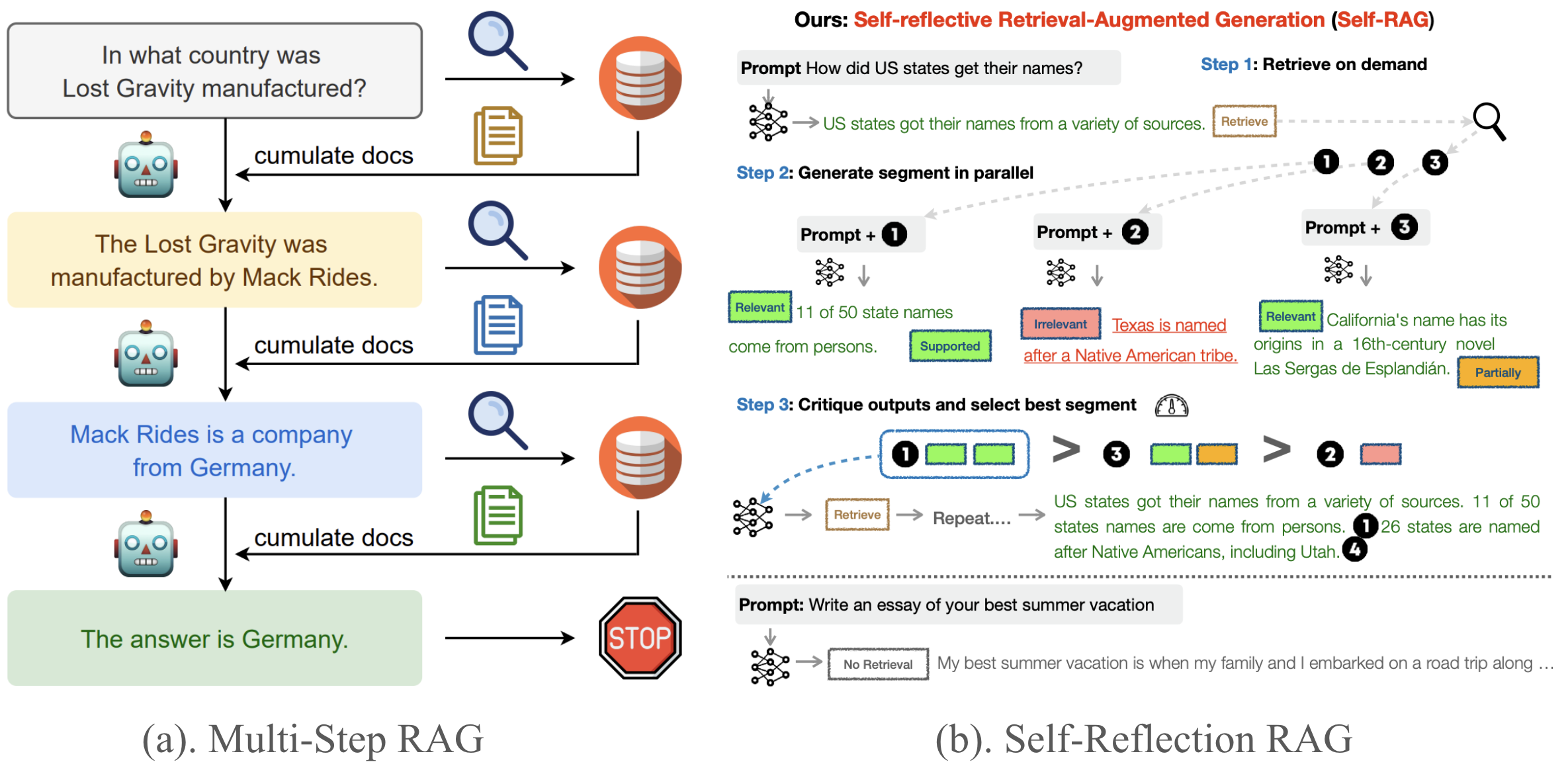}
    \caption{(a). Multi-Step RAG to divide the query into sub-queries and answer them sequentially based on previous sub-query's answer, (b). Self-Reflection RAG to classify retrieved documents as relevant, irrelevant or partially}
    \label{fig:MultiStep_SelfReflection}
\end{figure}

\begin{table}[h]
\centering
\caption{Summary of Important Works in RAG}
\label{tab:iterative_rag}
\begin{tabular}{p{2.5cm}|p{13cm}} 
\hline
\textbf{Method} & \textbf{Key Mechanism} \\
\hline
IRCoT \cite{trivedi2023interleavingretrievalchainofthoughtreasoning} 
& Interleaves retrieval and reasoning steps until final answer or maximum iterations reached. \\
\hline
Search-o1 \cite{li2025searcho1agenticsearchenhancedlarge} 
& Agentic RAG with special tokens for query injection; employs two-stage document refinement. \\
\hline
SELF-RAG \cite{asai2023selfraglearningretrievegenerate} 
& Fine-tuned generator with Retrieve, ISREL, ISSUP, ISUSE tokens; evaluates document relevance and response quality. \\
\hline
Active RAG \cite{jiang2023activeretrievalaugmentedgeneration} 
& Forward-looking mechanism; evaluates provisional responses and triggers retrieval when confidence is low. \\
\hline
\end{tabular}
\end{table}

\subsubsection{Multi-step RAG}

In the muti-step Q\&A, it is difficult to retrieve all relevant documents within one retrieval process. For example, for the question, "Where is the author of Harry Porter born?", it needs to 1. firstly identify the author of "Harry Porter" is J. K. Rowling and 2. then identify J. K. Rowling is born in Yate, United Kingdom. For such kind of queries, it is supposed to firstly split the muti-step question into multiple steps during the reasoning process where each reasoning step identify a sub-question that needs retrieval to be answered, a retrieval process is conducted based on the sub-query and the retrieved documents are utilized to answer this sub-query and move onto the next reasoning step. In different literature, multi-step Q\&A can also be named muti-hop Q\&A or muti-aspect Q\&A, while the essence of these problems are the same. The summary of some important works on multi-step RAG is in Table. \ref{tab:iterative_rag}.

Interleaved Retrieval and Chain-of-Thought (IRCoT) \cite{trivedi2023interleavingretrievalchainofthoughtreasoning}, as shown in part (a) of Figure \ref{fig:MultiStep_SelfReflection}, proposes to interleave between
\begin{itemize}
    \item \textit{Retrieval}: retrieving more documents based on the previous reasoning steps.
    \item \textit{Reasoning}: generating the next reasoning step.
\end{itemize}
Iterations of IRCoT continued until a final answer or the maximum number of iterations was reached. In Search-o1 \cite{li2025searcho1agenticsearchenhancedlarge}, it iteratively included 1. agentic RAG mechanism to automatically determine when to start query and inject knowledge with special tokens like $<|begin\_search\_query|> query\_text <|end\_search\_query|>$ and $<|begin\_search\_result|> docs <|end\_search\_result|>$ and 2. reason-in-documents module through a two-stage refinement by firstly generating an intermediate document analysis and then distilling key insights into a concise summary to avoid noisy or verbose insertions for the reasoning process. 

In previous Iterative RAG works, the decision of the retrieval query is a black box, and inner monologues RAG (IM-RAG) \cite{yang2024imragmultiroundretrievalaugmentedgeneration} makes the retrieval decision intrepetible, modular and optimizable through inner monologues. There are four modules: 1. reasoner including questioner and answerer, 2. retriever, 3. refiner and 4. progress tracker. To begin with, a query is sent to the reasoner to evaluate if more evidence is needed, and if yes, the reasoner will play the role of questioner to generate a query to retrieve more information from external knowledge source. Then, the retriever will retrieve relevant knowledge from external source and the refiner will re-rank and filter more relevant documents. Progress tacker evaluates how close the retrieved evidence supports the correct answer. Lastly, the reasoner evaluates and reasons why the evidence is sufficient or not. If the evidence is not sufficient, it will go back to the retriever with a new sub-query and its corresponding reasoning to retrieve more documents. Otherwise, the reasoner will play the role of answerer to generate the final answer. 

Previous works divided the query into sub-queries, rewrote them into higher quality based on dense retriever's embedding and then utilized for the embedding for retrieval and reasoning for the next iteration. However, this workflow of tight coupling between query rewriting and the dense retriever limits its compatibility with hybrid retrieval, impeding further RAG performance improvements. To solve this problem, LevelRAG \cite{zhang2025levelragenhancingretrievalaugmentedgeneration} proposed a two-tiered search architecture with 1. High-level searcher and 2. Low-level searchers for decoupling the query rewriting process from the retrieval and reasoning process. The high-level searcher performs four key operations for query rewriting: 1. decompose: breaks the complex query into simpler atomic sub-queries, 2. summarize: condenses retrieved documents into direct answers to each sub-query, 3. verify: checks if collected information sufficiently answers the original query and 4. supplement: adds new sub-queries if initial results are insufficient to break down the complex query into simpler ones. The low-level searcher for hybrid retrieval contained 1. sparse searcher: conducts keyword-based retrieval with Lucene syntax for query rewriting, 2. dense searcher: focuses on semantic similarity and 3. web Searcher: Interfaces with commercial search engines to refine the query for optimal retrieval. 

Lastly, is it possible to retrieve all relevant documents even if the query is muti-hop? Alignment-oriented retrieval (ARM) \cite{chen2025retrieveoncearmalignmentoriented} solved this problem using a three-step methodology: indexing, alignment (information + structure), and self-verification. The indexing process organized the data with embedding and N grams for search. In alignment, it contained information alignment and structure alignment. In information alignment, keywords were extracted by LLM and constrained decoding rephrased the keywords into N grams in the database by scoring them based on average logit probability. Next, the matched keywords were utilized to retrieve relevant documents based on BM25 and embedding similarity. For structure alignment, mixed-integer linear program (MIP) is applied to select different 1. the top-k subset and 2. their connection to form multiple drafts for maximizing relevancy to the question and compatibility between selected subsets. In the third process, the drafts are verified by LLM to get rid of irrelevant retrieved documents. Lastly, beam search is conducted and the best answer is voted by weighted aggregation. 

\subsubsection{Self-Reflection RAG}

In muti-step RAG, the query is divided into sub-query, and each sub-query is utilized to retrieve relevant documents and sent to LLM for generation. However, there is no judgment whether each step is reasonable and the error in each step is accumulated and eventually gets a wrong answer. In SELF-RAG, the generator is fine-tuned to generate four types of tokens: Retrieve (decides when to retrieve), ISREL (check if retrieved documents provide useful information to solve the query), ISSUP (check if the response is supported by the retrieved document), and ISSUE (check if the response is useful for the query) \cite{asai2023selfraglearningretrievegenerate}. As shown in Figure \ref{fig:MultiStep_SelfReflection}(b), given a query $x$, the retrieved document $z$ and the previous responses $y_t$, the generator firstly generates a "Retrieve" token. If "Retrieve" is no, the response is generated directly. Otherwise, $K$ documents will be retrieved and for each document, a response will generated, i.e., $K$ responses in total. For each retrieved document and generated response, it will be evaluated using a score function from the perspectives of "ISREL", "ISSUP", and "ISUSE". The best top-$B$ of the top $K$ responses by the scoring function will be utilized and continue for the next step. Similar ideas are in \cite{zhou2024metacognitiveretrievalaugmentedlargelanguage} when the similarity of the generated response with an expert response lies below a threshold, the monitoring module will activate the evaluating module to check if 1. the knowledge is sufficient, 2. is there any conflicting knowledge and 3. is there any erroneous reasoning. If not enough knowledge is available, a targeted new query will be generated to retrieve again. If the knowledge is conflicting, the correct information, either internal or external knowledge will be utilized. If the reasoning is wrong, the reasoning process will be checked for each step to see if it is supported and give suggestions to the error for the next round of reasoning. RetroRAG \cite{xiao2025retrievalaugmentedgenerationevidenceretroactivity} constructed a graph-based framework including two modules: 1. evidence-collation-discovery (ELLERY) and 2. Answer to discard erroneous past evidence and include new evidence. The ELLERY framework operates through an iterative process involving two core modules: 1. evidence collation and 2. evidence discovery. In evidence collation, missing documents are retrieved based on prior queries, merged with the current set of retrieved documents, and then scored by an LLM where the top-k documents are retained to ensure relevance and quality. In evidence discovery, the LLM derives new inferential evidence based on the collected source evidence, and the inferential envidence needs to pass question-relevance and reference-attribution checks. This cycle continues until the Answer module can produce responses that meet self-consistency criteria, with consistency measured by an LLM-based evaluator.

\begin{figure}
    \centering
    \includegraphics[width=\linewidth]{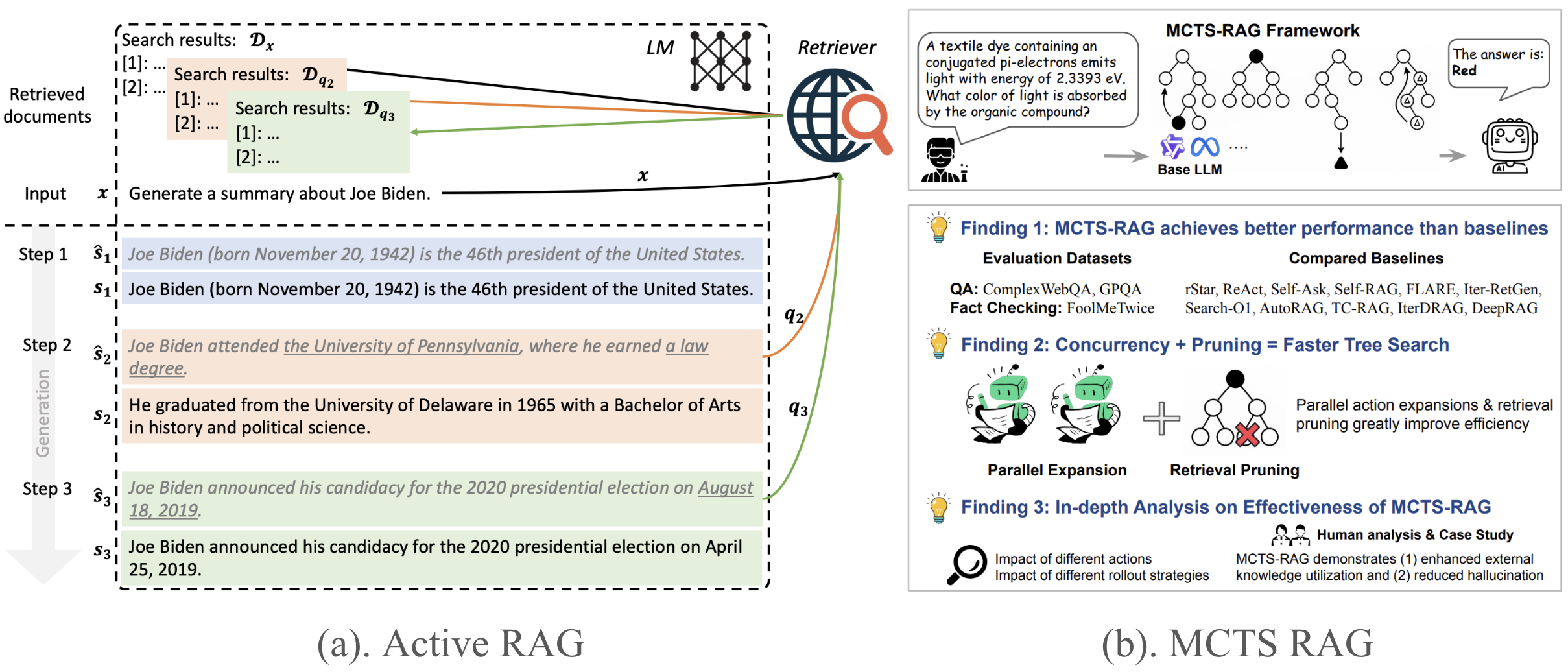}
    \caption{(a). Active RAG: retrieve based on generation that has low probability, (b). MCTS RAG: model the generator through MCTS with the capability of retrieval and reasoning.}
    \label{fig: Active_MCTS}
\end{figure}

\subsubsection{Multi-step RAG with Planning}

In multi-step RAG, the previous works focus on decomposing the query into sub-queries and answer them step by step. Another methodology introduces an explicit planning module and utilizes the initial plan to guide the retrieval and generation process for the final answer. In PlanRAG \cite{lee2024planragplanthenretrievalaugmentedgeneration}, the LLM takes a decision-making question Q, a database D with schema S, and business rules R as input, and generates an initial plan describing the data analyses needed for decision making. Based on this plan, the LLM generates structured queries to retrieve relevant data from the database. After executing these queries, the model evaluates whether the retrieved data are sufficient to answer the question. If not, a replanning procedure iteratively updates the plan until a confident decision is made. In Plan*RAG \cite{verma2025planragefficienttesttimeplanning}, the original workflow of RAG is modified into a two-phase execution: 1. plan generation by decomposing the main query into a direct acyclic graph (DAG) of sub-queries which are stored out of the context window of LLM and 2. plan execution with retrieval where independent nodes in the DAG are executed in parallel and the answers from the parent nodes are passed to the child nodes. Similar ideas are in \cite{lee2025rearagknowledgeguidedreasoningenhances}. In \cite{xiong2025outliningheterogeneousrecursiveplanning}, the authors propose a heterogeneous recursive planning framework for long-form writing with language models. Starting from the top-level writing goal, the planning framework recursively decomposes the goal into three type of tasks: 1. retrieval for information gathering, 2. reasoning for extracting information and resolving logical issues and 3. composition to generate or revise text. The planning process interleaves with the reasoning process with the memory and state updated until all goals are reached. 

Another direction research is applied planning to think ahead, i.e., guessing what LLM needs to in the future steps of generation. In Active RAG \cite{jiang2023activeretrievalaugmentedgeneration}, the authors enhance retrieval-augmented generation by introducing a forward-looking mechanism with three key components: (1) generating provisional responses before producing the final output, (2) evaluating the quality of these provisional responses, and (3) triggering retrieval when confidence in the provisional responses is low as shown in Figure \ref{fig: Active_MCTS}(a). To implement this idea, they propose two variants of Forward-Looking Active Retrieval-Augmented Generation (FLARE): FLAREinstruct and FLAREdirect. FLAREinstruct makes retrieval decisions explicitly by embedding search calls in the generated text.  However, this approach can suffer from unreliable or poorly formed queries. FLAREdirect instead adopts a more implicit strategy, leveraging the language model’s confidence to decide whether retrieval is necessary. To begin with, an initial answer is generated without retrieval. If the model’s probability for the next sentence is sufficiently high, the sentence is accepted directly without retrieval. If confidence is low, the uncertain sentence—or its refined version—is converted into a query, which is then used to retrieve supporting evidence before generating the final response. 

In the previous work of multi-step RAG and multi-step RAG with planning, the performances of RAG will be improved. However, two more questions may come into being: 1. context overload as the retrieved documents overwhelm the LLM's context window and 2. over-planning as continuing planning even if information is enough and redundant planning as new sub-queries will be very similar to previous sub-qeury leading to repetition problems. To solve this problem, the authors proposed to utilize a dual-function query-focused summarizer after retriever to record 1. global evidence memory to record documents relevant to the initial query and 2. local pathway memory to record documents relevant to the current sub-query in \cite{jiang2025retrievesummarizeplanadvancing}. Through the summarizer, the overwhelming context length problem is solved and with the memory, the repetition problem of sub-query is alleviated.

\subsubsection{Tree-based Search for RAG}

\textbf{Tree based Search for RAG}

Tree-based methods introduce explicit search structures into the reasoning process, allowing RAG systems to explore, evaluate, and refine multiple reasoning paths before reaching a final answer. In \cite{li2025deepsolutionboostingcomplexengineering}, SolutionRAG explored the reasoning path by alternating between solution node and comment node which mimic CoT for solution review, i.e., bi-point thinking in a tree structure to improve reliability. Because of the tree structure, different reasoning paths are explored and low quality paths are pruned through 1. reliability score to measure how well a solution solves a technical challenge and 2. helpfulness score to measure the relevance of a feedback comment. Similarly, in ReARTeR \cite{sun2025rearterretrievalaugmentedreasoningtrustworthy}, the authors included process reward model and process explaination model to improve the performance of RAG. In each step, the reasoning model generates multiple reasoning steps, the process reward model provides reward for every response. If any reward is above a threshold, it will be accepted. Otherwise, the process explaination model will provide explaination for the generator to refine its responses.

\textbf{Critic-based MDP for RAG}

By modeling the decision process as a Markov decision process (MDP), we can include search in the inference stage by either generating multiple responses and select the best one using critic based method or balance exploration and exploitation using MCTS. 

To begin with, the best response is selected using critic based method. In \cite{li2024elicitreasoningllmscriticguided}, CRPlanner models the planning as a MDP process where retrieval is a candidate action and multiple execution paths are generated and the best one is selected by a critic model. In the MDP, states refers to the full history of actions and observations. Actions include sub-goal selection and execution by LLM. To begin with, a sub-goal is selected like "RETRIEVE", "REASON" and "GENQUERY" by a sub-goal selection model. Then, the sub-goal will be sent to LLM to generate multiple possible execution paths and a critic model will evaluate the paths that most likely to reach the selected sub-goal as the next action. The workflow of sub-goal selection, execution, and critic with retrieval iterates until reaching out the final answer. 

\textbf{MCTS for RAG}

The other direction leverages MCTS to search for the optimal reasoning path, where retrieval is incorporated in the action space. To understand this line of work, it is helpful to firstly introduce Self-Play muTuAl Reasoning (rStar) \cite{qi2024mutualreasoningmakessmaller}. In rStar, a generator-discriminator framework using MCTS is proposed, while it does not involve retrieval. For the generator, it is realized through MCTS with five action options: 1. propose an one-step thought, 2. propose the remaining thought steps, 3. propose next sub-question along with its answer, 4. answer the sub-question again and 5. rephrase the question/sub-question. The discriminator utilizes another LLM to start from an intermediate reasoning path and see if it can reach the same answer to evaluate the reasoning path.

Building upon rStar, subsequent works extended MCTS to RAG. In MCTS-RAG \cite{hu2025mctsragenhancingretrievalaugmentedgeneration} shown in Figure \ref{fig: Active_MCTS}(b), the generator is modeled as MCTS with six actions including 1. direct answer, 2. quick reasoning, 3. decompose question from rStar, 4. retrieval reasoning, 5. retrieval decompose and 6. summarized answer, allowing dynamic interleaving of reasoning and retrieval. The retrieval process includes four steps: 1. query generation, 2. query execution, 3. knowledge reflection and 4. summary answer. Lastly, candidate answers from multiple reasoning paths are scored based on accumulated rewards along their trajectories and selected via a voting mechanism, ensuring consistency and factual accuracy rather than a discriminator. In RARE \cite{tran2025rareretrievalaugmentedreasoningenhancement}, two more actions are included: 6. search query generation and information retrieval and 7. sub-question retrieval and re-answering based on the five actions in rStar for MCTS to incorporate the retrieval capability. After the generation of different solution paths, retrieval-augmented factuality scorer (RAFS) evaluates each statement of each path against retrieved evidence, assigning Supported/Not Supported labels and a factuality score to select the most factually reliable reasoning trajectory.

\subsubsection{RAG with Tools and Rules}

\textbf{RAG with Tools} 

With some tools, the performances of RAG are greatly improved by avoiding letting LLM conduct problems that they are not good at. In this section, different tools will be discussed including 1. memory and 2. coding. When addressing long-context or multi-hop problems, incorporating a dedicated memory module to store refined information significantly enhances RAG performance. In the Retriever-and-Memory approach, an adaptive note mechanism iteratively retrieves relevant documents, refines the extracted information, and updates the note \cite{wang2025deepnotenotecentricdeepretrievalaugmented}. Firstly, with the initial query $x_0$ and initial retrieved documents $z_0$, an initial memory note $N_0$ is constructed. Then in the process of "Note-Centric Adaptive Retrieval", for the $t$-th iteration 1. the query $x_t$ is refined based on previous queries $x_1, \dots, x_{t-1}$ and current memory note $N_{t-1}$, and 2. the refined query $x_t$ is utilized to retrieve documents $z_t$, and 3. the memory note $N_t$ is refined by the retrieved documents $z_t$ and evaluated whether further retrieval is required. Eventually, the converged optimized memory note $N_{opt}$ is ultimately combined with the initial query $x_0$ to generate accurate responses. Similarly, TC-RAG \cite{jiang2024tcragturingcompleteragscasestudy} employs a Last-In-First-Out (LIFO) stack for memory management, where new information is added to the stack, and irrelevant or incorrect data is removed. MemoRAG \cite{qian2025memoragboostinglongcontext} adopts a dual-system framework. First, a lightweight global memory module—trained with Reinforcement Learning from Generation Feedback (RLGF)—compresses the full context into compact representations and provides draft answer clues. Second, a more expressive LLM generates the final response using evidence retrieved based on these clues. In practice, the database is pre-compressed into the memory module. At inference time, a query is first answered in draft form using only the memory, without retrieval. This draft answer, together with the original query, then guides retrieval of relevant documents, from which the LLM produces the final output.

\textbf{RAG with Rules}

In RAG, rules are frequently applied manually in the prompt template which will concatenate the initial query and retrieved documents before being sent to LLM for generating the final answer. Instead, rules can also be selected from a provided rule pool or learned from dataset. Some research has been done on this topic. In \cite{chen2025ruleragruleguidedretrievalaugmentedgeneration}, RuleRAG select specific rules from a rule pool to enrich query for retrieval documents and guide LLM for Q\&A in a specific manner. The authors proposed 1. RuleRAG-ICL and 2. RuleRAG-FT. In RuleRAG-ICL, a query is applied to retrieve relevant rules through cosine similarity in the embedding space by SentenceBERT. Then, for each rule, it is concatenated with the query to retrieve its most important documents. Next, all the rules and retrieved documents are mixed with the query for the final answer generation. RuleRAG-FT enhances RuleRAG-ICL with another fine-tuning stage with paired data using contrastive learning.

Another direction directly generates rules from LLM for answer generation. In \cite{zhang2024ruaglearnedruleaugmentedgenerationlarge}, Rule-Augmented Generation (RuAG) was proposed to automatically generate logic rules from dataset and utilized the learned logic rules in LLM prompt to generate the answer. To begin with, boolean-valued predicates were transformed from features, and then impossible or irrevalent predicates were removed. Next, logic rules were searched from the remained predicates using MCTS until the length or reward threshold was reached where state referred to partial logic rules, action referred to adding a predicate and reward referred to evaluating the logic rules. Lastly, the generated logic were 1. cleaned by removing replicate or low quality, 2. translated to natural language, 3. selected to fit given context window and 4. applied for final answer generation.

\subsubsection{Collaborative RAG}

Lots of high quality retrievers have been pretrained. However, the matching of the retriever and the generator has not been optimized. Some of the works in this subsection reply on fine-tuning, a little deviated from inference-time scaling topic. However, it is too important to remove them.

\begin{figure}
    \centering
    \includegraphics[width=\linewidth]{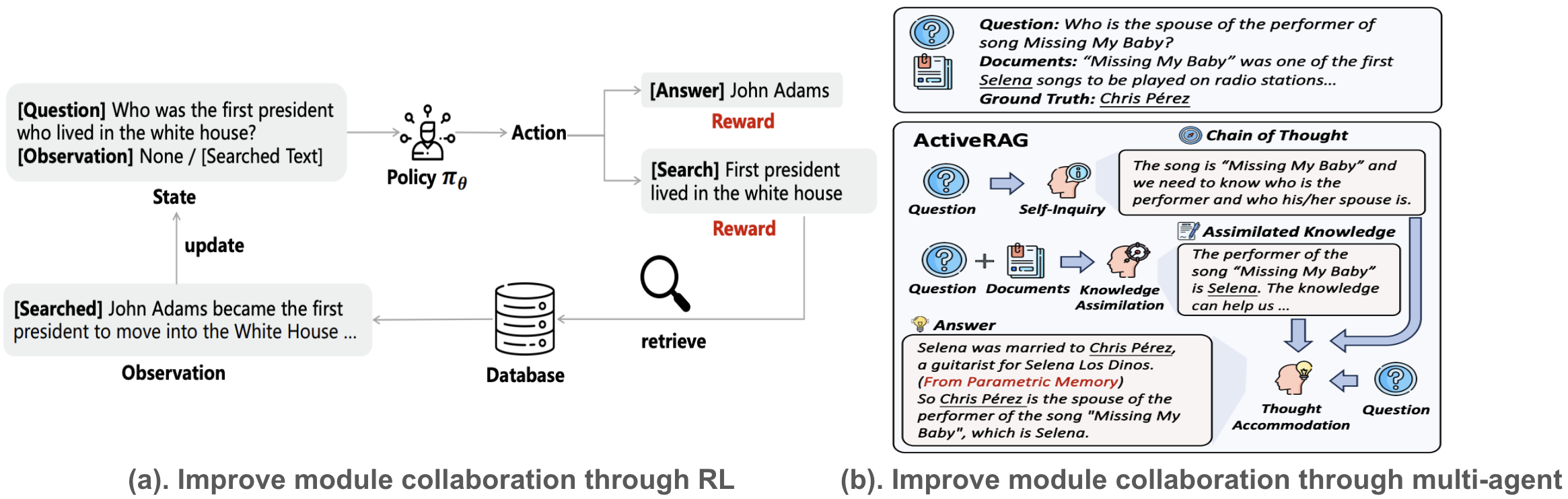}
    \caption{Collaborative RAG: (a) fine-tune all modules through RL to improve collaboration and (b) use multi-agent to improve collaboration}
    \label{fig: collaborative RAG}
\end{figure}

RePlug \cite{shi2023replugretrievalaugmentedblackboxlanguage} proposed to utilize the LLM's response generation probability based on query and retrieved documents, i.e., $P(y|z,x)$ as information to improve the performances of the retriever. To be more specific, the probabaility of retrieving documents $z$ from x, i.e., $P(z|x)= \frac{e^{s(z,x)/\gamma}}{\sum_{z' \in D'} e^{s(z',x)/\gamma}}$ where $s(z, x)=cos(E(z), E(x))$ measured the relevancy between $z$ and $x$ through cosine similarity between embedding. The probability of generator generating document $z$ from query $x$ and response $y$, i.e., $Q(z|x,y)$ was computed through Bayes theorem $Q(z|x, y) = \frac{e^{P(y|z,x)/\beta}}{\sum_{z' \in D'} e^{P(y|z',x)/\beta}}$. Lastly, the KL divergence between $P(z|x)$ and $Q(z|x, y)$ is minimized to optimize the retriever so that it will have a better performance when working with the generator. In the inference, RePlug firstly retrieves multiple documents, and then apply each document with the query and send to LLM to compute a next token probability. By ensembling across different retrieved documents, the final next token probability is obtained and then sampled to obtain the final response. In \cite{gao2025smartragjointlylearnragrelated}, RAG is trained using SFT for warmup and reinforcement learning to coordinate different modules and balance retrieval and answering as shown in Figure \ref{fig: collaborative RAG} (a). In the RL setting, the action includes "Answer" to generate answer or "Retrieval" to rewrite query and retrieve and the reward includes the correctness of the final answer and each retrieval penalty. When the action is predicted or when the maximum retrieval quota is reached, the answer will be generated and the reward will be collected to update the policy. In \cite{chen2025improvingretrievalaugmentedgenerationmultiagent}, the authors model RAG in a multi-agent reinforcement learning framework to optimize them together. The system is mainly composed of 1. query rewriter, 2. retriever, 3. selector and 4. generator. A common LLM is utilized as the backbone of these modules, and a share reward is utilized for these different modules and each module has their own reward like a query rewriting quality measurement. Lastly, PPO is utilized to optimize the LLM to maximize the accumulated rewards. 

In CoRAG \cite{muhamed2025coragcollaborativeretrievalaugmentedgeneration}, the authors extend RAG using local datasets of different clients into a collaborative setting. For each agent, it continuously trains the retriever and generator based on its own private data. Then, after training, local updates are aggregated into a global model for the next iteration of collaborative training. In the inference time, the same global retriever and generator models will be utilized across different agents, the retrieval process will happen in both the shared central dataset and agent-specific dataset before being sent to LLM for final answer generation. This methodology preservers the data privacy of different agents and improve the performances of RAG for different clients. In ActiveRAG \cite{xu2024activeragautonomouslyknowledgeassimilation}, the authors propose to collaborate among multiple agents to improve performances on downstream tasks as shown in Figure \ref{fig: collaborative RAG} (b). The multi-agents include 1. self-inquiry agent for generating initial response based on LLM parameters, 2. knowledge assimilation agent for analyzing the retrieved documents to extract useful insights and 3. thought accommodation agent to refine initial responses by resolving the confilicts with external knowledge.

\subsection{Muti-modal RAG}

\begin{figure}
    \centering
    \includegraphics[width=\linewidth]{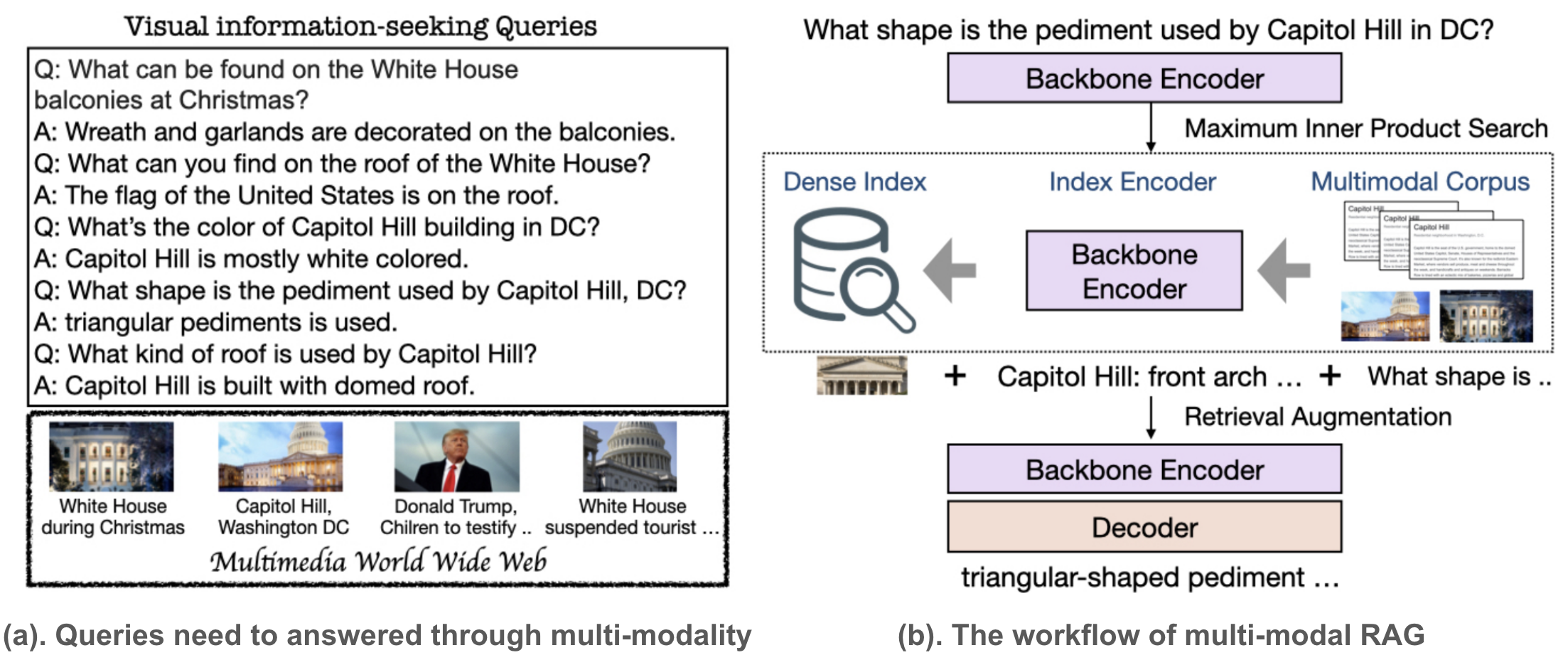}
    \caption{Muti-modal RAG: (a) some example questions that need to be answer using muti-model RAG and (b) the general workflow of multi-modal RAG}
    \label{fig: multi-modal RAG}
\end{figure}

Previous work focuses on Q\&A in the text domain. However, lots of questions need to be answered across multiple modalities like text, speech and image as shown in Figure \ref{fig: multi-modal RAG}. To extend RAG to multi-modality, a multi-modal encoders is utilized to extract embedding from different modalities into a common embedding space for retrieval. Then, a multi-modal LLM is applied to generate responses based on query and retrieved multi-modal documents. Different encoders are utilized for embedding extraction from images and texts \cite{chen2022muragmultimodalretrievalaugmentedgenerator}, \cite{yu2025visragvisionbasedretrievalaugmentedgeneration}, \cite{cho2024m3docragmultimodalretrievalneed}. To mitigate the fairness of image generation, FairRAG \cite{shrestha2024fairragfairhumangeneration} utilize a fair retrieval system to retrieve demographically balanced reference images and then apply these reference images in the output generation. 

In LLM-based ASR RAG (LARAG) \cite{li2024laragenhancingllmbasedasraccuracy} where ASR refers to automatic speech recognition, the authors applied RAG in the audio domain for speech transcript correction. To begin with, a speech tokenizer like (CTC/AED) is utilized to obtain speech tokens from speech which is aligned with a corresponding text. Then, a speech database of speech-text pairs is constructed using a pre-trained ASR model, and it is stored as a speech inverted index from speech token to speech transcription to facilicate term frequency search. When given an audio, the speech tokens are utilized obtain speech token and then retrieve similar speech tokens from the previously built database with token-level KNN. The retrieved results are grouped and scored at the sequence-level, with low-similarity matches filtered and error-prone tokens pruned using N-best list analysis. Next, a feedforward network transforms the speech embedding to produce candidate text embeddings. Lastly, speech embeddings, corresponding text embeddings, current input speech tokens and N-best candidate text embeddings are sent to LLM for generating the final correct text transcription.

\section{Input: Few-Shot}
Few-shot prompt tuning is one of the most commonly used methods to improve LLM performance. In zero-shot prompt tuning, the model is provided only with a task description or instructions, without any labeled examples \cite{kojima2023largelanguagemodelszeroshot}. Surprisingly, LLMs often perform well in zero-shot settings, demonstrating strong reasoning capabilities even without examples. In few-shot prompt tuning, the model is given a small number of labeled examples (typically 2–10) alongside the task description. A special case is one-shot prompt tuning, where exactly one labeled example is included. The success of GPT-3 demonstrated the effectiveness of few-shot prompting, particularly in industry scenarios where a specific output format is desired without full model fine-tuning \cite{brown2020languagemodelsfewshotlearners}. 

Research has investigated how and why few-shot prompting improves LLM performance \cite{min2022rethinkingroledemonstrationsmakes}. To begin with, they discover that replacing in-distribution examples with out-of-distribution examples significantly reduces performance. Besides, they discovered that direct models, which predict labels directly, rely heavily on correct labels in few-shot examples. In contrast, channel models, which predict the likelihood of input given a label, are less sensitive to the correctness of the examples. Lastly, they observed that the structure of input-output demonstrations matters: leaving only the input or only the output substantially decreases performance.

\begin{figure}
    \centering
    \includegraphics[width=0.7\linewidth]{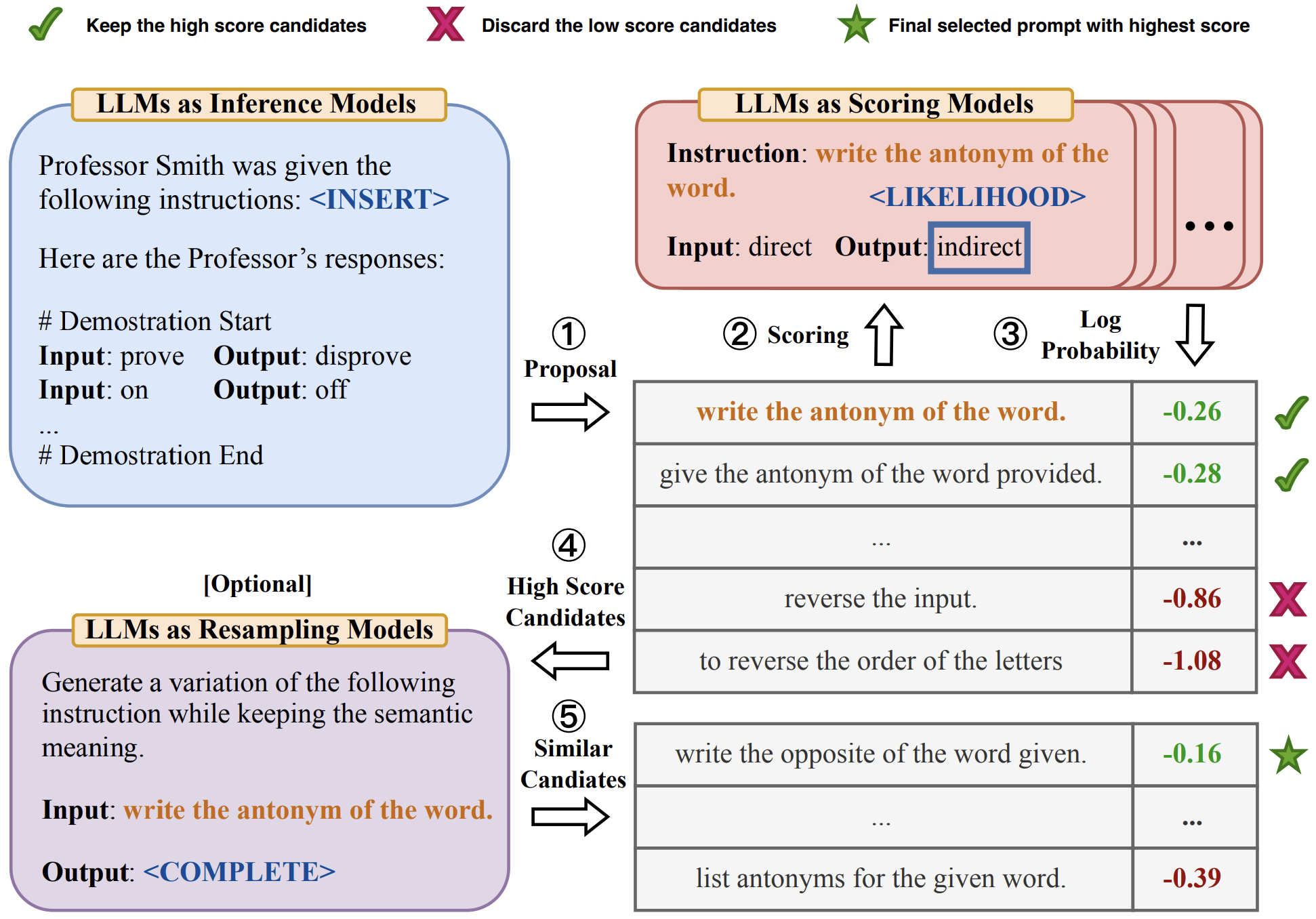}
    \caption{APE: automatic prompt engineer to generate lots of prompts and scoring based on a small set of test examples using answer accuracy and log probability}
    \label{fig:fewshot}
\end{figure}

Another research direction focuses on automatically generating prompts to further improve LLM performance. In Automatic Prompt Engineer (APE) \cite{zhou2023largelanguagemodelshumanlevel}, prompts are generated using two strategies:
\begin{itemize}
    \item \textit{Forward generation}: The model is given task examples and instructed to continue a story-like setup, completing the prompt.
    \item \textit{Reverse generation (infilling)}: The model fills in blanks in the middle of input-output examples.
\end{itemize}
Generated prompts are then scored on a small validation set using metrics like answer accuracy and log probability. The top prompts are selected, and the process iterates for multiple epochs to optimize prompt quality (Figure \ref{fig:fewshot}). Similarly, Automate-CoT \cite{shum2024automaticpromptaugmentationselection} automates the generation of prompts with a focus on: 1. Order Sensitivity, 2. Complexity Sensitivity, 3. Diversity Sensitivity and 4. Style Sensitivity. Here, the LLM generates multiple reasoning paths, low-quality paths are filtered, and a Variance-Reduced Policy Gradient (VR-PG) method estimates the importance of specific examples for downstream tasks. Finally, the highest-quality examples are selected and ordered to form the final prompt for the LLM.

\section{Conclusion}

Our review has explored the shift in LLM research from the traditional focus on scaling \textbf{training FLOPs and data}---a path increasingly constrained by the scarcity of high-quality data---to \textbf{inference-time scaling}. This paradigm uses increased computational resources during inference to significantly enhance LLM performance on downstream tasks.

We systematically organized the techniques contributing to this shift based on two key perspectives: the \textbf{output} and the \textbf{input}.

On the \textbf{output} side, we discussed a range of methods for more sophisticated generation, including:
\begin{itemize}
    \item \textbf{Reasoning} techniques like CoT, ToT, and ReAct.
    \item \textbf{Search} methods such as MCTS and beam search.
    \item Advanced \textbf{decoding} strategies, including Best-of-N, speculative decoding, and constrained decoding.
    \item Methods focused on \textbf{training for long CoT}, such as RLVR and GRPO.
    \item The emerging fields of \textbf{multi-modal reasoning} and \textbf{model ensemble}.
\end{itemize}

Regarding the \textbf{input} side, we examined \textbf{few-shot learning} and dedicated significant attention to \textbf{RAG}. Our analysis of RAG covered its critical components and latest advancements:
\begin{itemize}
    \item \textbf{Query expansion}.
    \item The role of high-quality \textbf{data}.
    \item Improvements in \textbf{retrieval and reranker} mechanisms.
    \item Optimization of the \textbf{LLM generation} process.
    \item The integration of \textbf{multi-modal RAG}.
\end{itemize}

In summary, inference-time scaling provides a powerful, practical avenue for improving LLM capabilities in the face of data limitations. The rich landscape of techniques reviewed---from complex reasoning and search at the output level to sophisticated RAG architectures at the input level---underscores the depth and potential of this rapidly evolving area of research.

\section{Symbols}

$G$: knowledge graph

$E$: Edges in knowledge graph

$N$: Nodes in knowledge graph

$r_j$: the average reward of node $j$ in MCTS

$N$: the total number of visits to the parent in MCTS

$n_j$: the number of visits to node $j$ in MCTS

$P(y|x)$: the probability of generating response $y$ given prompt $x$

$s_t$: state at step $t$ for MDP

$a_t$: action at step $t$ for MDP

$p$: draft model in speculative decoding

$q$: target model in speculative decoding

$M$: large base model

$m^+$: fine-tuned domain model

$m^-$: anti-domain model

$\text{KL}$: KL divergence

$z$: retrieved documents in RAG

$\text{KNN}$: K nearest neighborhood

$N_{\text{opt}}$: the optimal memory note 

\bibliographystyle{unsrt}  
\bibliography{references}

\end{document}